\documentclass[acmtog]{acmart}
\AtBeginDocument{%
  }

\setcopyright{cc}
\setcctype{by}
\acmJournal{TOG}
\acmYear{2026} \acmVolume{45} \acmNumber{4} \acmArticle{52}
\acmMonth{7} \acmDOI{10.1145/3811390}
\acmConference[SIGGRAPH 2026]{SIGGRAPH’26 Conference Proceedings}{July 19--23, 2026}{Los Angeles, CA}
\acmISBN{978-1-4503-XXXX-X/2018/06}

\usepackage{hyperref}
\usepackage{cleveref}

\usepackage{booktabs}

\usepackage[export]{adjustbox}
\usepackage{float}
\usepackage{makecell}
\usepackage{comment}
\usepackage{overpic}
\usepackage{multirow} 
\usepackage{subcaption}
\usepackage[export]{adjustbox}
\usepackage{enumitem}
\usepackage{wrapfig}

\usepackage{algorithm}
\usepackage{algpseudocode}
\usepackage{amsmath}

\begin{document}

\title{DissolveStereo: Coarse Depth Injection for Zero-Shot Stereo Video Generation}

\author{Jian Shi}
\affiliation{%
  \institution{KAUST}
  \city{Thuwal}
  \country{Saudi Arabia}}
\email{jian.shi@kaust.edu.sa}

\author{Qian Wang}
\affiliation{%
  \institution{KAUST}
  \city{Thuwal}
  \country{Saudi Arabia}}
\email{qian.wang@kaust.edu.sa}

\author{ZhenYu Li}
\affiliation{%
  \institution{KAUST}
  \city{Thuwal}
  \country{Saudi Arabia}}
\email{zhenyu.li.1@kaust.edu.sa}

\author{WenQing Cui}
\affiliation{%
  \institution{KAUST}
  \city{Thuwal}
  \country{Saudi Arabia}}
\email{wenqing.cui@kaust.edu.sa}

\author{Ramzi Idoughi}
\affiliation{%
  \institution{KAUST}
  \city{Thuwal}
  \country{Saudi Arabia}}
\email{ramzi.idoughi@kaust.edu.sa}

\author{Peter Wonka}
\affiliation{%
  \institution{KAUST}
  \city{Thuwal}
  \country{Saudi Arabia}}
\email{peter.wonka@kaust.edu.sa}

\renewcommand{\shortauthors}{Shi et al.}

\newcommand{\modified}[1]{{\textcolor{red}{#1}}}
\begin{abstract}
Generating high-quality stereo videos requires consistent depth perception and temporal coherence across frames. Despite advances in image and video synthesis using diffusion models, producing high-quality stereo videos remains a challenging task due to the difficulty of maintaining consistent temporal and spatial coherence between left and right views.
We introduce \textit{DissolveStereo}, a novel framework for zero-shot stereo video generation that leverages video diffusion priors without requiring paired training data. Our key innovations include a noisy restart strategy to initialize stereo-aware latent representations and an iterative refinement process that progressively harmonizes the latent space, addressing issues like temporal flickering and view inconsistencies.
Importantly, we propose the use of dissolved depth maps to streamline latent space operations by reducing high-frequency depth information.
Our comprehensive evaluations, including quantitative metrics and user studies, demonstrate that \textit{DissolveStereo} produces high-quality stereo videos with enhanced depth consistency and temporal smoothness. In terms of epipolar consistency, our method achieves an $11.7\%$ improvement in MEt3R score over the current state-of-the-art. Furthermore, user studies indicate strong perceptual gains over the previous arts, with an $8.0\%$ higher perceived frame quality and $10.9\%$ higher perceived temporal coherence.
Our code is in~\url{https://github.com/shijianjian/DissolveStereo}.
\end{abstract}


\settopmatter{printacmref=false} 
\renewcommand\footnotetextcopyrightpermission[1]{}

  
\begin{teaserfigure}
    \includegraphics[width=.98\linewidth,trim={.8cm 0 .8cm 0},clip]{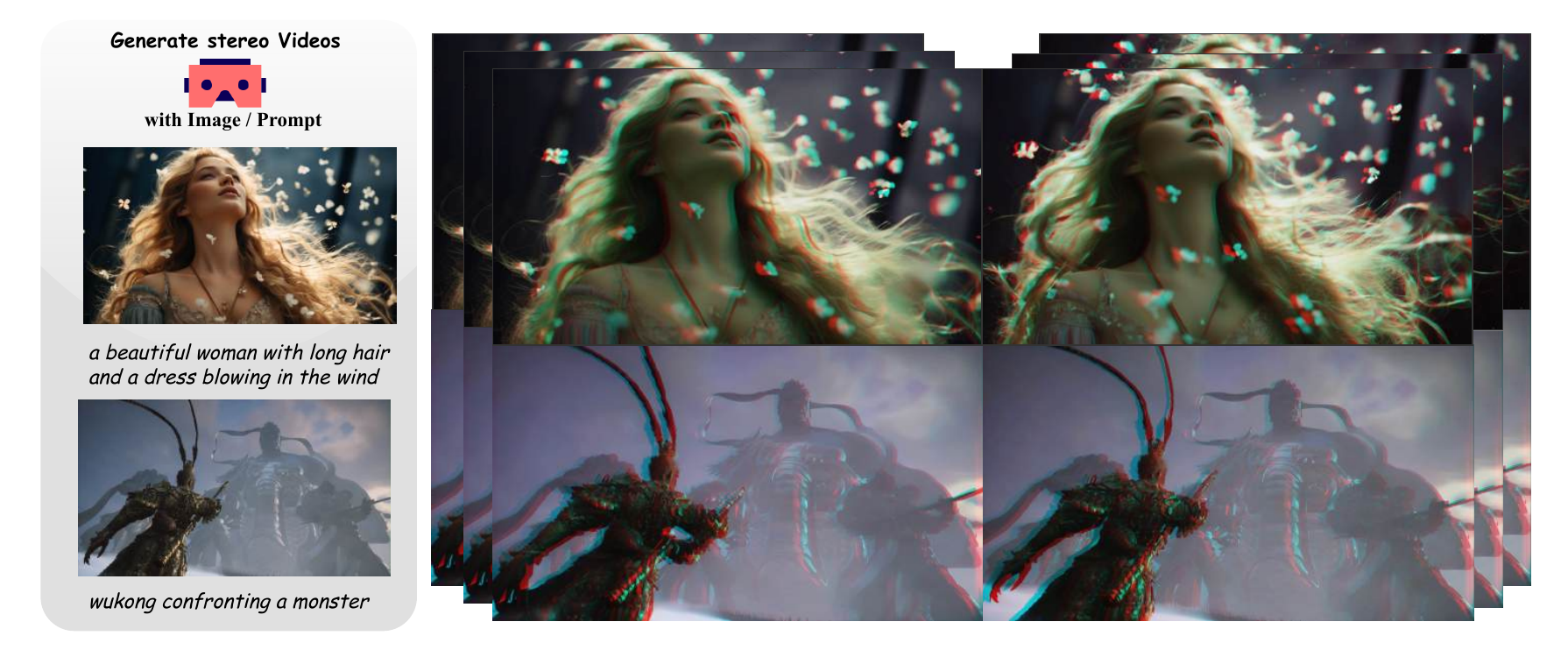}
    \caption{
    From only a single image and text prompt as input (left), our method generates stereo video sequences, visualized as an anaglyph (right).
    }
    \Description{
    From only a single image and text prompt as input (left), our method generates stereo video sequences, visualized as an anaglyph (right).}
    \vspace{3em}
  \label{fig:teaser}
\end{teaserfigure}

\maketitle

\section{Introduction}
\label{sec:intro}
The rapid adoption of head-mounted displays for virtual reality (VR) has created a growing demand for high-quality stereo videos, which provide immersive depth perception through paired left and right views. Given the lack of naturally acquired stereo videos, we propose exploring generative methods, particularly diffusion models, for creating these videos from scratch.

Several previous works initially focused on stereo image conversion~\cite{wang2019web, Shih3DP20, watson2020learning, Ranftl2022}, where one view is given, and its stereo pair is synthesized using techniques such as depth estimation and image warping. These approaches prioritized geometric accuracy over content generation.
With the rise of diffusion models~\cite{ho2020denoising, rombach2022high, ramesh2022hierarchical}, StereoDiffusion~\cite{wang2024stereodiffusion} demonstrated the potential of zero-shot stereo image generation to synthesize both views simultaneously.

Recent studies have extended image stereo conversion techniques to video, emphasizing the importance of temporal consistency to produce smooth and coherent stereo video sequences~\cite{shi2024immersepro, zhao2024stereocrafter}. Yet, zero-shot stereo video generation remains unexplored. This task is inherently more complex than image generation, as it requires meticulous handling of depth cues to produce realistic parallax and consistent depth perception in both views~\cite{hartley2003multiple}, while maintaining temporal coherence across frames and inter-view consistency between stereo pairs. Advanced video depth estimation models have improved temporal consistency by accurately capturing and maintaining depth information across video frames~\cite{luo2020consistent,kopf2021robust,hu2024-DepthCrafter,chen2025video}. They played a crucial role in enhancing the overall visual fidelity and temporal smoothness of the output of existing video generation models. However, these methods lack mechanisms for dynamically adapting depth information during the diffusion process, which is necessary to accurately represent scene dynamics in stereo video generation.

To address these challenges, we introduce \textbf{DissolveStereo}, a novel framework for zero-shot stereo video generation that leverages video diffusion priors to generate high-quality stereo videos without the need for paired training data. 
The straightforward solution to this problem is to break it down into two known components: video generation and stereo video conversion.
Instead of pixel-level generation and conversion, we propose a stronger coupling of generation and conversion by improving and refining the consistency within the latent features of a video diffusion model.
One significant advantage is that our method does not require precise disparity maps. We introduce the concept of \textit{dissolved depth maps}, which retain only the low-frequency structural depth information. Our key insight is that latent-space warping benefits more from coarse geometry than fine-grained depth. Unlike image-space warping, where high-frequency depth helps preserve pixel-level accuracy, latent-space warping prioritizes coarse geometry and semantic consistency without requiring accurate depth maps.
Furthermore, our approach seamlessly integrates advanced video depth estimation into the diffusion-based synthesis process, ensuring both depth consistency and temporal coherence.
Our main contributions are:
\begin{itemize}[itemsep=0pt, parsep=0pt, topsep=-.5em, leftmargin=*]

    \item We enhance latent consistency in stereo generation by employing \textit{noisy restart} to create stereo-aware initial latents, followed by \textit{iterative refinement} that injects controlled noise into the diffusion process, progressively improving the harmony of the latent space.
    \item We introduce \textit{dissolved depth maps}, a novel depth representation that retains only low-frequency structural depth while suppressing high-frequency information, effectively reducing artifacts in the generated right view.
    \item We perform thorough evaluations, including statistical analysis and user studies, to validate our method for high-quality stereo video generation. Our approach achieves a new state-of-the-art in epipolar consistency, with user studies confirming significant improvements in visual clarity and temporal smoothness.
\end{itemize}

\begin{figure*}
    \centering
    \includegraphics[width=.98\textwidth]{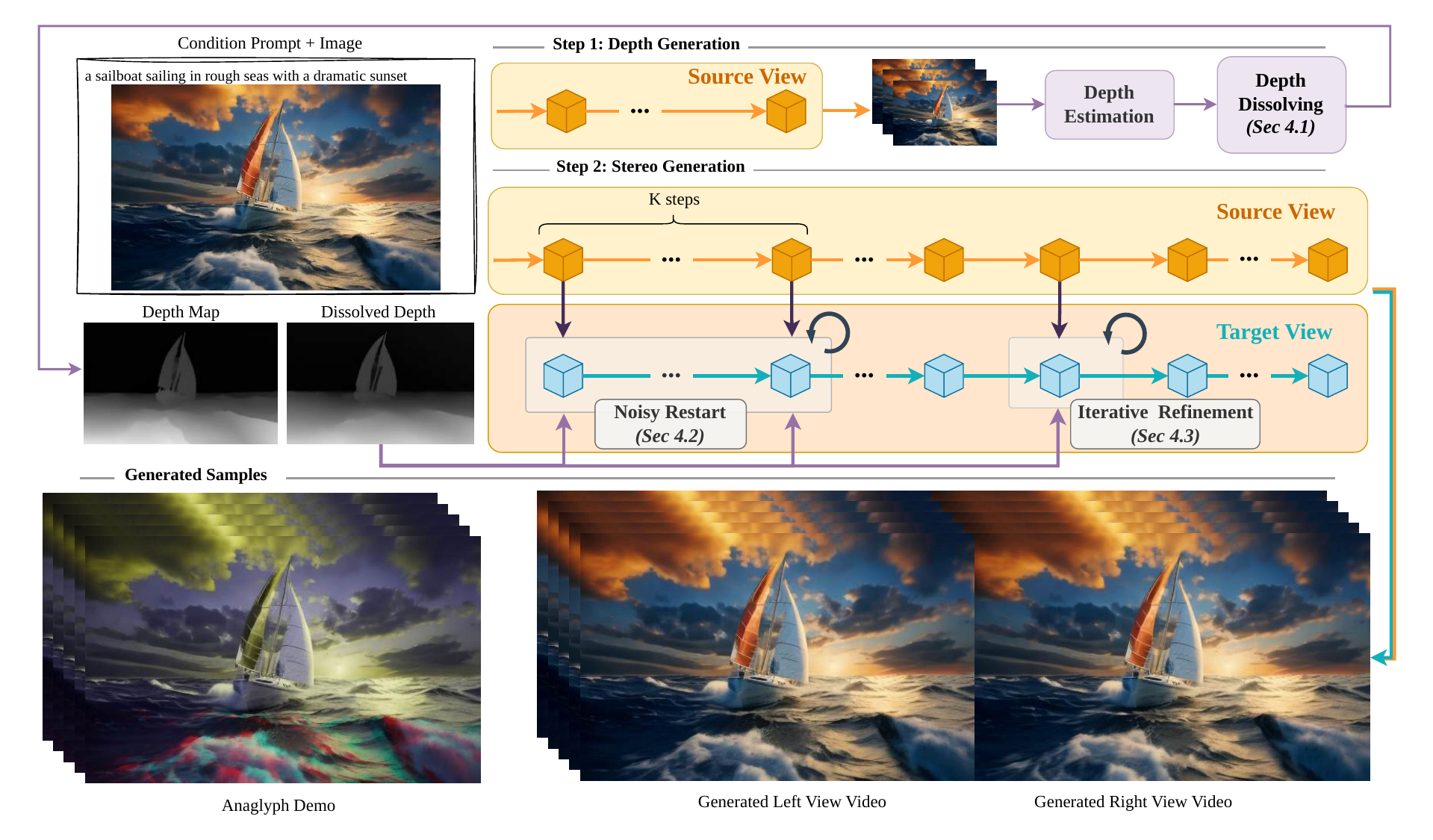}
    \setlength{\abovecaptionskip}{0.5em}
    \Description{An overview of the \textit{DissolveStereo} pipeline. Top: Given an image and text prompt, we first generate a source/left-view video, then obtain its depth maps with a depth estimation model. A \textbf{depth dissolving} technique (\Cref{sec:dslv_depth}) is used to improve compatibility with latent warping. Along with the dissolved depth maps, we perform stereo generation conditioned on the same image and prompt. Our method contains two main components: (1) \textbf{Noisy Restart} for a robust initial latent estimation (\Cref{sec:noisy_restart}) and (2) \textbf{Iterative Refinement} for latent refinement (\Cref{sec:iter_refine}). These components act on target/right view latents (blue) for temporal and inter-view consistency with the source/left view (orange). Bottom: Our pipeline generates stereo videos with a strong and consistent stereoscopic effect.}
    \caption{An overview of the \textit{DissolveStereo} pipeline. Top: Given an image and text prompt, we first generate a source/left-view video, then obtain its depth maps with a depth estimation model. A \textbf{depth dissolving} technique (\Cref{sec:dslv_depth}) is used to improve compatibility with latent warping. Along with the dissolved depth maps, we perform stereo generation conditioned on the same image and prompt. Our method contains two main components: (1) \textbf{Noisy Restart} for a robust initial latent estimation (\Cref{sec:noisy_restart}) and (2) \textbf{Iterative Refinement} for latent refinement (\Cref{sec:iter_refine}). These components act on target/right view latents (blue) for temporal and inter-view consistency with the source/left view (orange). Bottom: Our pipeline generates stereo videos with a strong and consistent stereoscopic effect.}
    \label{fig:main_pipeline}
\end{figure*}

\section{Related Works}
\label{sec:related_work}

\subsection{Novel-View Synthesis}
Novel-view synthesis task aims to generate images from new camera perspectives based on one or more source images.
Recent novel-view synthesis works such as ~\cite{Li_2023_CVPR,liu2023robust,yu2023hifi,tang2023make,sun2023dreamcraft3d,yu2024viewcrafter,bai2024syncammaster} demonstrate good results in creating a stereo pair of a given scene. However, these methods require scene-specific optimization, which limits their applicability to video data.
Another category of approaches~\cite{shriram2024realmdreamer,chung2023luciddreamer,zhang2024text2nerf} employs the depth-warping technique to synthesize novel views and subsequently refines the warped images. These approaches suffer from visual artifacts in inpainted regions, particularly in complex scenes with large disparities or occlusions. More importantly, these methods cannot enforce temporal consistency, which is an important requirement for handling video-based novel-view synthesis. 
The Collaborative Video Diffusion (CVD) technique~\citep{kuang2024cvd} utilized a cross-video synchronization module to directly generate multi-view videos for predefined camera trajectories. Another line of work~\citep{yu20244real,bahmani20244dfy,zhang20244diffusion} focuses on explicit 3D scene construction with temporal dynamics, enabling the generation of novel-view videos. However, these methods are limited to simple object-centric scenes and struggle with complex scenes. In contrast, our task is not arbitrary viewpoint synthesis. We target epipolar-constrained stereoscopic video generation, a more structured and geometrically grounded problem. By leveraging a pretrained video prior, we achieve stereoscopic video generation without requiring stereo supervision or scene-specific optimization.

\subsection{Stereo Content Generation}
Stereo content generation has evolved significantly from traditional disparity estimation and view synthesis methods to modern deep learning approaches that enhance both depth accuracy and visual realism. Early approaches~\cite{xie2016deep3d, wang2019web} utilized neural networks and generative networks to predict disparity maps and synthesize the corresponding stereo image pairs. Recent advances have extended these techniques to video by incorporating temporal coherence, thereby creating smooth and immersive stereo video sequences ~\cite{zhang2022temporal3d,shi2024immersepro, zhao2024stereocrafter}. Notably, diffusion models have been successfully adapted for zero-shot stereo image synthesis~\cite{wang2024stereodiffusion}. However, applying these models to stereo video generation remains challenging, primarily due to the need to maintain both spatial depth fidelity and temporal consistency across frames. 
A concurrent work, namely SVG~\cite{dai2024svg}, proposes a \textit{frame matrix} approach, which involves placing multiple camera views and warping the latent space on every DDIM (Denoising Diffusion Implicit Models) sampling step. In contrast, our method generates new viewpoints without the need for warping at every step, resulting in approximately 3 times faster generation speed.

\section{Preliminaries}
\label{sec:preliminary}

\subsection{Image Warping for Stereo View Synthesis.} Generating a stereo pair from a single input image involves warping the image according to a disparity map $d$ that can be computed as:
\begin{equation}
    d(x,y)=\frac{f\cdot B}{I_{depth}(x,y)},
\label{eq:disparity}
\end{equation}
where $f$ is the focal length, $B$ is an adjustable baseline distance, and $I_{depth} (x,y)$ is the depth value at pixel location $(x,y)$. Given the disparity map, the right-view image can be obtained by warping the original 2D image with:
\begin{equation}
    I_{right}(x-d(x,y),y)=I(x,y).
\label{eq:warp}
\end{equation}
This shift creates a horizontal parallax, mimicking the viewpoint difference.
However, the warping operation requires high-precision depth maps and can introduce artifacts and misalignments. In particular, due to occlusions, some pixels in the right view may not have a corresponding pixel in the original image.
Previous works~\cite{watson2020learning,mehl2024stereo,zhao2024stereocrafter} handled these occluded and disoccluded regions with inpainting.

\subsection{Stereo Pair Generation with Diffusion Models}
Recent works, such as~\cite{wang2024stereodiffusion}, propose warping the latent space of an image diffusion model. This is feasible because the early stages of the diffusion sampling process involve low-frequency signals that primarily define the overall structure of the generated image, while the later stages, corresponding to a large time step $t$, refine high-frequency details.
Formally, each DDIM denoising step can be described as:
\begin{equation}
\begin{aligned}
    x_{t-1} =& \sqrt{\alpha_{t-1}}\big( \frac{x_t-\sqrt{1-\alpha_t}\epsilon_\theta^t(x_t)}{\sqrt{\alpha_t}} \\
    & + \sqrt{1-\alpha_{t-1}-\sigma_t^2\epsilon_\theta^{(t)}(x_t)} + \sigma_t \epsilon_t,
\end{aligned}
\label{eq:ddim}
\end{equation}
where $\epsilon_t \sim \mathcal{N}(0, {I})$ is Gaussian noise independent of $x_t$, and $\alpha_t$ controls the noise scale at step $t$ with $\alpha_0=1$. Since $\epsilon_t$ is the sole source of randomness in the denoising process, we share in our work the same $\epsilon$ across views by default.
Meanwhile, as demonstrated in~\Cref{sec:noisy_restart,sec:exp}, we find the noise level $\epsilon$ plays an important role in producing stereoscopic effects.

\subsection{Cross-View Attention}
\label{sec:cross_view_attention}
Maintaining consistency between the left- and right-view latents is crucial for stereo content generation. To produce multi-view consistent content, several works~\cite{cao2023masactrl,chen2023fec,chen2024anydoor,huang2023kv,khandelwal2023infusion,mou2023dragondiffusion,nam2024dreammatcher,wang2024stereodiffusion} modify the diffusion model's attention mechanism by replacing the query, key, and value pairs, thereby ensuring coherent generation across views. In this work, we adopt \textbf{cross-view attention}, following~\cite{wang2024stereodiffusion}. Let $Q_L, K_L, V_L$ and $Q_R, K_R, V_R$ represent the query, key, and value matrices for the left and right views, respectively.
To align the right view with the left view’s features, we use $Q_L$ to attend to $K_R$ and $V_R$. The resulting cross-view feature $F_{R \to L}$ can be expressed as:
\begin{equation}
\begin{split}
F_{R \to L} &= \text{softmax}\left(\frac{Q_L K_R^\top}{\sqrt{d}}\right) \cdot V_R,
\end{split}
\end{equation}
where the left view's queries ($Q_L$) attend directly to the right view's keys ($K_R$) and values ($V_R$). This mechanism promotes stronger alignment and coherence across views by sharing essential spatial and contextual information.

\section{Method}

This section describes \textit{DissolveStereo}, which leverages video diffusion priors for zero-shot stereo-consistent video generation, without requiring stereo-paired data for training or fine-tuning.
In this work, we refer to the reference and synthesized views as the “source” (left) and “target” (right), respectively. An overview of the pipeline is shown in~\Cref{fig:main_pipeline}.
Let \(\mathbf{X} = \{x_T, x_{T-1}, \dots, x_0\}\) denote the DDIM latent sequence from a video diffusion model with \(T\) diffusion steps (e.g., $T=50$). Each latent \(x_t \in \mathbb{R}^{B \times C \times T \times H \times W}\) is a five-dimensional tensor representing a batch of video frames, where \(B\) is the batch size, \(C\) the number of channels, \(T\) the temporal dimension (number of frames), and \(H\) and \(W\) the spatial height and width, respectively.
By decoding $x_0$, we obtain a video sequence $V\in \mathbb{R}^{B \times 3 \times T \times H' \times W'}$, where $H'$ and $W'$ are the corresponding height and width of the decoded videos.
To capture the scene geometry and enable accurate latent warping, we generate depth maps \(\mathbf{D} \in \mathbb{R}^{B \times T \times H' \times W'}\), which provide per-pixel, temporally consistent depth information. Using these depth maps, the latent features are warped to achieve stereo consistency and account for parallax effects.
Let \(\Delta\) denote the disparity map, which represents the horizontal shift between the left and right stereo views. The warp operation \(\mathbf{W}(x, \Delta)\) (see~\Cref{eq:warp}) is then applied to produce the warped latent representation $x^{\text{warp}}_{t}=\mathbf{W}(x_t, \Delta)$. Details of our efficient implementation (\textbf{1000 $\times$ faster} than a traditional non-vectorized warping method) of the warping algorithm are provided in the supplementary.
This warping aligns the latent representations according to depth-induced disparities, which is essential for generating realistic stereo views.
The warping process, however, introduces blank regions \(x^{blank} \in \mathbb{R}^{B \times C \times T \times H \times W}\) in the warped latents due to occlusions or disocclusions, defined as:
\begin{equation}
    x^{blank}_{t} = 1 ~~ \text{if}~ x^{\text{warp}}_{t} ~\text{ is undefined}, ~~~~~~0 ~~\text{otherwise}.
    \label{eq:blank_areas}
\end{equation}
With the aid of a disparity map, each latent feature can be decomposed into $x^{blank}_{t}$ and $x^{\text{warp}}_{t}$ after warping. 
Maintaining consistency between these two latent parts is critical for creating meaningful and harmonized outputs for each individual view with correct depth cues.
In the diffusion process, $\epsilon_t$ serves as the sole source of randomness (see~\Cref{eq:ddim}). By iteratively injecting $\epsilon_t$ into the diffusion steps, we refine the latent features, promoting coherence and fidelity in the generated results.

In this section, we first introduce \textbf{dissolved depth maps} (\Cref{sec:dslv_depth}), which convert depth maps into lower-frequency representations to enhance the consistency of warped latents with the video diffusion prior.
Next, during stereo generation, we improve the consistency between $x^{blank}_{t}$ and $x^{\text{warp}}_{t}$ through two key steps: (1) \textbf{Noisy Restart} (\Cref{sec:noisy_restart}), which initializes a reasonable $x^{blank}$, and (2) \textbf{Iterative Refinement} (\Cref{sec:iter_refine}) that improves the consistency between $x^{blank}$ and $x^{warp}$.

\begin{figure}[b]
     \centering
     \begin{subfigure}[b]{0.49\linewidth}
         \centering
         \adjincludegraphics[width=\linewidth,trim={3px {.5\height + 3px} {.875\width} 3px},clip]{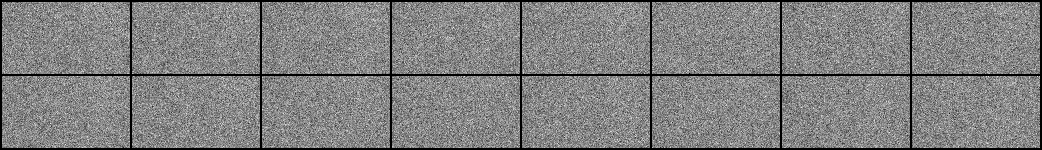}
         \caption{$t=15$}
     \end{subfigure}
     \begin{subfigure}[b]{0.49\linewidth}
         \centering
         \adjincludegraphics[width=\linewidth,trim={3px {.5\height + 3px} {.875\width} 3px},clip]{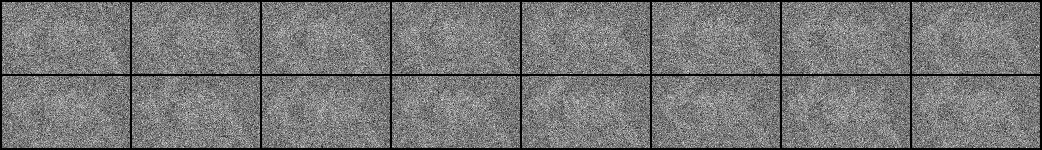}
         \caption{$t=25$}
     \end{subfigure}
     \\
     \begin{subfigure}[b]{0.49\linewidth}
         \centering
         \adjincludegraphics[width=\linewidth,trim={3px {.5\height + 3px} {.875\width} 3px},clip]{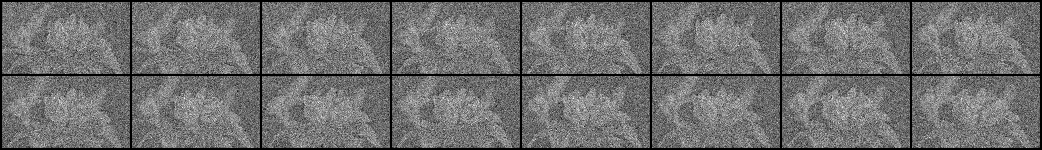}
         \caption{$t=35$}
     \end{subfigure}
     \begin{subfigure}[b]{0.49\linewidth}
         \centering
         \adjincludegraphics[width=\linewidth,trim={3px {.5\height + 3px} {.875\width} 3px},clip]{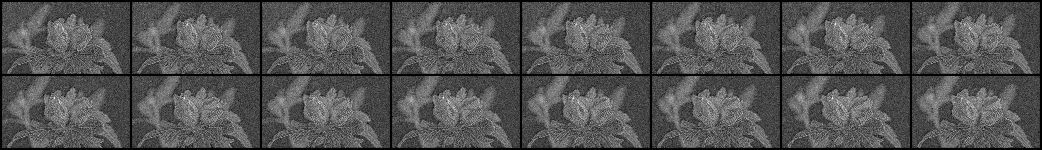}
         \caption{$t=45$}
     \end{subfigure}
    \Description{We visualize the tensor of the first timeframe in the first channel of the latent space of video diffusion. We can observe that the significant structural information does not show before $t=25$.}
    \caption{We visualize the tensor of the first timeframe in the first channel of the latent space of video diffusion. We can observe that the significant structural information does not show before $t=25$.}
    \label{fig:latent_space}
\end{figure}

\begin{figure*}
     \centering
     \begin{subfigure}[b]{0.24\linewidth}
        \includegraphics[width=1.\linewidth]{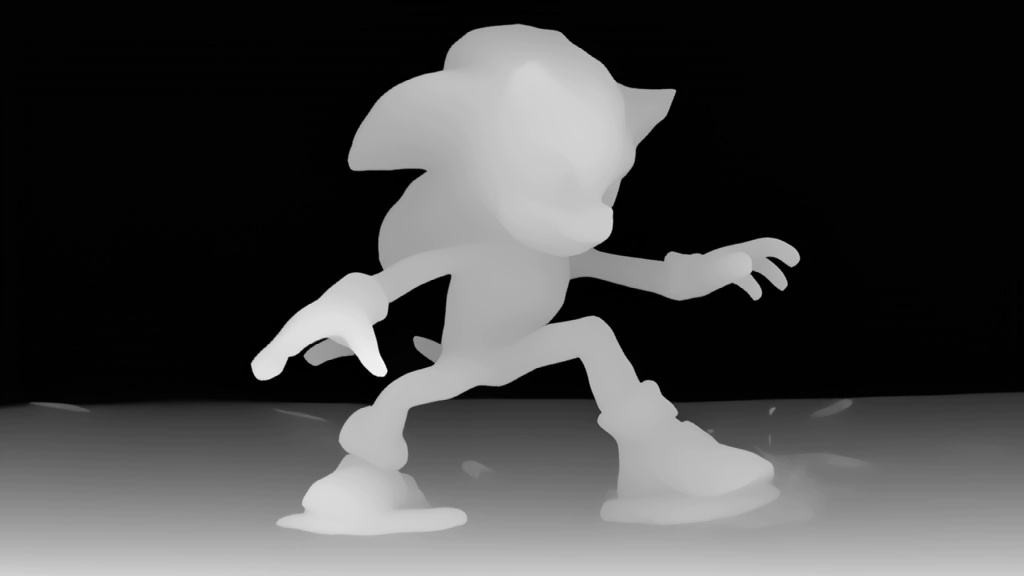}
     \end{subfigure}
     \begin{subfigure}[b]{0.24\linewidth}
        \includegraphics[width=1.\linewidth]{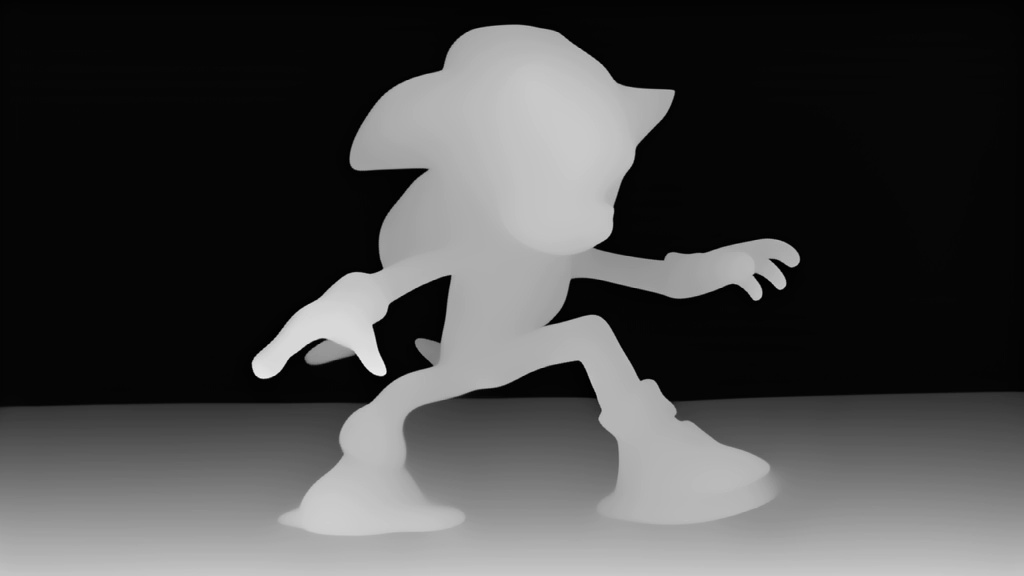}
     \end{subfigure}
     \begin{subfigure}[b]{0.24\linewidth}
        \includegraphics[width=1.\linewidth]{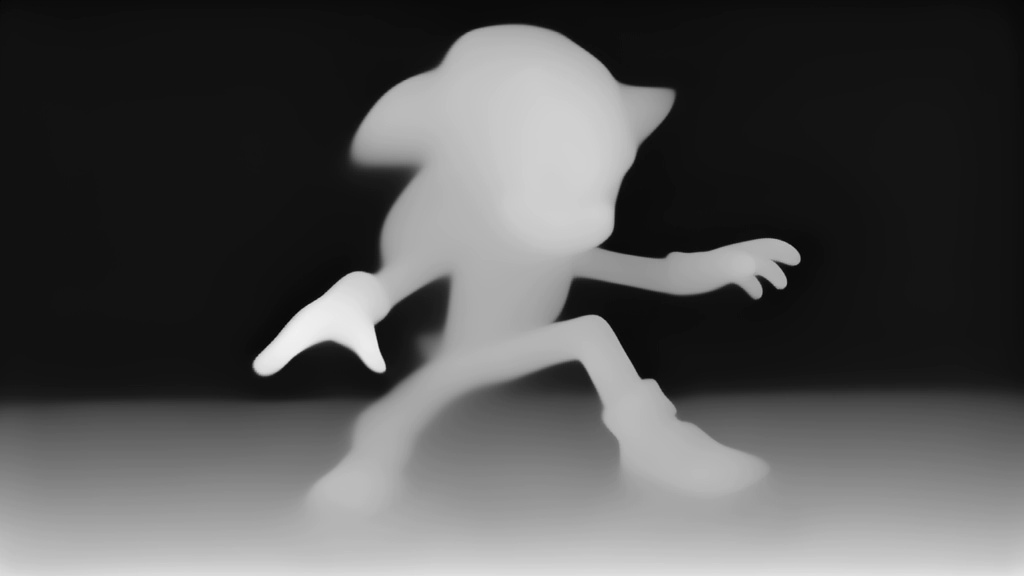}
     \end{subfigure}
     \begin{subfigure}[b]{0.24\linewidth}
        \includegraphics[width=1.\linewidth]{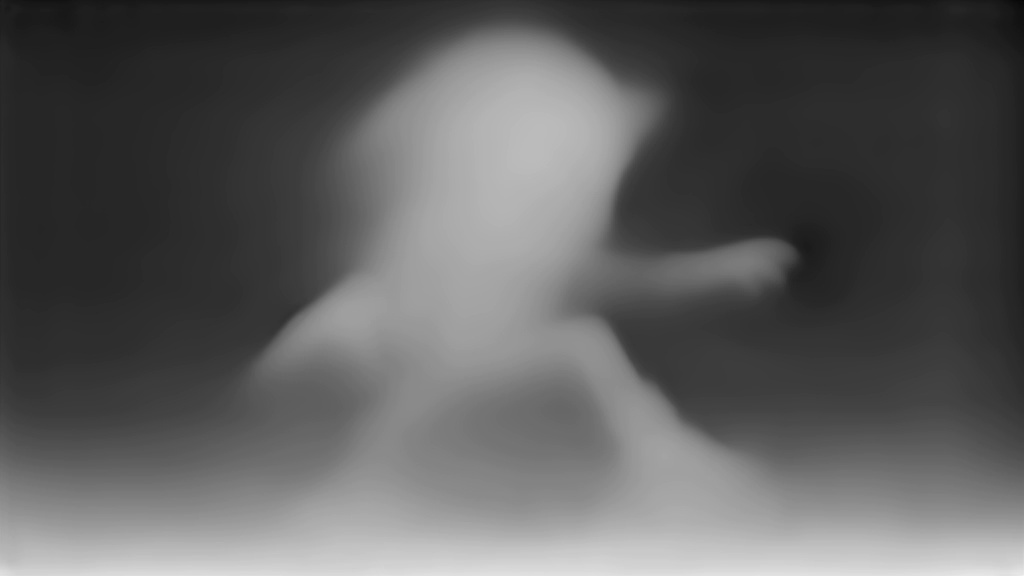}
     \end{subfigure}
     \begin{subfigure}[b]{0.24\linewidth}
        \begin{overpic}[width=\textwidth]{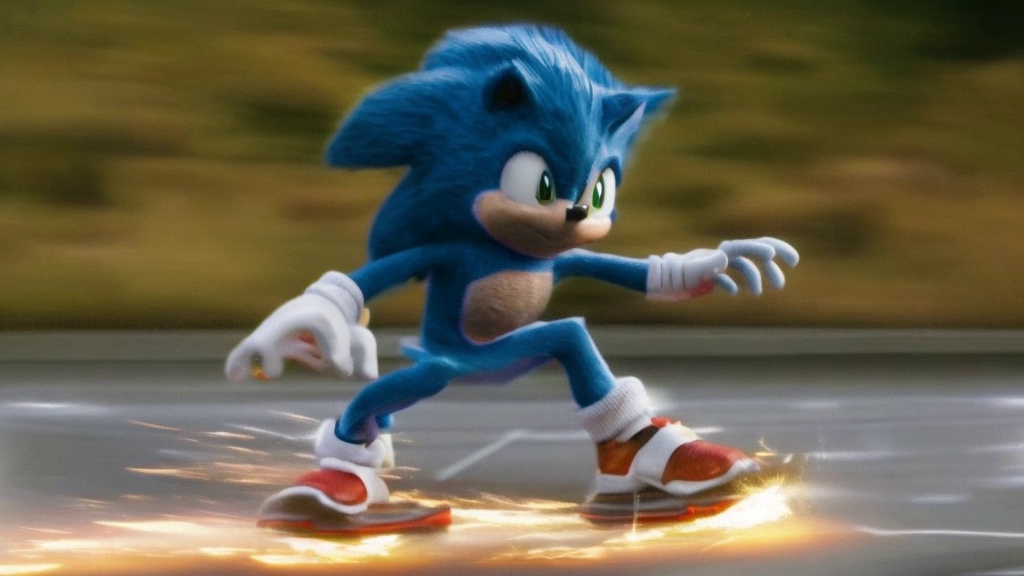}
            \put(83,23.5){\makebox[0pt]{\adjincludegraphics[height=0.3\textwidth,trim={{.50\width} {.65\height} {.32\width} {.02\height}},clip, cfbox=purple 1pt 0cm]{misc/dslv_sonic/frame_000__dlsv_none_r.jpg}}}
            \put(22,7.5){\makebox[0pt]{\adjincludegraphics[height=0.23\textwidth,trim={{.20\width} {.33\height} {.60\width} {.43\height}},clip, cfbox=purple 1pt 0cm]{misc/dslv_sonic/frame_000__dlsv_none_r.jpg}}}
        \end{overpic}
        \caption{\footnotesize Source Depth Map}
     \end{subfigure}
     \begin{subfigure}[b]{0.24\linewidth}
        \begin{overpic}[width=\textwidth]{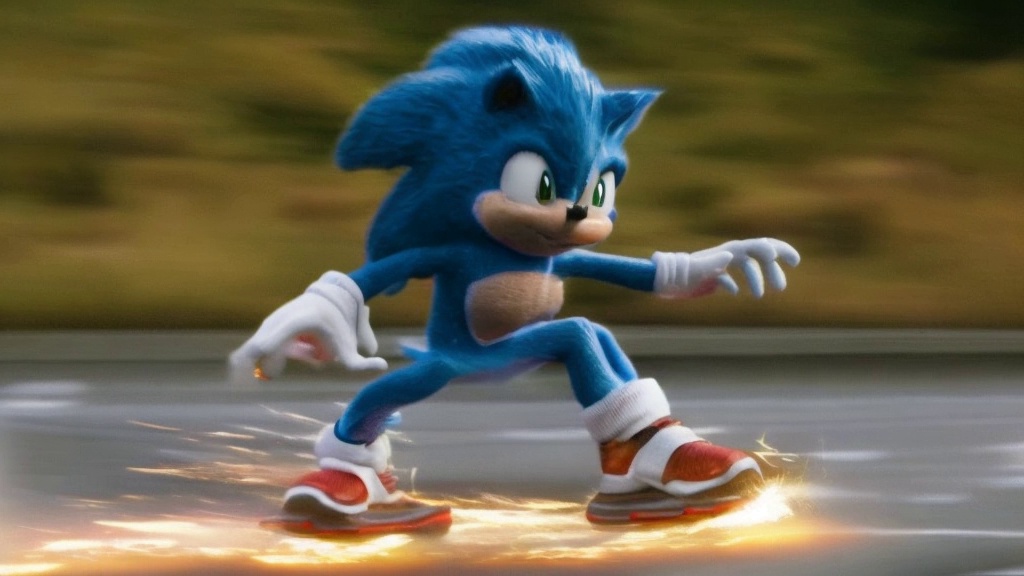}
            \put(83,23.5){\makebox[0pt]{\adjincludegraphics[height=0.3\textwidth,trim={{.50\width} {.65\height} {.32\width} {.02\height}},clip, cfbox=purple 1pt 0cm]{misc/dslv_sonic/frame_000__dlsv_20_r.jpg}}}
            \put(22,7.5){\makebox[0pt]{\adjincludegraphics[height=0.23\textwidth,trim={{.20\width} {.33\height} {.60\width} {.43\height}},clip,  cfbox=purple 1pt 0cm]{misc/dslv_sonic/frame_000__dlsv_20_r.jpg}}}
        \end{overpic}
        \caption{\footnotesize Dissolved Depth, $t=20$}
     \end{subfigure}
     \begin{subfigure}[b]{0.24\linewidth}
        \begin{overpic}[width=\textwidth]{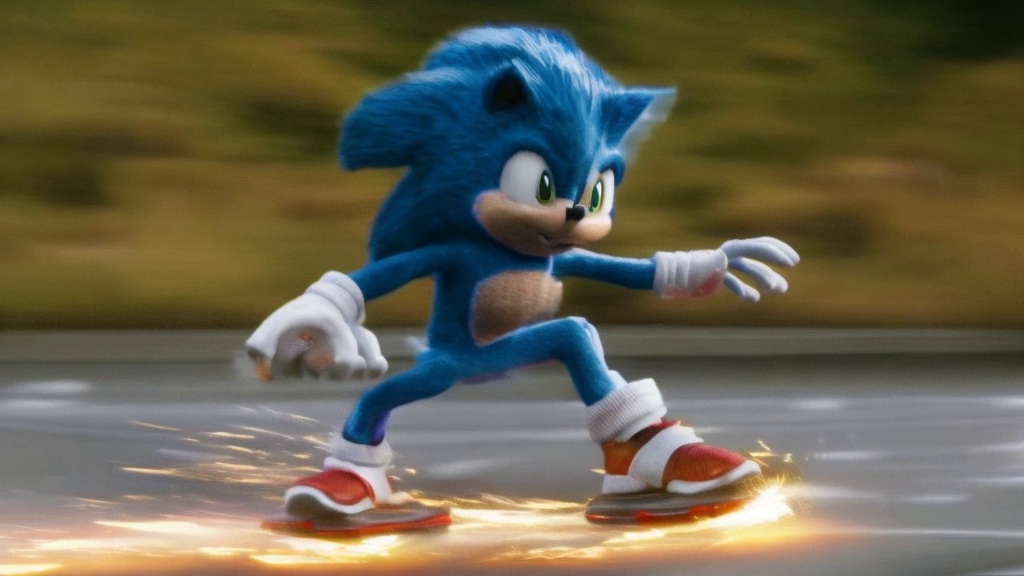}
            \put(83,23.5){\makebox[0pt]{\adjincludegraphics[height=0.3\textwidth,trim={{.50\width} {.65\height} {.32\width} {.02\height}},clip, cfbox=purple 1pt 0cm]{misc/dslv_sonic/frame_000__dlsv_30_r.jpg}}}
            \put(22,7.5){\makebox[0pt]{\adjincludegraphics[height=0.23\textwidth,trim={{.20\width} {.33\height} {.60\width} {.43\height}},clip, cfbox=purple 1pt 0cm]{misc/dslv_sonic/frame_000__dlsv_30_r.jpg}}}
        \end{overpic}
        \caption{\footnotesize Dissolved Depth, $t=30$}
     \end{subfigure}
     \begin{subfigure}[b]{0.24\linewidth}
        \begin{overpic}[width=\textwidth]{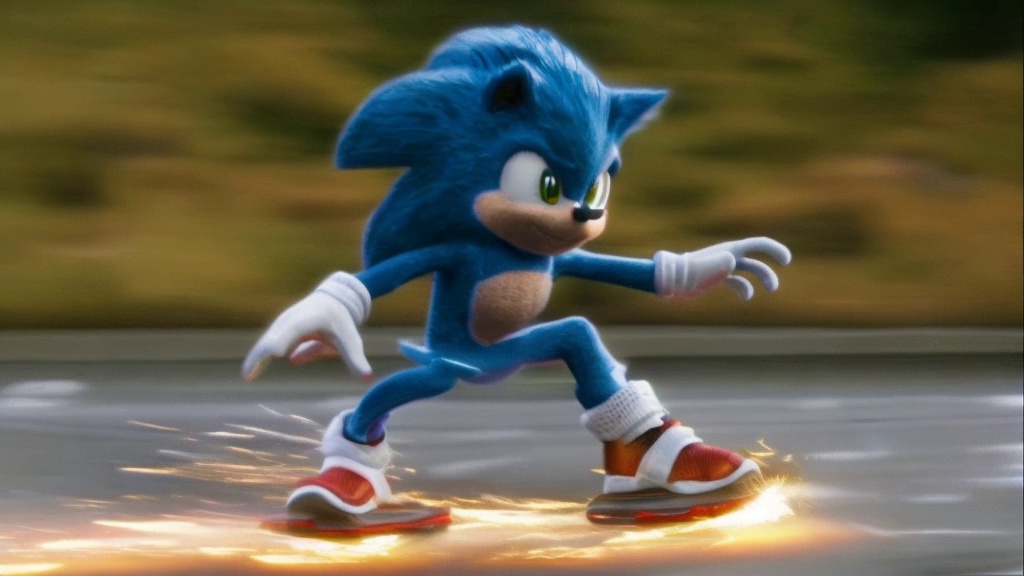}
            \put(83,23.5){\makebox[0pt]{\adjincludegraphics[height=0.3\textwidth,trim={{.50\width} {.65\height} {.32\width} {.02\height}},clip, cfbox=purple 1pt 0cm]{misc/dslv_sonic/frame_000__dlsv_40_r.jpg}}}
            \put(22,7.5){\makebox[0pt]{\adjincludegraphics[height=0.23\textwidth,trim={{.20\width} {.33\height} {.60\width} {.43\height}},clip, cfbox=purple 1pt 0cm]{misc/dslv_sonic/frame_000__dlsv_40_r.jpg}}}
        \end{overpic}
        \caption{\footnotesize Dissolved Depth, $t=40$}
     \end{subfigure}
     \Description{Dissolved depth maps from \textit{DepthCrafter} (50-step schedule). Top: progressive removal of high-frequency details. Bottom: reduced ghosting effects.}
     \caption{Dissolved depth maps from \textit{DepthCrafter} (50-step schedule). Top: progressive removal of high-frequency details. Bottom: reduced ghosting effects.}
     \label{fig:dslv_demo}
\end{figure*}

\subsection{Dissolved (Low-Frequency) Depth Maps}
\label{sec:dslv_depth}

Depth estimation models are typically designed to capture fine, detailed depth maps. However, unlike image-space warping, where high-frequency depth helps preserve pixel-level accuracy, latent-space warping prioritizes coarse geometry and semantic consistency. As demonstrated in~\Cref{fig:latent_space}, the diffusion latent spaces mostly emphasize the low-frequency structure information. Thus, when warping the latent space of a diffusion model, these high-frequency details can compromise the latent space consistency. As shown in~\Cref{sec:evaluation}, high-precision depth maps often introduce artifacts, such as ghosting, during the warping process. We suspect that this fine-grained warping, especially the sharp edges within depth maps, interferes with temporal coherence, making them less compatible with video diffusion priors.

\begin{figure}[h]
    \centering
     \begin{subfigure}[b]{0.49\linewidth}
        \begin{overpic}[width=\textwidth]{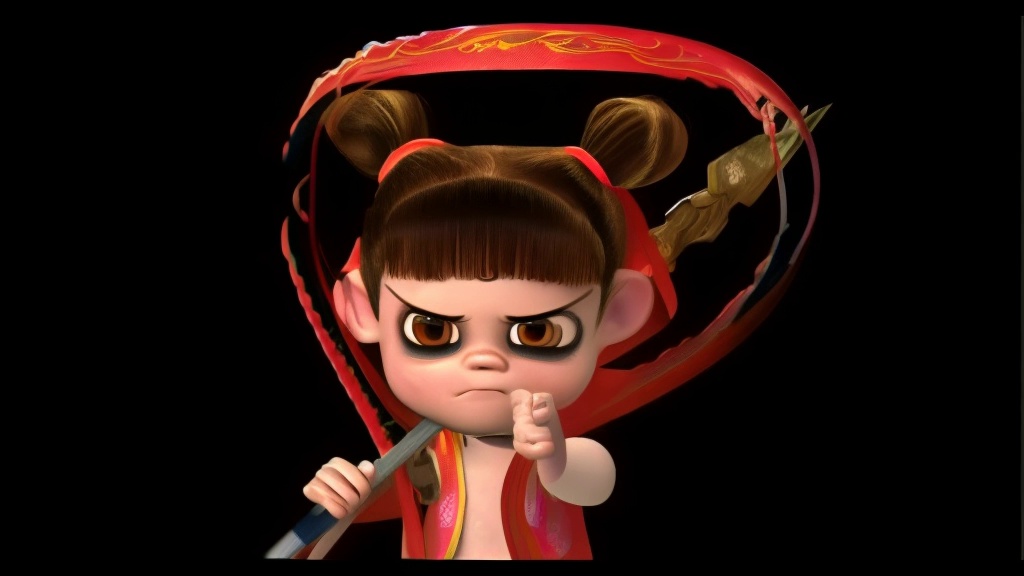}
            \put(90,4.5){\makebox[0pt]{\adjincludegraphics[height=0.5\textwidth,trim={{.74\width} {.55\height} {.19\width} {.05\height}},clip, cfbox=purple 2pt 0cm]{misc/dslv/a_boy_is_facing_the_front-r-d_none.jpg}}}
            \put(12,3.5){\makebox[0pt]{\adjincludegraphics[height=0.5\textwidth,trim={{.26\width} {.28\height} {.60\width} {.05\height}},clip, cfbox=purple 2pt 0cm]{misc/dslv/a_boy_is_facing_the_front-r-d_none.jpg}}}
        \end{overpic}
        \caption{w/o Dissolved Depth}
     \end{subfigure}
     \begin{subfigure}[b]{0.49\linewidth}
        \begin{overpic}[width=\textwidth]{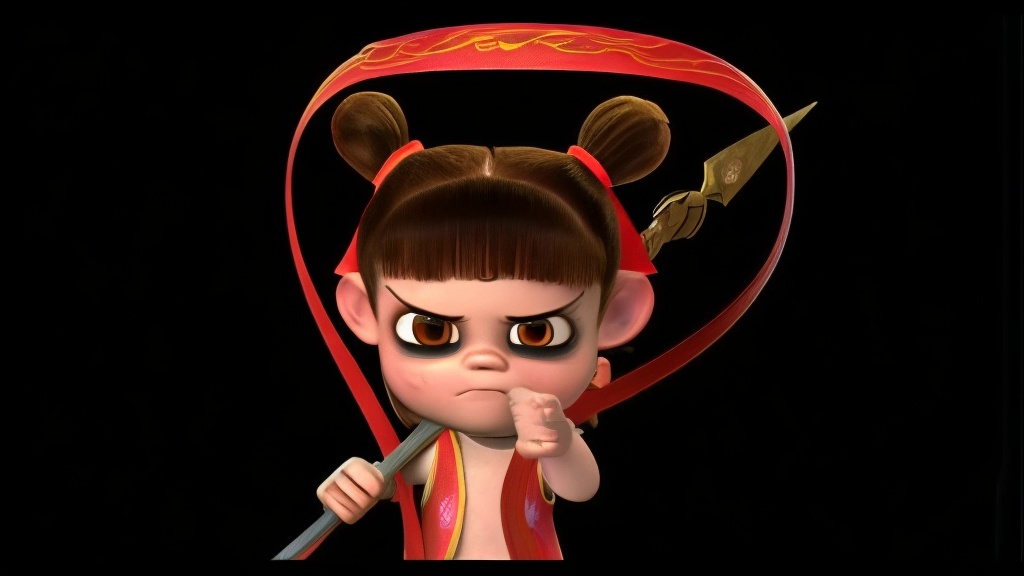}
            \put(90,4.5){\makebox[0pt]{\adjincludegraphics[height=0.5\textwidth,trim={{.74\width} {.55\height} {.19\width} {.05\height}},clip, cfbox=purple 2pt 0cm]{misc/dslv/a_boy_is_facing_the_front-r-d4.jpg}}}
            \put(12,3.5){\makebox[0pt]{\adjincludegraphics[height=0.5\textwidth,trim={{.26\width} {.28\height} {.60\width} {.05\height}},clip, cfbox=purple 2pt 0cm]{misc/dslv/a_boy_is_facing_the_front-r-d4.jpg}}}
        \end{overpic}
        \caption{w/ Dissolved Depth}
     \end{subfigure}
    \captionof{figure}{Dissolved depth effectively reduces the artifacts.}
    \label{fig:dslv_effects}
    \Description{Visualizing the effects with and without dissolved depth.}
\end{figure}

To address this issue, we propose a depth-dissolving technique that transforms the depth maps into a lower-frequency representation. Inspired by the semantic simplification technique proposed by \textit{Dissolving Is Amplifying (DIA)}~\cite{shi2024dissolving} and Wang \textit{et al.}~\cite{wang2024layered}, we generate \textbf{dissolved depth maps} by leveraging the inherent properties of diffusion models to act as a low-pass filter on the latent space.
Using a diffusion-based depth estimation model, we first obtain a depth latent $x_T$ at the final diffusion step $T$. Instead of performing a full reverse diffusion process, we only execute a single-step reverse diffusion on $x_T$. This approximation is designed to suppress high-frequency details, thereby reducing noise and artifacts in the latent representation.
Formally, we denote the approximated initial state as $\hat{x}_{t\rightarrow 0}$, which depends on the selected time step $t$. The process is defined as follows:
\begin{align}
    \hat{x}_{t\rightarrow 0}=\sqrt{\frac{1}{\Bar{\alpha}_t}} \cdot x - \sqrt{\frac{1}{\Bar{\alpha}_t} - 1}\cdot \epsilon_\theta(x,t),
\end{align}
where $\Bar{\alpha}_t$ represents the cumulative product of the diffusion coefficients up to time $t$, $\epsilon_\theta$ is the predicted noise at time $t$.
By emphasizing global structure over pixel-level depth variations, our approach enables smoother disparity transitions, as shown in~\Cref{fig:dslv_demo}. As a result, the warped latents maintain better temporal coherence and align more effectively with the video diffusion priors.
Experimental results confirm our hypothesis that dissolved depth maps reduce artifacts such as ghosting and staircase effects.
We present a representative case in~\Cref{fig:dslv_effects}, where ghosting and jaggies artifacts are significantly reduced.

\begin{figure}[b]
    \centering
    \includegraphics[trim={0 0 0 .8cm}, clip, width=.95\linewidth]{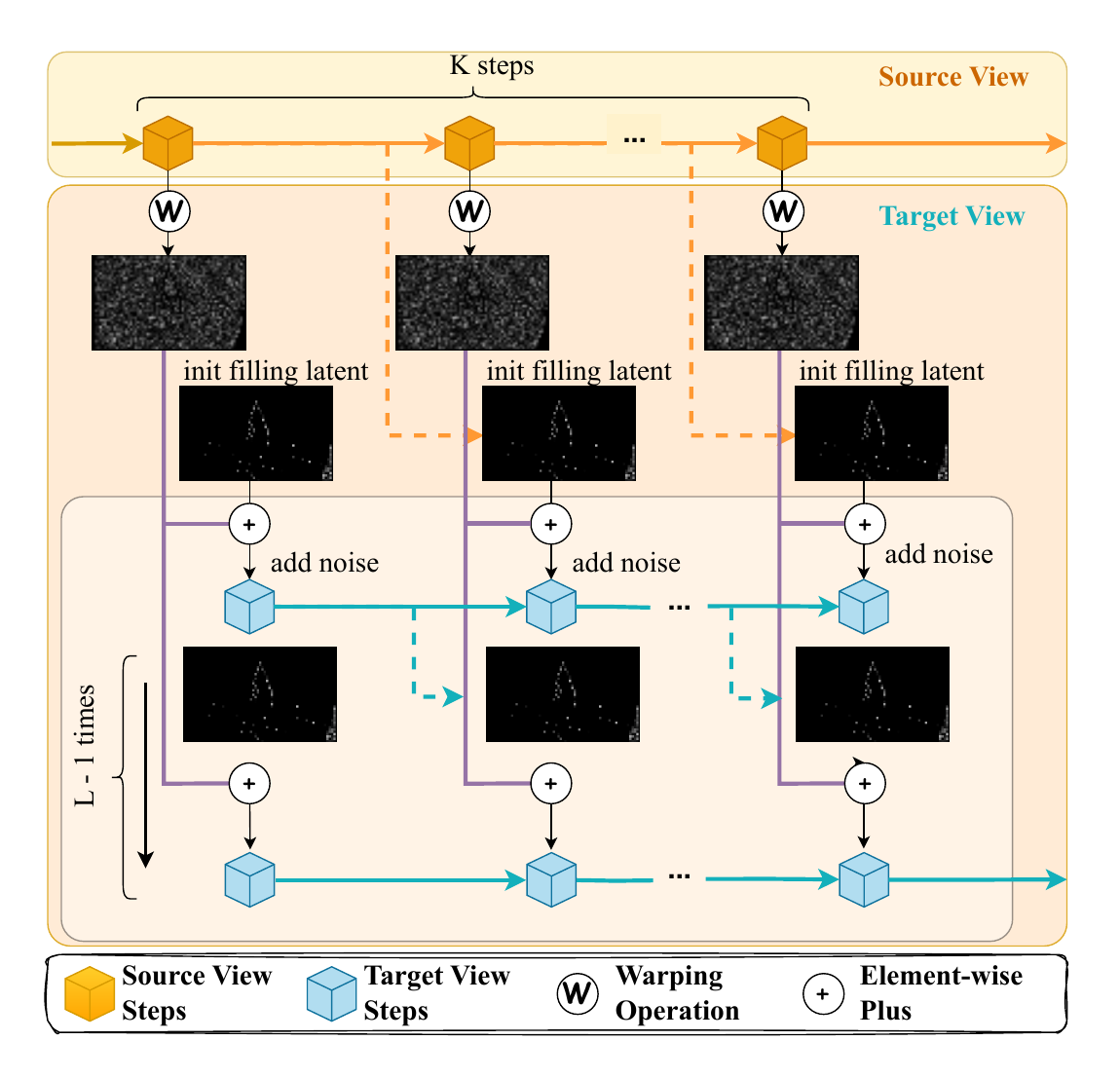}
    \Description{Illustration of noisy start. At selected steps, we replace target view sampling with a warped source view. Occluded/disoccluded areas are filled using the non-warped source view latent with added noise injected. Subsequent iterations update the latents with values from the preceding iteration while preserving non-occluded regions.}
    \caption{Illustration of noisy start. At selected steps, we replace target view sampling with a warped source view. Occluded/disoccluded areas are filled using the non-warped source view latent with added noise injected. Subsequent iterations update the latents with values from the preceding iteration while preserving non-occluded regions.}
    \label{fig:noisy_restart_fig}
\end{figure}
\begin{figure*}
     \centering
     \begin{subfigure}[b]{0.32\linewidth}
         \centering
         \includegraphics[width=\linewidth]{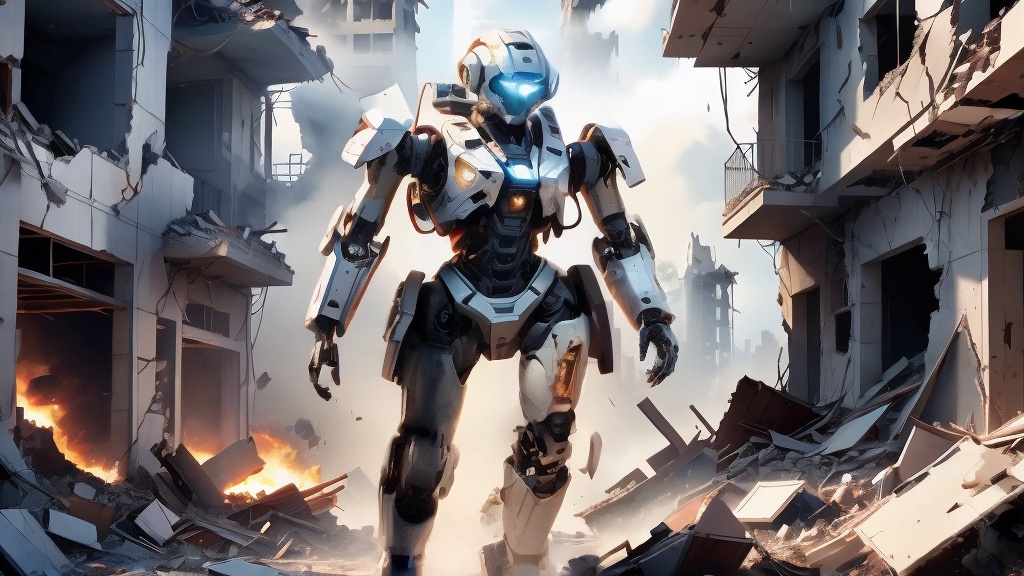}
         \caption{w/o Noisy Restart ($L=1$)}
     \end{subfigure}
     \begin{subfigure}[b]{0.32\linewidth}
         \centering
         \includegraphics[width=\linewidth]{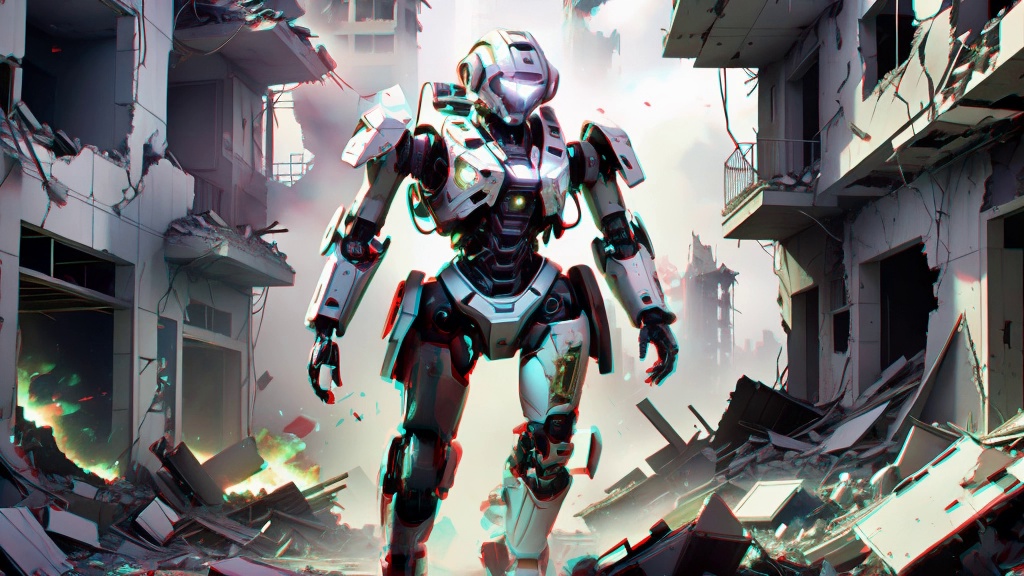}
         \caption{w/ Noisy Restart ($L=3$)}
     \end{subfigure}
     \begin{subfigure}[b]{0.32\linewidth}
         \centering
         \includegraphics[width=\linewidth]{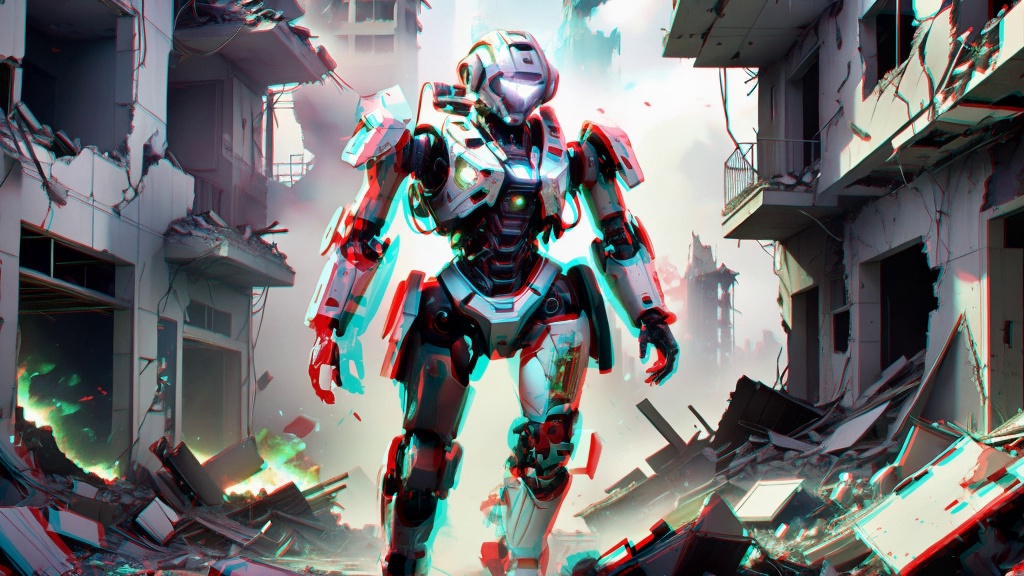}
         \caption{w/ Noisy Restart ($L=7$)}
     \end{subfigure}
    \Description{Impact of the Noisy Restart. The anaglyph shows stronger stereo effects with increasing restart iterations. Results use noisy restart only.}
    \caption{Impact of the Noisy Restart. The anaglyph shows stronger stereo effects with increasing restart iterations. Results use noisy restart only.}
    \label{fig:noisy_restart}
\end{figure*}
\subsection{Noisy Restart}
\label{sec:noisy_restart}

The noisy restart mechanism operates over $K$ diffusion steps for $L$ iterations. We denote the $K$ diffusion steps as $\{ x_t \}_{t=0}^{K}$, where $x_t$ represents the latent state at step $t$. While $K$ can be a set of discrete timesteps, we use $K=\{49,...,45\}$ in our implementation.
The main purpose of the noisy restart is to introduce random noise into the latent space to prevent structural repetition and ensure smooth transitions. In the first iteration, $K$ filling latents are generated for each diffusion step, while the subsequent iterations will refine the obtained $K$ latents.
Referring to~\Cref{eq:ddim}, a random noise $\epsilon_t$ is introduced at each sampling step. If each iteration uses different noise values, consistency across iterations can be disrupted. To address this issue, we initialize fixed noise tensors $\epsilon{'}_t$ before starting the diffusion sampling sequence and use the same random seed for each iteration. This operation ensures that the random generator status remains the same across multiple iterations, and the noise injection is controlled solely by $\epsilon'_t$ and $\alpha_t$ in~\Cref{eq:noisy_restart}. Here $\alpha_t$ is a weighting factor that regulates the relative contributions of the previous latent state and the injected noise.
Meanwhile, such a controlled approach promotes structural stability throughout the generation process, ensuring consistent low-frequency noise that preserves major features across frames.
We used $L = 7$ iterations, with the first iteration directly using the left-view latents as the filling latents. Noise is injected into the latent state through a weighted addition, balancing the contributions of the existing latent and the injected noise. We update $x_{t-1}$ by:
\begin{equation}
x_{t-1} = x_{t} \cdot (1 - \alpha_t) + \sigma_t \epsilon{'}_t \alpha_t
\label{eq:noisy_restart}
\end{equation}
where $\sigma_t$ is the noise magnitude at timestep $t$, scaling the impact of injected noise, while $\alpha_t$ is the balancing coefficient to control the mixture between the latent state and the noise. This approach significantly enhances stereoscopic effects, as shown in~\Cref{fig:noisy_restart}.

\begin{figure}[h]
     \centering
     \begin{subfigure}[b]{0.49\linewidth}
         \centering
         \includegraphics[width=\linewidth]{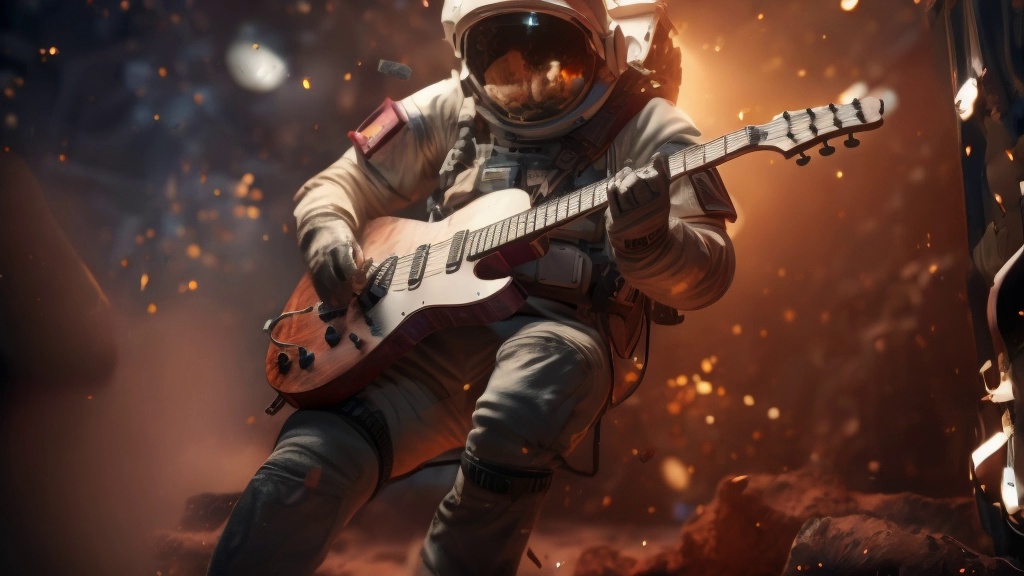}
     \end{subfigure}
     \begin{subfigure}[b]{0.49\linewidth}
         \centering
         \includegraphics[width=\linewidth]{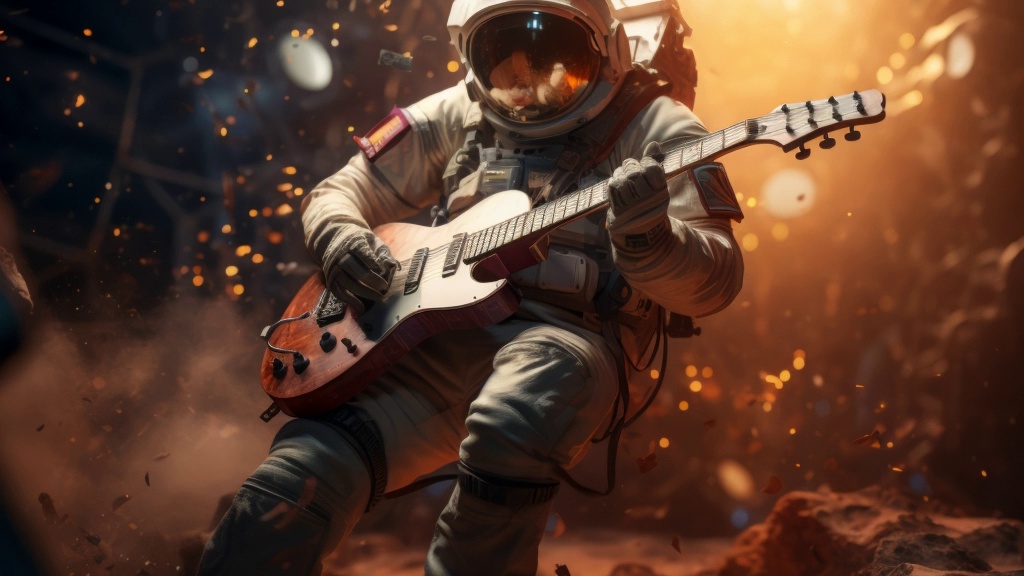}
     \end{subfigure}
     \\
     \begin{subfigure}[b]{0.49\linewidth}
         \centering
         \includegraphics[width=\linewidth]{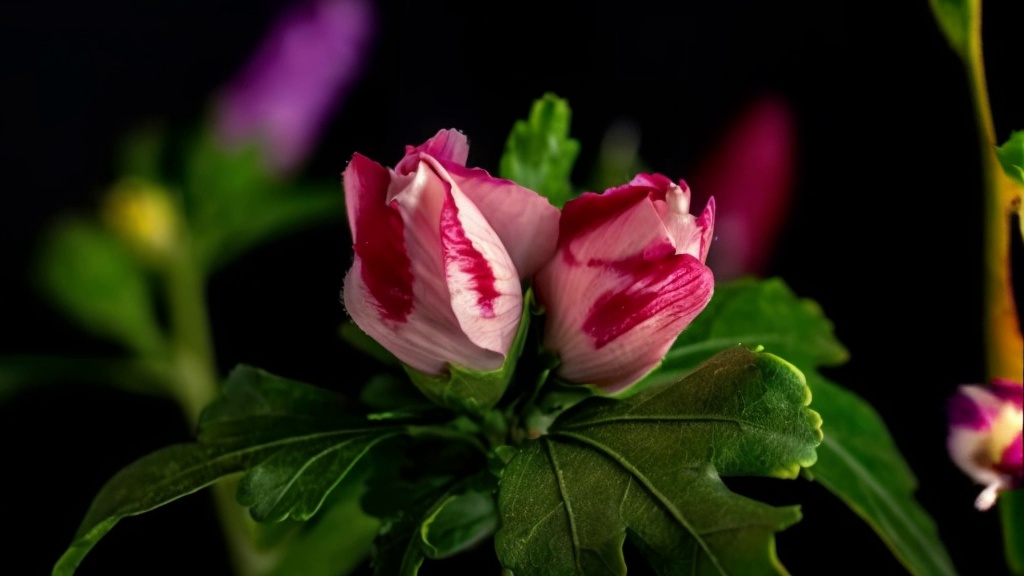}
         \caption{w/o border handling}
     \end{subfigure}
     \begin{subfigure}[b]{0.49\linewidth}
         \centering
         \includegraphics[width=\linewidth]{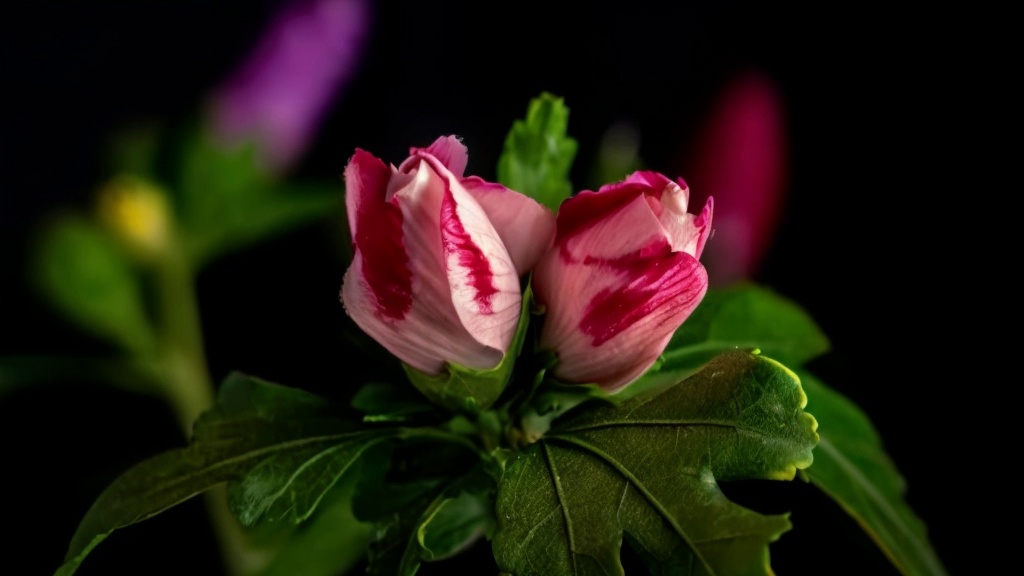}
         \caption{w/ border handling}
     \end{subfigure}
    \Description{Abrupt border handling. (a) Images with noticeable abrupt artifacts along the right edge. (b) Border artifacts are effectively removed.}
    \caption{Abrupt border handling. (a) Images with noticeable abrupt artifacts along the right edge. (b) Border artifacts are effectively removed.}
\label{fig:bad_border}
\end{figure}
\paragraph{Abrupt Border Handling.}
Direct warping of the right view can result in blank regions along the right border. These blank areas may create irrelevant or distracting content in the border regions during the iterative process. In order to maximally preserve the widest possible stereoscopic field of view, we introduce a border-cleaning function that selectively masks and fills these regions using information from the left view, ensuring visual coherence.
A border mask, ${M}_{\text{border}}$, is created using a heuristic function to select the border areas. We then inpaint these areas with the corresponding regions in the left-view latents, denoted as ${L}_l$. The resulting border refined latent ${L}_{r,\text{refined}}$ is defined as:
\begin{equation}
{L}_{r,\text{refined}} = {L}_r\; (1 - {M}_{\text{border}}) + {L}_l\; {M}_{\text{border}}.
\end{equation}
This operation replaces the blank columns in the right view with the content from the left view, ensuring visual continuity and eliminating the reasonable but distracting artifacts of the stereo pair.
A qualitative evaluation of this technique is presented in~\Cref{fig:bad_border}.

\subsection{Iterative Refinement}
\label{sec:iter_refine}

\begin{figure}[h]
    \centering
     \begin{subfigure}[b]{0.49\linewidth}
         \centering
         \includegraphics[width=\linewidth]{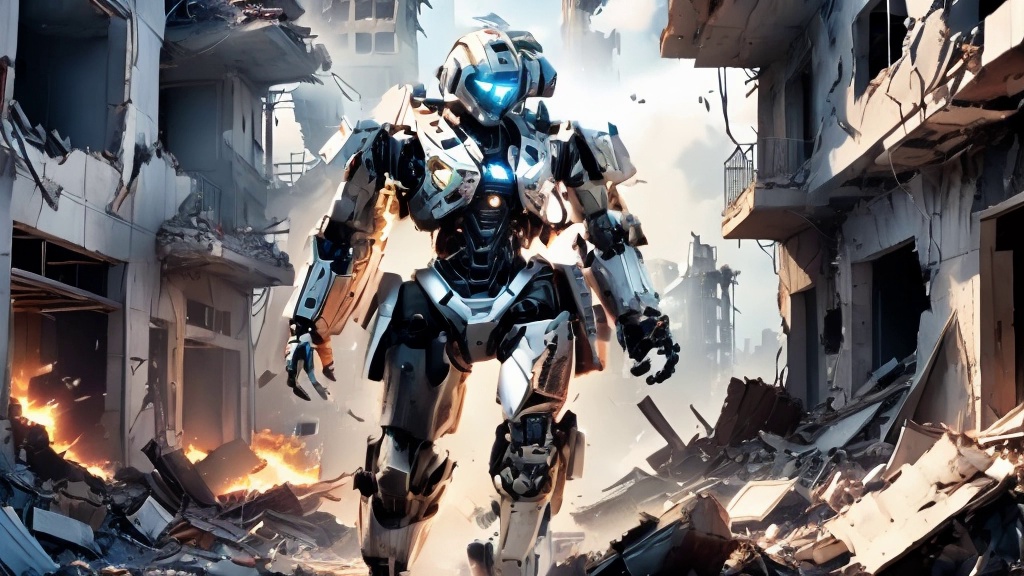}
         \caption{w/o Iterative Refine}
     \end{subfigure}%
     \hspace{.1em}
     \begin{subfigure}[b]{0.49\linewidth}
         \centering
         \includegraphics[width=\linewidth]{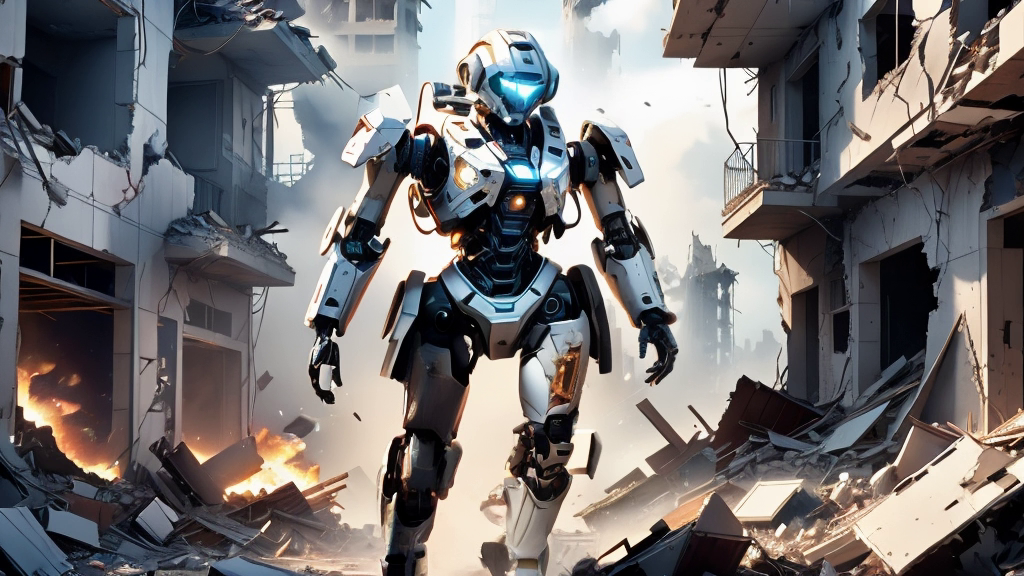}
         \caption{w/ Iterative Refine}
     \end{subfigure}
    \Description{Impact of Iterative Refinement. Without it, warping artifacts may degrade the filling areas.}
    \caption{Impact of Iterative Refinement. Without it, warping artifacts may degrade the filling areas.}
    \label{fig:iter_refine_performance}
\end{figure}

To optimize computational resources, we limit noisy restart to initial sampling steps, where it has the highest impact on stereo effects. For later stages, we introduce Iterative Refinement, which leverages video diffusion priors to enhance details in occluded regions. It operates at specific diffusion steps, repeating the denoising operation $N$ times on the occluded regions. For each refinement step, we first obtain a predicted latent $x_t^{j=1}$ with the UNet denoiser $\epsilon_\theta$. For subsequent iterations ($j > 1$), the latent is updated as follows:
\begin{equation}
{x}_t^{ j} = (1-M)\; {x}_t^{ (j-1)} + M \;\, \epsilon_\theta\!\left({x}_t^{ (j-1)}\right),
\label{eq:blank_areas_itrefine}
\end{equation}
where M is the mask for occluded regions.
This approach optimizes computational resource usage for efficient refinement while delivering substantial quality improvements, as can be seen in \Cref{fig:iter_refine_performance}.

\subsection{Pipeline Summary}

We summarize the overall \textit{DissolveStereo} pipeline in \Cref{alg:pipeline}.
Given a conditioning signal $c$ (image and text prompt), we first generate the left-view video using DDIM sampling. To synthesize the stereo pair, we estimate depth maps from the decoded video and convert them into dissolved depth maps.
The stereo generation process warps the left-view latents based on depth-induced disparity (derived from the dissolved depth maps and a desired baseline distance), yielding partially observed latents with occluded regions. 
These regions are initialized and refined through \textit{noisy restart} at early diffusion steps and \textit{iterative refinement} at later steps, while preserving non-occluded regions. 
We denote a standard DDIM sampling step with cross-view attention (\Cref{sec:cross_view_attention}) as \texttt{StereoStep}.

\begin{algorithm}
\caption{DissolveStereo}
\label{alg:pipeline}
\begin{algorithmic}[1]
\Require Condition $c$, baseline distance $s$,\;
       $T{=}50$,\; $L{=}7$,\; $K{=}\{T,\ldots,T{-}5\}$
\Ensure  Left video $\mathbf{v}_L$,\quad Right video $\mathbf{v}_R$
\State $\{\mathbf{z}_L^{(t)}\}_{t=T}^{1}, \mathbf{v}_L
     \gets \mathrm{DDIMSample}(c)$
\Comment{save left-view trajectory}
\State $\mathbf{d} \gets \mathrm{DissolvedDepth}(\mathbf{v}_L)$

\Statex \textbf{Phase 1 --- Noisy Restart}
\State $\mathbf{f}_t \gets \mathbf{z}_L$, $\forall\, t \!\in\! K$
\For{$j = 1$ \textbf{to} $L$}
\State $[\mathbf{z}_L, \mathbf{z}_R] \sim \mathcal{N}(0, I)$
  \Comment{fresh noise every iteration}
\For{$t \in K$ \textbf{(descending)}}
  \State $\hat{\mathbf{z}}_R,\,\mathbf{m}
         \gets \mathrm{Warp}(\mathbf{z}_L, \mathbf{d}, s)$;\quad
         $\hat{\mathbf{z}}_R[\mathbf{m}] \gets \mathbf{f}_t[\mathbf{m}]$
  \State $[\mathbf{z}_L,\,\mathbf{z}_R]
         \gets \mathrm{StereoStep}([\mathbf{z}_L,\,\hat{\mathbf{z}}_R],\,t)$
  \State $\mathbf{f}_t \gets \mathbf{z}_R + \delta\boldsymbol{\varepsilon}$
    \Comment{update per-step fill for next iteration}
\EndFor
\EndFor
\Statex \textbf{Phase 2 --- Iterative Refinement}
\State $[\mathbf{z}_L,\,\mathbf{z}_R] \gets$ Phase~1 output at $t{=}T{-}5$
\For{$t = T{-}6$ \textbf{downto} $1$}
\If{$t \in \mathrm{warp\_steps}$}
  \State $\hat{\mathbf{z}}_R,\,\mathbf{m}
         \gets \mathrm{Warp}(\mathbf{z}_L, \mathbf{d}, s)$;\quad
         $\hat{\mathbf{z}}_R[\mathbf{m}] \gets \mathbf{z}_R[\mathbf{m}]$
  \For{$k = 1$ \textbf{to} $N(t)$}
    \State $[\mathbf{z}_L,\,\mathbf{z}_R]
           \gets \mathrm{StereoStep}([\mathbf{z}_L,\,\hat{\mathbf{z}}_R],\,t)$
    \State $\hat{\mathbf{z}}_R[\lnot\mathbf{m}]
           \gets \mathbf{z}_R[\lnot\mathbf{m}]$
  \EndFor
\Else
  \State $[\mathbf{z}_L,\,\mathbf{z}_R]
         \gets \mathrm{StereoStep}([\mathbf{z}_L,\,\mathbf{z}_R],\,t)$
\EndIf
\State $\mathbf{z}_L \gets \mathbf{z}_L^{(t)}$
  \Comment{anchor left to pre-computed trajectory}
\EndFor
\State \Return $\mathrm{Decode}(\mathbf{z}_L)$, $\mathrm{Decode}(\mathbf{z}_R)$
\end{algorithmic}
\end{algorithm}

\section{Experiments}
\label{sec:exp}

\subsection{Implementation details}
\label{sec:exp_setup}

We implement our method based on \textit{DynamiCrafter}~\cite{xing2024dynamicrafter}. We also evaluated our method with other diffusion-based video generation methods, such as~\cite{he2022lvdm,zhang2025motion} in our supplementary material. We infer video depth maps using \textit{DepthCrafter}~\cite{hu2024-DepthCrafter}. We apply the cross-view attention mechanism on all sampling steps apart from the cross-attention between the latents and the conditions. We use $L=7$ for noisy restart from $t=49$ to $t=45$. Afterward, warping is performed to update the side-view latents every 5 steps, followed by iterative refinement at each warping stage. We found that the last iterative refinement step of $t=15$ is sufficient for most cases. The number of refinements is set to $N=4$.

\subsection{Results}
\label{sec:result}

\begin{figure*}[t]

\setlength{\tabcolsep}{0pt}
\renewcommand{\arraystretch}{0}
\begin{tabular}{cccc}
    Source Frames & ProPainter & StereoDiffusion & RoDynRF \\
    ~
    \\
    \begin{overpic}[width=0.24\linewidth,height=0.135\linewidth]{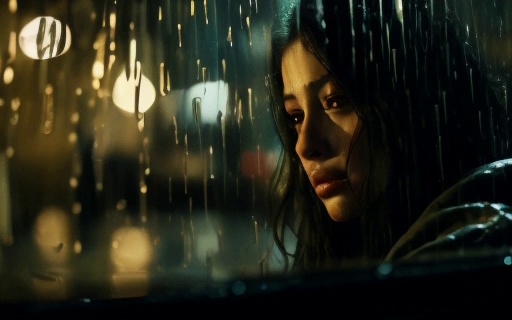}
        \put(32,20.5){\makebox[0pt]{\adjincludegraphics[height=.25\height,trim={{.2\width} {.5\height} {.5\width} 0},clip, cfbox=purple 2pt 0cm]{misc/main_figure/inputs/a_woman_looking_out_in_the_rain_input_000__out.jpg}}}
    \end{overpic}
        &
    \begin{overpic}[width=0.24\linewidth,height=0.135\linewidth]{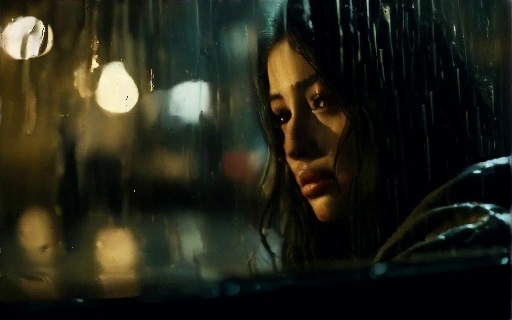}
        \put(32,20.5){\makebox[0pt]{\adjincludegraphics[height=.25\height,trim={{.2\width} {.5\height} {.5\width} 0},clip, cfbox=purple 2pt 0cm]{misc/main_figure/propainter/a_woman_looking_out_in_the_rain_masked_000.jpg}}}
    \end{overpic}
        &
    \begin{overpic}[width=0.24\linewidth,height=0.135\linewidth,trim={1024px 0 0 0},clip]{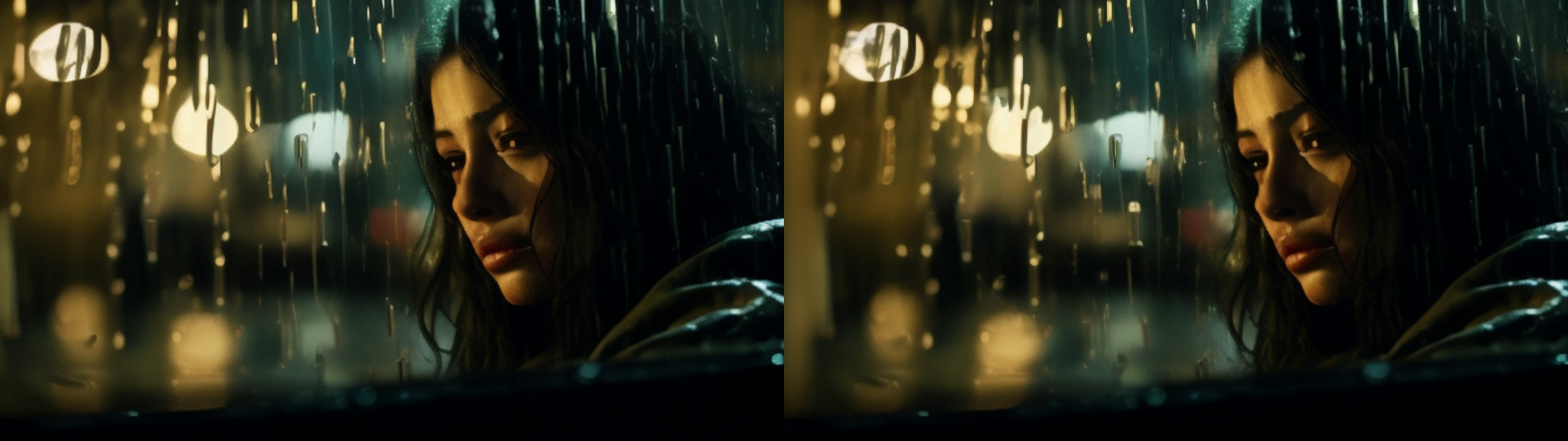}
        \put(32,20.5){\makebox[0pt]{\adjincludegraphics[height=.085\height,width=.125\width,trim={{.6\width} {.2\height} {.25\width} 0},clip, cfbox=purple 2pt 0cm]{misc/main_figure/stereodiffusion/a_woman_looking_out_in_the_rain-r_right_000.png}}}
    \end{overpic}
        &
    \begin{overpic}[width=0.24\linewidth,height=0.135\linewidth]{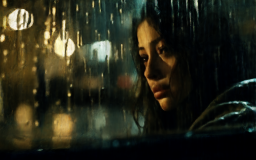}
        \put(32,20.5){\makebox[0pt]{\adjincludegraphics[height=.48\height,trim={{.2\width} {.5\height} {.5\width} 0},clip, cfbox=purple 2pt 0cm]{misc/main_figure/rodynrf/woman_robustnrf.png}}}
    \end{overpic}
    \\
    \begin{overpic}[width=0.24\linewidth,height=0.135\linewidth]{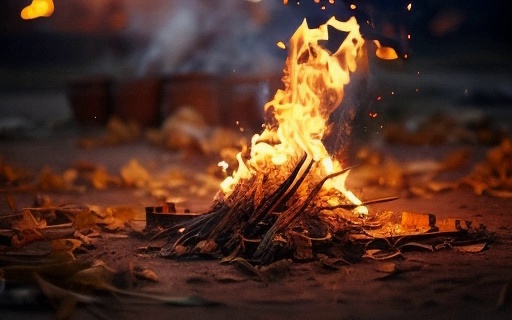}
        \put(78,16.5){\makebox[0pt]{\adjincludegraphics[height=.28\height,trim={{.53\width} {.5\height} {.2\width} 0},clip, cfbox=purple 2pt 0cm]{misc/main_figure/inputs/a_bonfire_is_lit_in_the_middle_of_a_fiel_input_000__out.jpg}}}
    \end{overpic}
        &
    \begin{overpic}[width=0.24\linewidth,height=0.135\linewidth]{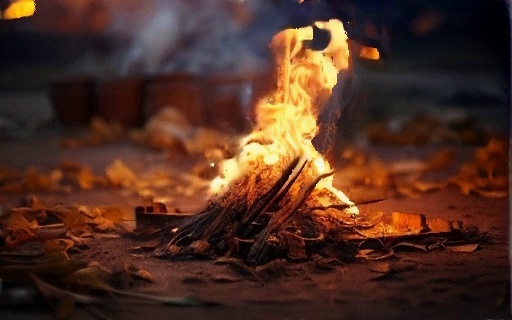}
        \put(78,16.5){\makebox[0pt]{\adjincludegraphics[height=.28\height,trim={{.53\width} {.5\height} {.2\width} 0},clip, cfbox=purple 2pt 0cm]{misc/main_figure/propainter/a_bonfire_is_lit_in_the_middle_of_a_fiel_masked_000.jpg}}}
    \end{overpic}
        &
    \begin{overpic}[width=0.24\linewidth,height=0.135\linewidth,trim={1024px 0 0 0},clip]{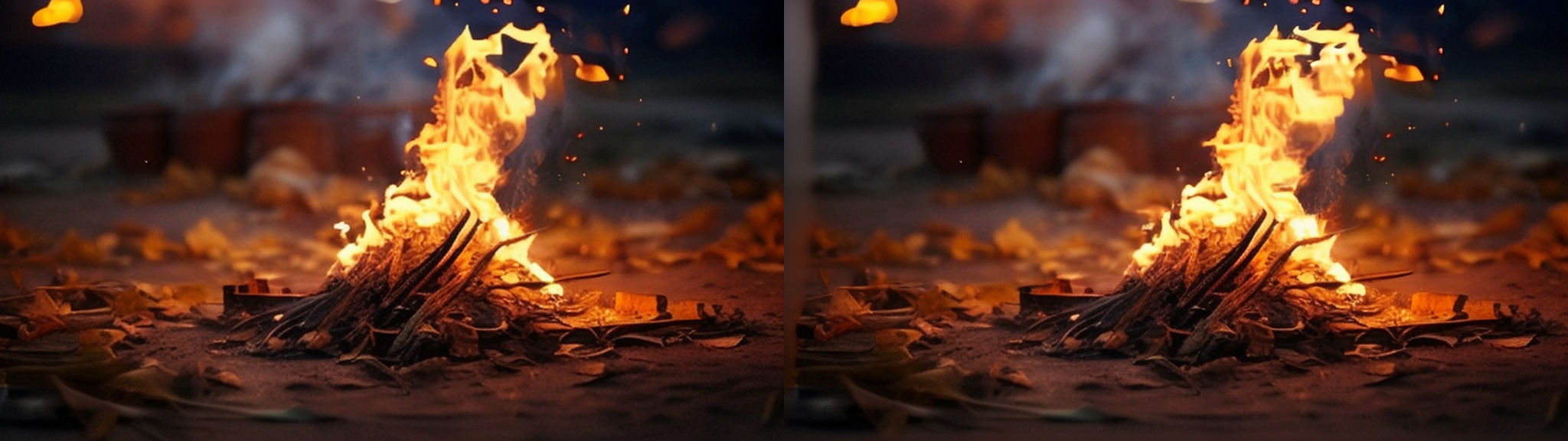}
        \put(78,16.5){\makebox[0pt]{\adjincludegraphics[height=.155\height,trim={{.78\width} {.5\height} {.1\width} 0},clip, cfbox=purple 2pt 0cm]{misc/main_figure/stereodiffusion/a_bonfire_is_lit_in_the_middle_of_a_fiel-r_right_000.png}}}
    \end{overpic}
        &
    \begin{overpic}[width=0.24\linewidth,height=0.135\linewidth]{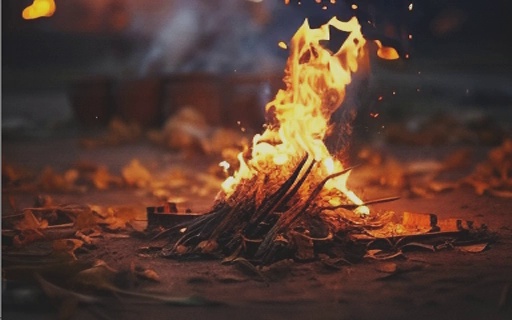}
        \put(78,16.5){\makebox[0pt]{\adjincludegraphics[height=.11\linewidth,trim={{.53\width} {.5\height} {.2\width} 0},clip, cfbox=purple 2pt 0cm]{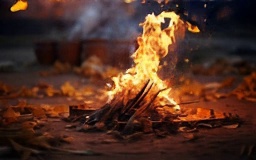}}}
    \end{overpic}
    \\
    \begin{overpic}[width=0.24\linewidth,height=0.135\linewidth]{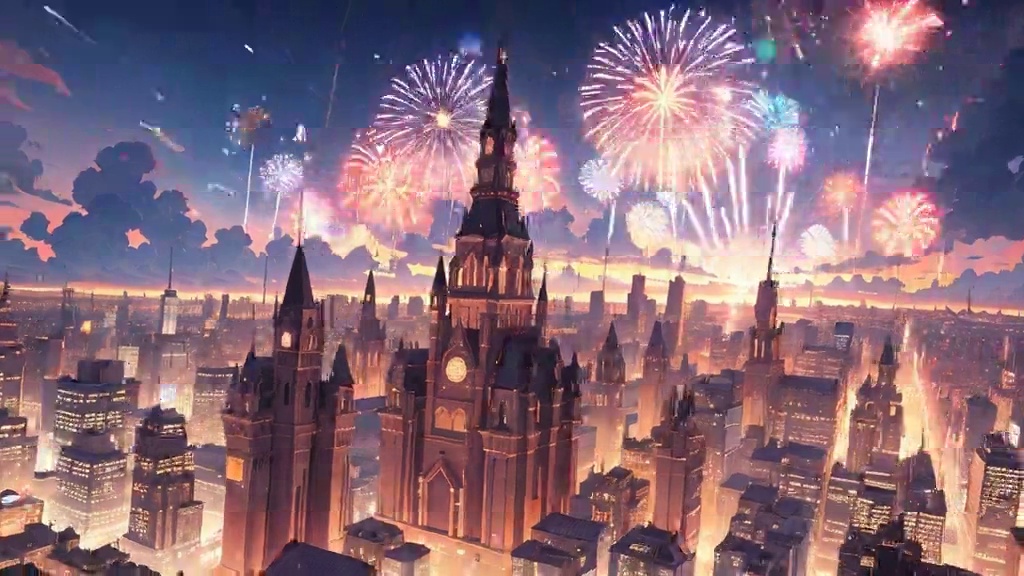}
        \put(57,20.5){\makebox[0pt]{\adjincludegraphics[height=0.1\linewidth,trim={{.3\width} {.5\height} {.2\width} 0},clip, cfbox=purple 2pt 0cm]{misc/main_figure/inputs/fireworks_display_input_015__out.jpg}}}
    \end{overpic}
        &
    \begin{overpic}[width=0.24\linewidth,height=0.135\linewidth]{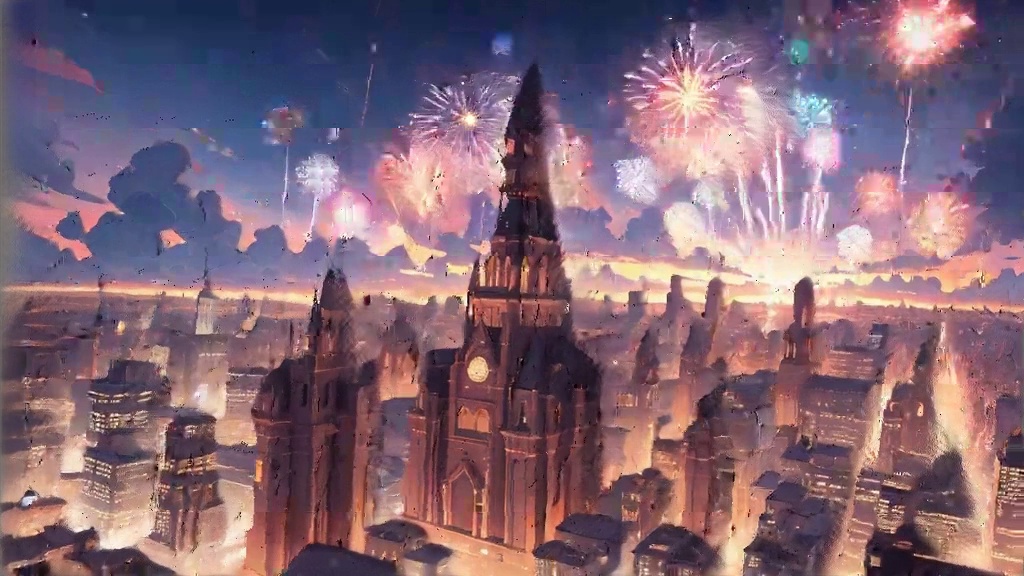}
        \put(57,20.5){\makebox[0pt]{\adjincludegraphics[height=0.1\linewidth,trim={{.3\width} {.5\height} {.2\width} 0},clip, cfbox=purple 2pt 0cm]{misc/main_figure/propainter/fireworks_display_masked_015.jpg}}}
    \end{overpic}
        &
     \begin{overpic}[width=0.24\linewidth,height=0.135\linewidth,trim={1024px 0 0 0},clip]{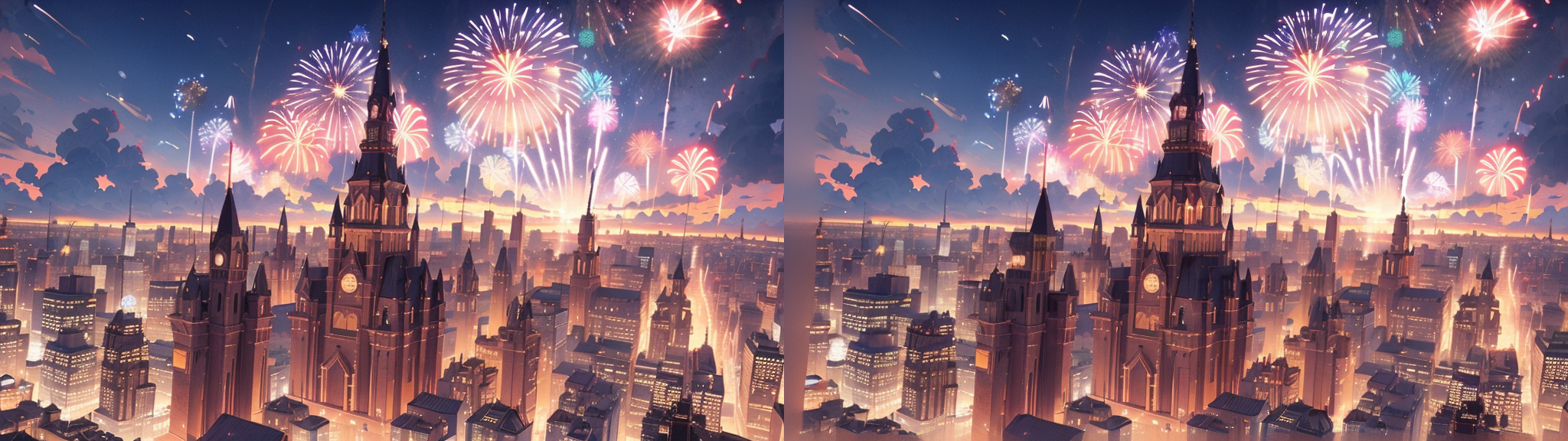}
        \put(57,20.5){\makebox[0pt]{\adjincludegraphics[height=0.1\linewidth,width=0.181\linewidth,trim={{.65\width} {.6\height} {.1\width} 0},clip, cfbox=purple 2pt 0cm]{misc/main_figure/stereodiffusion/fireworks_display-r_right_015.png}}}
    \end{overpic}
        &  
    \begin{overpic}[width=0.24\linewidth,height=0.135\linewidth]{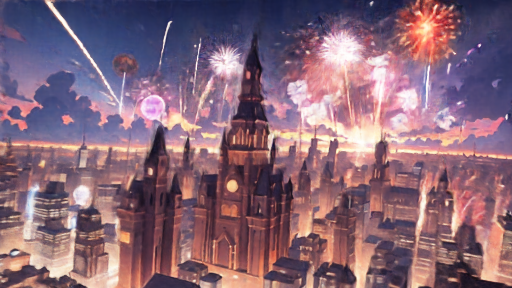}
        \put(57,20.5){\makebox[0pt]{\adjincludegraphics[height=0.1\linewidth,trim={{.3\width} {.5\height} {.2\width} 0},clip, cfbox=purple 2pt 0cm]{misc/main_figure/rodynrf/fireworks_robustnrf.png}}}
    \end{overpic}
    \\
    \begin{tabular}{cc}
         & ~ \\
    \end{tabular}
    \\
    ImmersePro & Traj. Crafter & StereoCrafter & Ours \\
    \\
    \begin{overpic}[width=0.24\linewidth,height=0.135\linewidth]{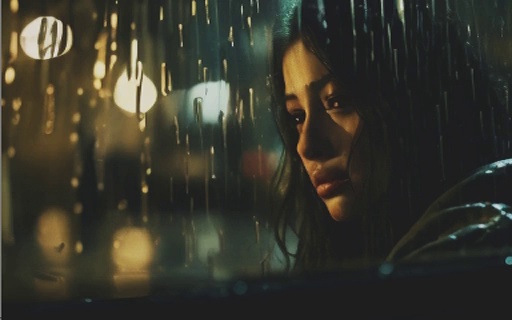}
        \put(32,20.5){\makebox[0pt]{\adjincludegraphics[height=.25\height,trim={{.2\width} {.5\height} {.5\width} 0},clip, cfbox=purple 2pt 0cm]{misc/main_figure/immersepro/a_woman_looking_out_in_the_rain_input_immersepro_000__out.jpg}}}
    \end{overpic}
        &
    \begin{overpic}[width=0.24\linewidth,height=0.135\linewidth]{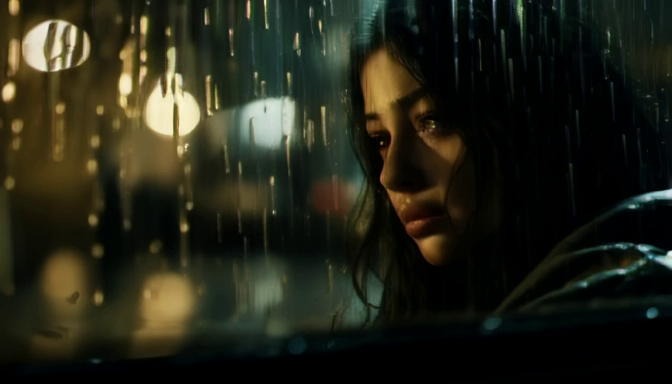}
        \put(35,20.5){\makebox[0pt]{\adjincludegraphics[height=.21\height,trim={{.2\width} {.52\height} {.54\width} 0},clip, cfbox=purple 2pt 0cm]{misc/main_figure/trajcrafter/a_woman_looking_out_in_the_rain_input_000.png}}}
    \end{overpic}
        &
    \begin{overpic}[width=0.24\linewidth,height=0.135\linewidth]{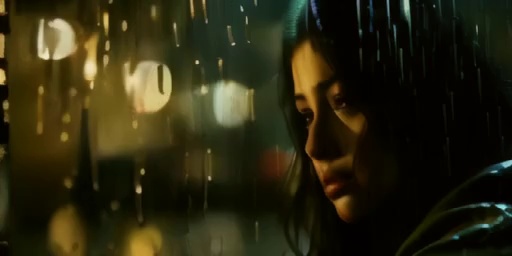}
        \put(35,20.5){\makebox[0pt]{\adjincludegraphics[height=.25\height,trim={{.23\width} {.4\height} {.47\width} 0},clip, cfbox=purple 2pt 0cm]{misc/main_figure/stereocrafter/a_woman_looking_out_in_the_rain_input_inpainting_results_sbs_000_.jpg}}}
    \end{overpic}
        &
    \begin{overpic}[width=0.24\linewidth,height=0.135\linewidth]{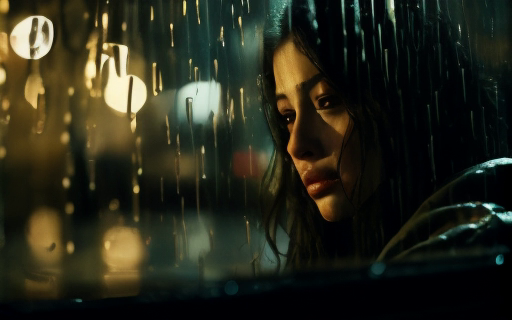}
        \put(35,20.5){\makebox[0pt]{\adjincludegraphics[height=.25\height,trim={{.18\width} {.5\height} {.52\width} 0},clip, cfbox=purple 2pt 0cm]{misc/main_figure/ours/a_woman_looking_out_in_the_rain-r_000__out.png}}}
    \end{overpic}
    \\
    \begin{overpic}[width=0.24\linewidth,height=0.135\linewidth]{misc/main_figure/immersepro/a_bonfire_is_lit_in_the_middle_of_a_fiel_input_immersepro_000__out.jpg}
        \put(78,16.5){\makebox[0pt]{\adjincludegraphics[height=.28\height,trim={{.53\width} {.5\height} {.2\width} 0},clip, cfbox=purple 2pt 0cm]{misc/main_figure/immersepro/a_bonfire_is_lit_in_the_middle_of_a_fiel_input_immersepro_000__out.jpg}}}
    \end{overpic}
        &
    \begin{overpic}[width=0.24\linewidth,height=0.135\linewidth]{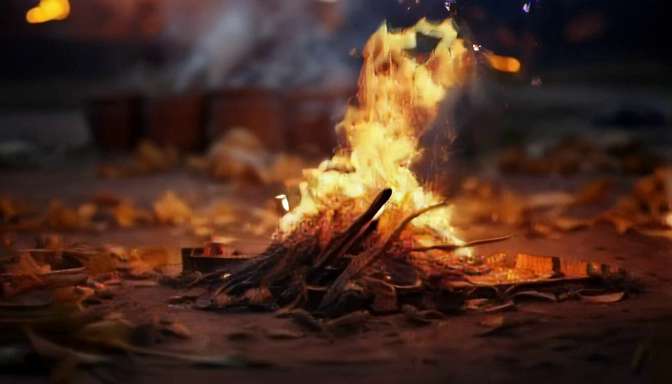}
        \put(78,16.5){\makebox[0pt]{\adjincludegraphics[height=.22\height,trim={{.52\width} {.48\height} {.22\width} 0},clip, cfbox=purple 2pt 0cm]{misc/main_figure/trajcrafter/a_bonfire_is_lit_in_the_middle_of_a_fiel_input_000.png}}}
    \end{overpic}
        &
    \begin{overpic}[width=0.24\linewidth,height=0.135\linewidth]{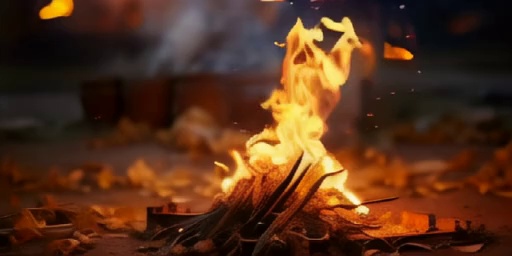}
        \put(78,16.5){\makebox[0pt]{\adjincludegraphics[height=.28\height,trim={{.53\width} {.4\height} {.2\width} 0},clip, cfbox=purple 2pt 0cm]{misc/main_figure/stereocrafter/a_bonfire_is_lit_in_the_middle_of_a_fiel_input_inpainting_results_sbs_000_.jpg}}}
    \end{overpic}
        &
    \begin{overpic}[width=0.24\linewidth,height=0.135\linewidth]{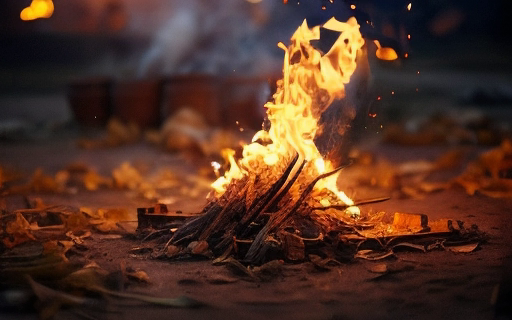}
        \put(78,16.5){\makebox[0pt]{\adjincludegraphics[height=.28\height,trim={{.53\width} {.5\height} {.2\width} 0},clip, cfbox=purple 2pt 0cm]{misc/main_figure/ours/a_bonfire_is_lit_in_the_middle_of_a_fiel-r_000__out.png}}}
    \end{overpic}
    \\
    \begin{overpic}[width=0.24\linewidth,height=0.135\linewidth]{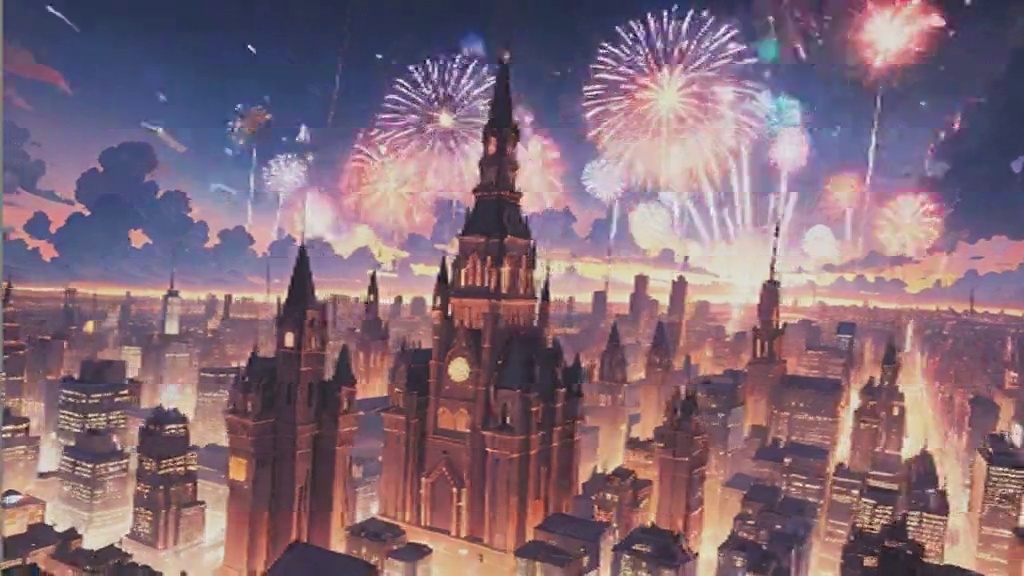}
        \put(57,20.5){\makebox[0pt]{\adjincludegraphics[height=0.1\linewidth,trim={{.3\width} {.5\height} {.2\width} 0},clip, cfbox=purple 2pt 0cm]{misc/main_figure/immersepro/fireworks_display_input_immersepro_015__out.jpg}}}
    \end{overpic}
        &
    \begin{overpic}[width=0.24\linewidth,height=0.135\linewidth]{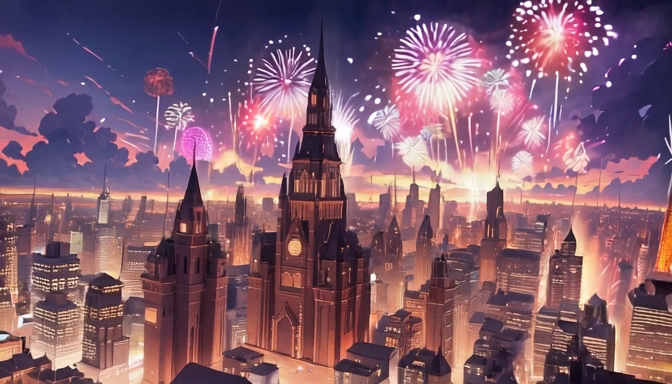}
        \put(57,20.5){\makebox[0pt]{\adjincludegraphics[height=0.1\linewidth,trim={{.3\width} {.5\height} {.2\width} 0},clip, cfbox=purple 2pt 0cm]{misc/main_figure/trajcrafter/fireworks_display_input_015.png}}}
    \end{overpic}
        &
    \begin{overpic}[width=0.24\linewidth,height=0.135\linewidth]{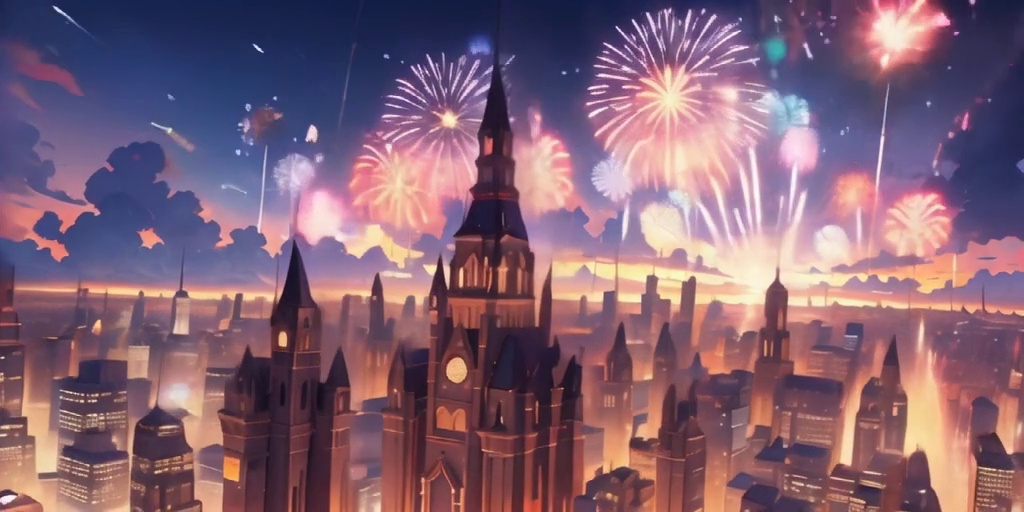}
        \put(57,20.5){\makebox[0pt]{\adjincludegraphics[height=0.1\linewidth,trim={{.32\width} {.5\height} {.22\width} 0},clip, cfbox=purple 2pt 0cm]{misc/main_figure/stereocrafter/fireworks_display_input_inpainting_results_sbs_045_.jpg}}}
    \end{overpic}
        &
    \begin{overpic}[width=0.24\linewidth,height=0.135\linewidth]{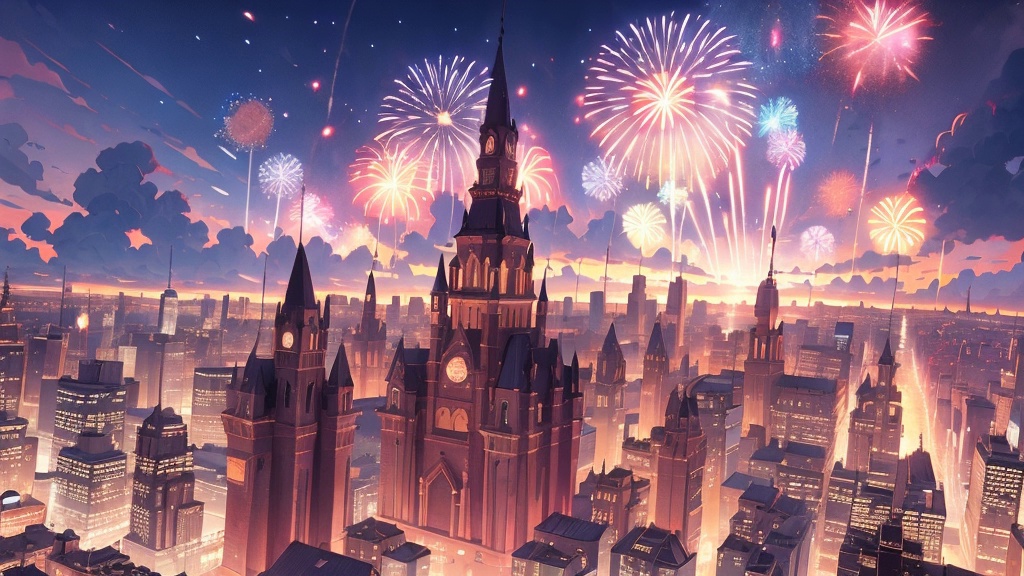}
        \put(57,20.5){\makebox[0pt]{\adjincludegraphics[height=0.1\linewidth,trim={{.3\width} {.5\height} {.2\width} 0},clip, cfbox=purple 2pt 0cm]{misc/main_figure/ours/fireworks_display-r_015__out.jpg}}}
    \end{overpic}
    \\

\end{tabular}
    \Description{Visual comparison with {ProPainter}, {RoDynRF}, {ImmersePro}, {TrajactoryCrafter}, {StereoCrafter}, and {StereoDiffusion}.
    Low-resolution methods such as {ImmersePro} and {StereoCrafter} produce videos with fewer fine details. {ProPainter} can hardly handle the fine details such as raindrops and fire flames, {RoDynRF} can hardly maintain the scene structure correctly, {StereoDiffusion} may produce more distortion since there are no temporal constraints (stronger when viewed in videos), and {TrajctoryCrafter} can hardly handle complex scenes. We provide video samples in the supplementary material.}
    \caption{
    Visual comparison with \textit{ProPainter}, \textit{RoDynRF}, \textit{ImmersePro}, \textit{TrajactoryCrafter}, \textit{StereoCrafter}, and \textit{StereoDiffusion}.
    Low-resolution methods such as \textit{ImmersePro} and \textit{StereoCrafter} produce videos with fewer fine details. \textit{ProPainter} can hardly handle the fine details such as raindrops and fire flames, \textit{RoDynRF} can hardly maintain the scene structure correctly, \textit{StereoDiffusion} may produce more distortion since there are no temporal constraints (stronger when viewed in videos), and \textit{TrajctoryCrafter} can hardly handle complex scenes. We provide video samples in the supplementary material.}
    \label{fig:main_figure}
\end{figure*}

\paragraph{Baseline Methods.}
Since there are no direct competitors for zero-shot stereo video generation, we compare against methods including \textit{ProPainter}, \textit{RoDynRF}, \textit{ImmersePro}, \textit{StereoDiffusion}, \textit{TrajectoryCrafter} , \textit{SVG}, and \textit{StereoCrafter}. For benchmarking purposes, we generated 40 video clips with \textit{DynamiCrafter} spanning diverse domains such as anime, human subjects, animals, various objects, and imaginary contents.
Note that generating one video clip with \textit{RoDynRF} requires approximately 10 hours on an NVIDIA A100, we therefore neglected it in our benchmark tables but presented a visualization in~\Cref{fig:main_figure}.
\paragraph{Quantitative Results.}
Unlike stereo conversion, the stereo generation task lacks a ground-truth right view. Direct geometry-based metrics such as disparity error, depth accuracy, or pixel-wise correspondence checks cannot be reliably computed.
Therefore, our evaluation focuses on two key aspects: semantic consistency and multi-view consistency between the left and right views of the generated video.
For semantic consistency, we follow prior works in novel-view rendering~\cite{cai2023diffdreamer,cai2024generative,kuang2024cvd}, and report CLIP-F~\cite{taited2023CLIPScore} to assess the consistency between the source-view videos and the generated views.
For multi-view consistency, we employ MEt3R~\cite{asim2025met3r} to assess the epipolar consistency between the left and right views.

\Cref{tab:main_performance} shows the performances against baselines.
In general, low-resolution video conversion methods such as \textit{ImmersePro} and \textit{StereoCrafter} produce better \textit{MEt3R} scores. We suspect this is due to the lower resolution depth maps, which come with fewer distracting fine depth details.
Our approach leverages this insight directly. 
By using dissolved depth maps, we selectively remove these potentially distracting details from the depth information without downsampling the video itself. Thus, our method is capable of producing high-resolution videos without compromising the epipolar consistency.

\begin{table}[h]
    \small
    \centering
    \caption{Benchmark results. The best and second best results are highlighted in \textbf{\color{red}red} and \textbf{\color{cyan} cyan} colors, respectively.}
    \setlength{\tabcolsep}{.5pt}
    \label{tab:main_performance}
    \begin{tabular}{r|cccccc|c}
         \toprule
         & {\footnotesize ProPainter}
         & {\footnotesize ImmersePro}
         & {\footnotesize StereoDiff.}
         & {\footnotesize Traj. Crafter*} 
         & {\footnotesize StereoCrafter}
         & {\footnotesize SVG}
         & {\footnotesize Ours} \\
         \midrule
         {\footnotesize CLIP-F $\uparrow$} & 96.45  & 96.99 & 91.09 & \textbf{\color{red} 97.43} & 93.59 & 96.76 & \textbf{\color{cyan} 97.16} \\
         {\footnotesize MEt3R $\downarrow$} & 8.82 &  5.69 & 6.09 & 6.21 & \textbf{\color{cyan}5.61} & 5.96 & \textbf{\color{red} 4.95} \\
         \bottomrule
         \multicolumn{8}{l}{\scriptsize *: \textit{TrajectoryCrafter} uses the first 10 frames for camera pose estimation. Thus, we tripled }\\
         \multicolumn{8}{l}{\scriptsize the frames. The metrics are computed with frames downsampled back.}
    \end{tabular}
\end{table}

\paragraph{Qualitative Results.}
\Cref{fig:main_figure} presents visual comparisons against various competing methods. The results show that our method consistently generates high-quality stereo videos, outperforming other approaches in terms of temporal coherence, resolution, and stereo effects.
In general, \textit{ProPainter} struggles to accurately reconstruct fine details (\textit{e.g.} shifted raindrops). \textit{RoDynRF} fails to maintain the scene structure during view changes. \textit{StereoDiffusion} introduces distortions due to the lack of temporal constraints, while \textit{ImmersePro} alters scene brightness with weaker stereo effects. \textit{TrajectoryCrafter} may wrongly interpret complex scenes, while \textit{StereoCrafter} uses downsampled videos without fine details.
We strongly encourage readers to watch the supplementary videos, since temporal coherence and spatial jittering are difficult to assess in static images.

\subsection{User Study} 

Since stereoscopic video quality is not fully captured by quantitative metrics alone, we conduct a user study to assess subjective visual experience from a perceptual standpoint.

\paragraph{Protocol}
We conducted a single-blind user study involving 29 participants (6 female, 23 male), none of whom reported binocular vision issues, and five of whom had prior VR headset experience. The study followed institutional guidelines, and all participants provided informed consent. Participants evaluated short stereoscopic clips on a Meta Quest 3 headset (using personal glasses if needed), rating each along four dimensions on a 5-point Likert scale:
\begin{enumerate}
    \item \textit{Frame Quality (5 is best, 1 is poor)}
    \item \textit{Temporal Coherence (5 is best, 1 is poor)}
    \item \textit{Stereoscopic Effects (5 is best, 1 is poor)}
    \item \textit{Overall Conformity (5 is best, 1 is poor)}
\end{enumerate}

Specifically, we define frame quality as the quality of the generated 2D images, temporal coherence as the consistency between frames, and stereoscopic effects as a subjective experience of the stereo videos. For overall conformity, while some stereo videos may exhibit stronger stereo effects, they can also induce discomfort or motion sickness in viewers. Therefore, "overall conformity" reflects our users' subjective perception of the generated videos.

\begin{figure}[t]
    \centering
    \adjincludegraphics[width=.9\linewidth,trim={0 {.15\height} 0 {.25\height}},clip]{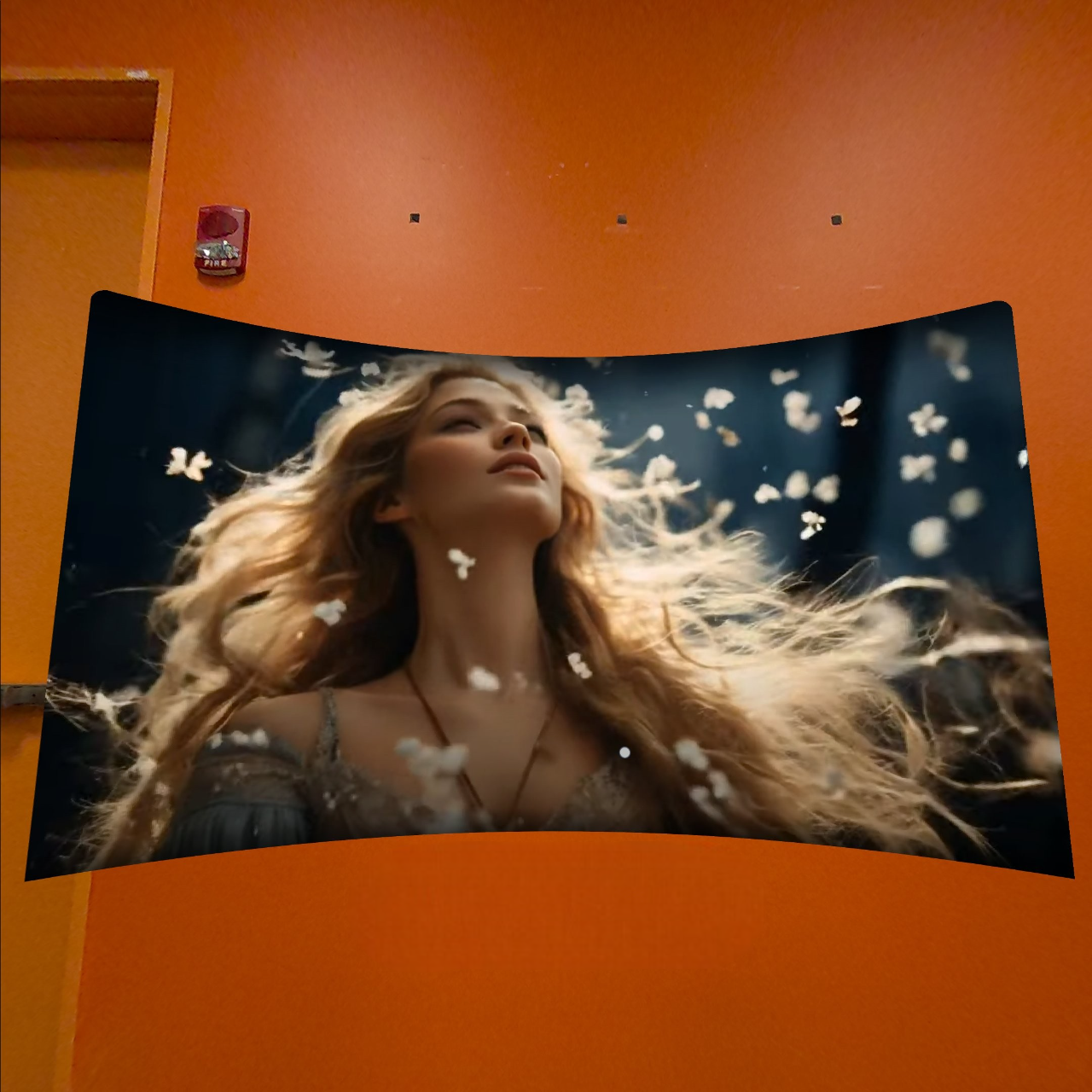}
    \Description{A Meta Quest 3 screenshot of a 3D video in action.}
    \caption{A Meta Quest 3 screenshot of a 3D video in action.}
    \label{fig:quest}
\end{figure}

During the evaluation, participants were unaware of which method produced each clip. First, they viewed all clips once for scale familiarity and setup confirmation, then replayed clips as needed before rating. Two participants reported slight 3D sickness afterward. \Cref{fig:quest} showcases a 3D video being viewed through the headset.

\begin{figure}
    \centering
    \includegraphics[width=.9\linewidth]{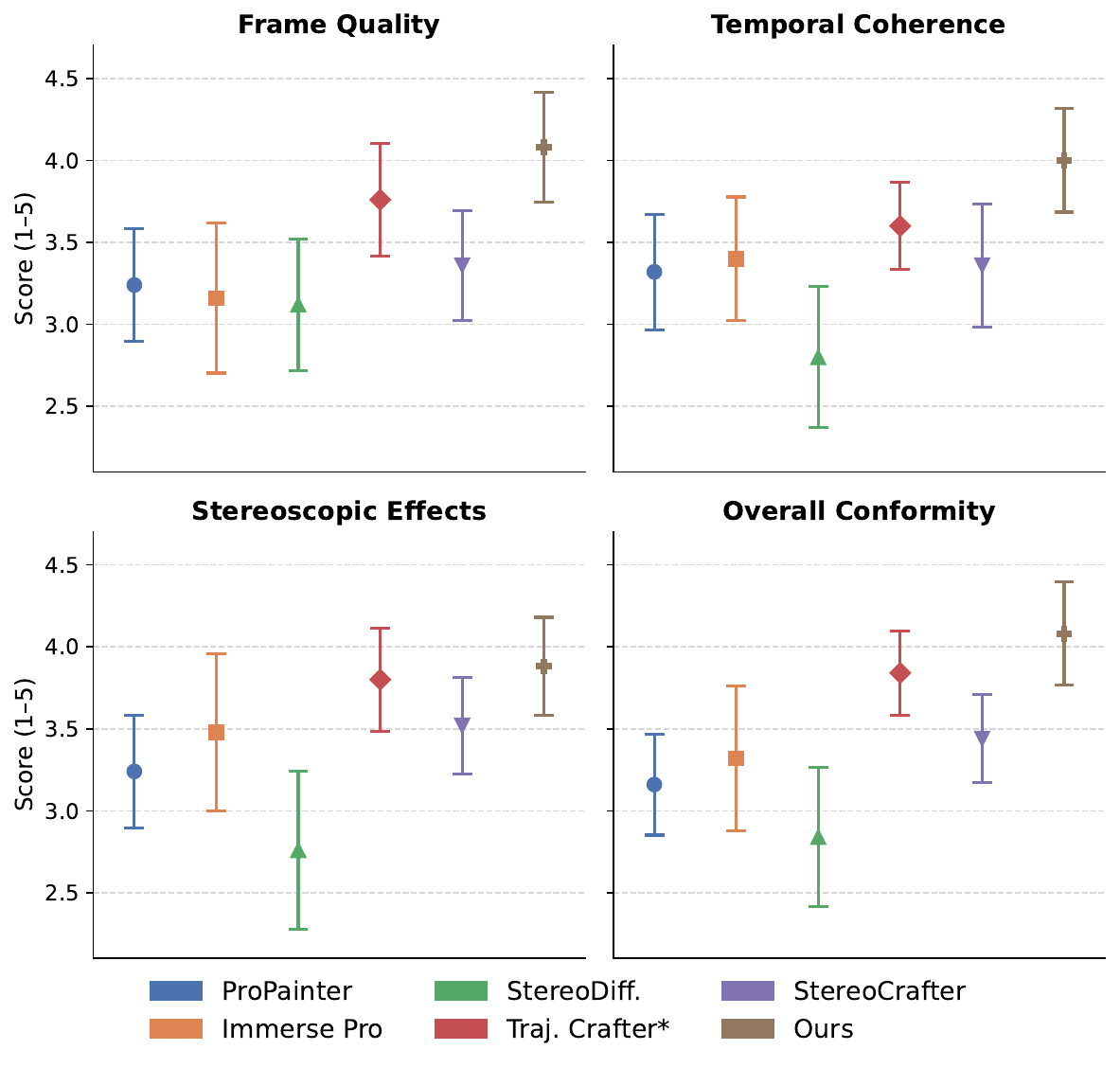}
    \Description{95\% confidence interval plot for the user study.}
    \caption{95\% confidence interval plot for the user study.}
    \label{fig:user_study_ci}
\end{figure}

\paragraph{Results}
The user study results and confidence interval plots are summarized in~\Cref{tab:user_study} and~\Cref{fig:user_study_ci}, respectively. Our method achieved the highest overall user rating.
Although \textit{TrajectoryCrafter} was not originally designed for stereo generation, it nevertheless delivered the second-best performance, particularly demonstrating strong stereoscopic effects.

\begin{table}[h]
    \small
    \centering
    \caption{User study results comparing preference scores (1–5). The highest and second-highest values are highlighted in \textbf{\color{red}red} and \textbf{\color{cyan}cyan}, respectively.}
    \setlength{\tabcolsep}{1pt}
    \label{tab:user_study}
    \begin{tabular}{c|ccccc|c}
         \toprule
         & {\scriptsize ProPainter}
         & {\scriptsize ImmersePro}
         & {\scriptsize StereoDiff.}
         & {\scriptsize Traj. Crafter*} 
         & {\scriptsize StereoCrafter}
         & {\scriptsize Ours} \\
         \midrule
         {\scriptsize Frame Quality} & 3.27 & 3.20 & 3.12 & \textbf{\color{cyan} 3.75} & 3.38 & \textbf{\color{red} 4.05} \\
         {\scriptsize Temporal Coherence} & 3.34 & 3.42 & 2.83 & \textbf{\color{cyan} 3.57} & 3.38 & \textbf{\color{red} 3.96} \\
         {\scriptsize Stereoscopic Effects} & 3.27 & 3.50 & 2.75 & \textbf{\color{red} 3.79} & \textbf{\color{cyan} 3.52} & \textbf{\color{red} 3.79} \\
         {\scriptsize Overall Conformity} & 3.20 & 3.34 & 2.83 & \textbf{\color{cyan} 3.83} & 3.46 & \textbf{\color{red} 3.98}\\
         \bottomrule
    \end{tabular}
\end{table}

\begin{figure*}
    \centering
     \setlength{\tabcolsep}{0pt}
     \renewcommand{\arraystretch}{0}
    \begin{tabular}{cccc}
        $t=10$ & $t=20$ & $t=30$ & $t=40$ \\
        \toprule
        \includegraphics[width=0.24\linewidth,height=0.13\linewidth]{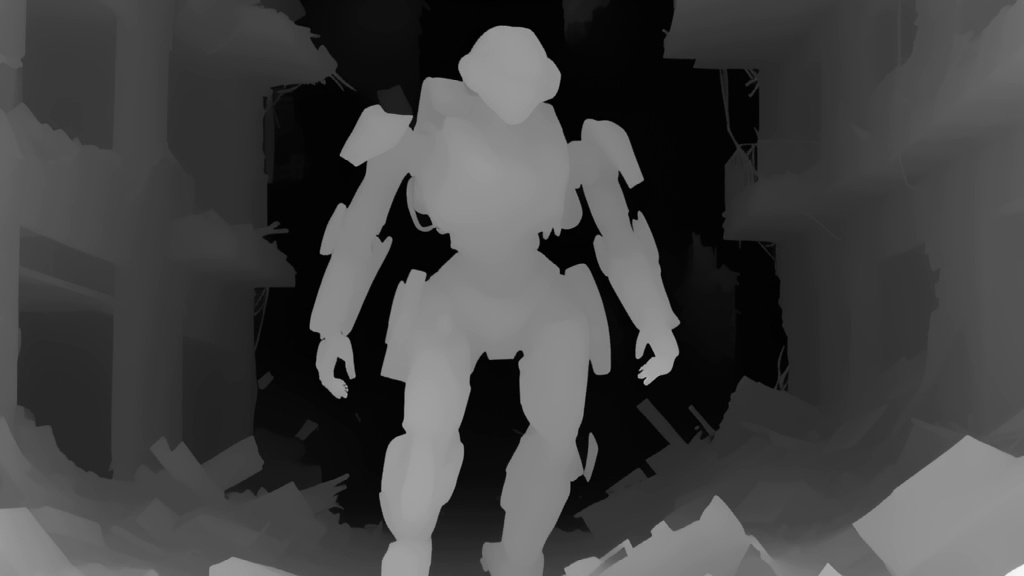}
        & 
        \includegraphics[width=0.24\linewidth,height=0.13\linewidth]{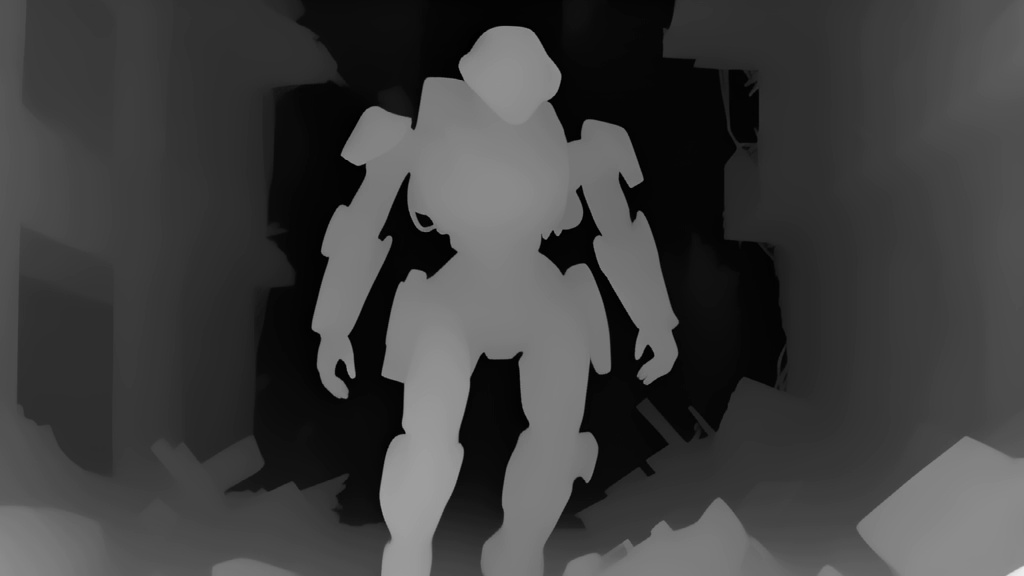} 
        & 
        \includegraphics[width=0.24\linewidth,height=0.13\linewidth]{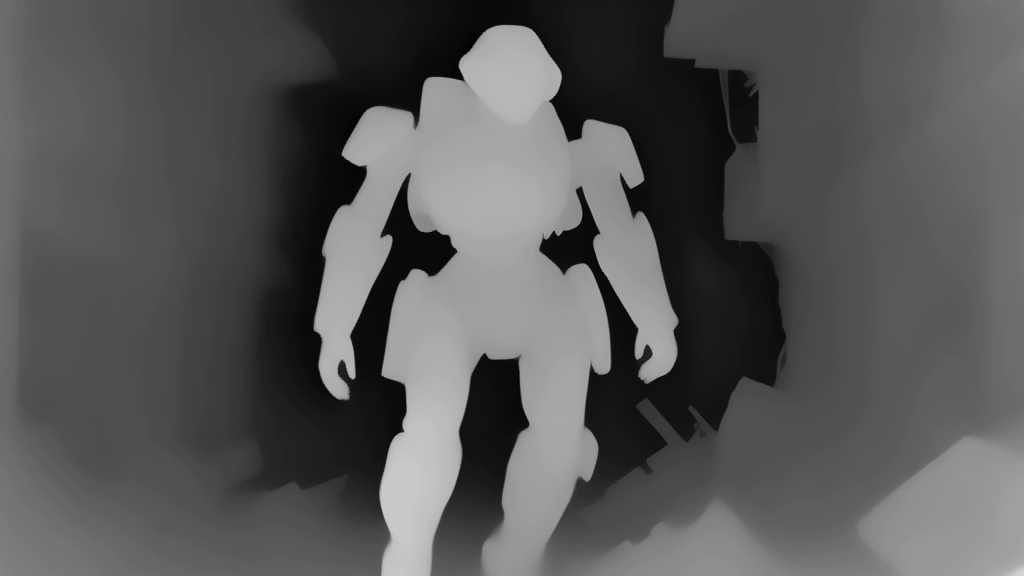} 
        &
        \includegraphics[width=0.24\linewidth,height=0.13\linewidth]{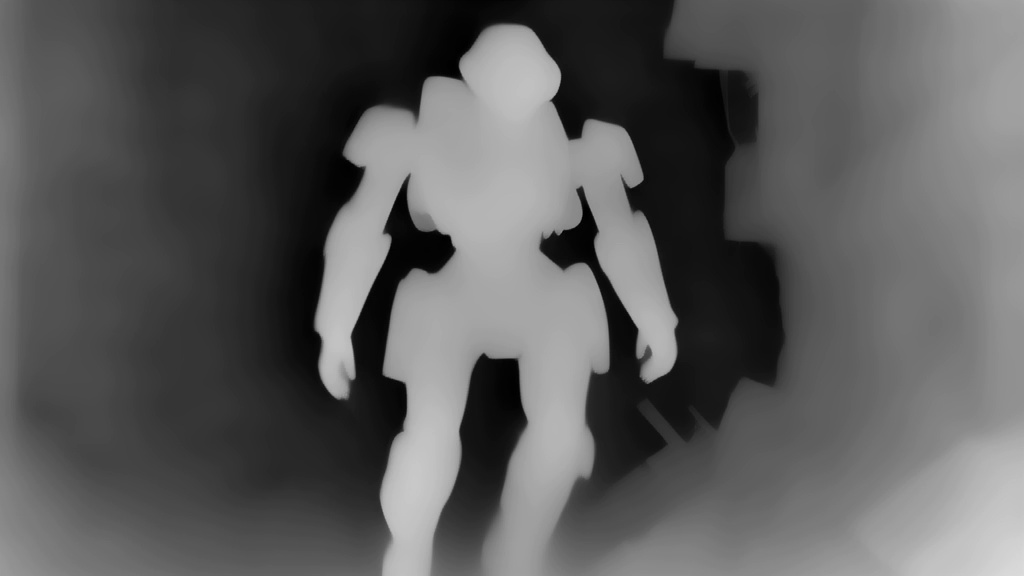} 
        \\
        \begin{overpic}[width=0.24\linewidth,height=0.13\linewidth]{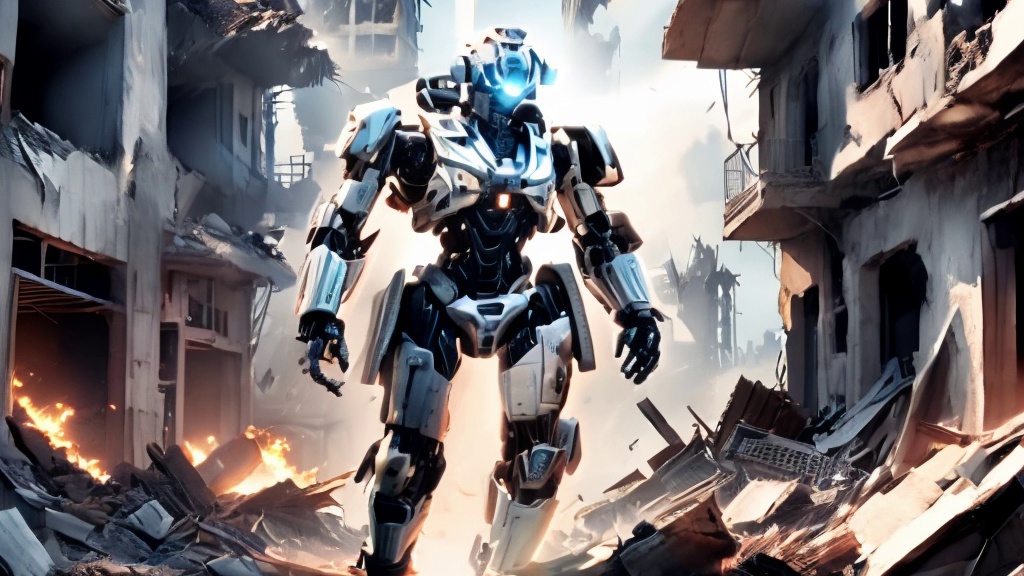}
            \put(15,1.5){\makebox[0pt]{\adjincludegraphics[width=0.065\linewidth,trim={{.41\width} {.4\height} {.44\width} {.1\height}},clip, cfbox=purple 2pt 0cm]{misc/dslv_showcase/a_robot_is_walking_through_a_destroyed_c-r-dslv40-00.jpg}}}
        \end{overpic}
        & 
        \begin{overpic}[width=0.24\linewidth,height=0.13\linewidth]{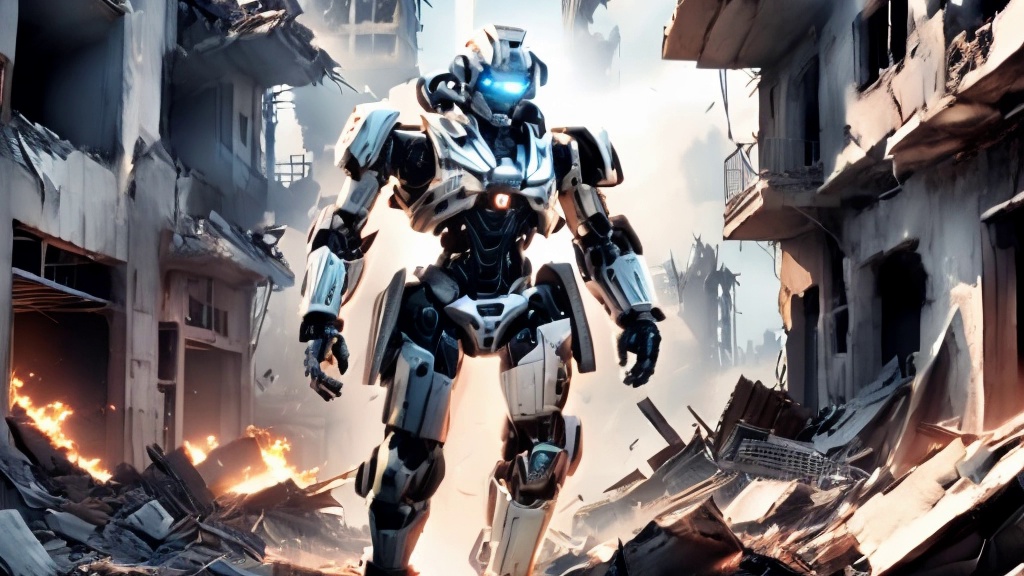}
            \put(15,1.5){\makebox[0pt]{\adjincludegraphics[width=0.065\linewidth,trim={{.41\width} {.4\height} {.44\width} {.1\height}},clip, cfbox=purple 2pt 0cm]{misc/dslv_showcase/a_robot_is_walking_through_a_destroyed_c-r-dslv30-00.jpg}}}
        \end{overpic}
        & 
        \begin{overpic}[width=0.24\linewidth,height=0.13\linewidth]{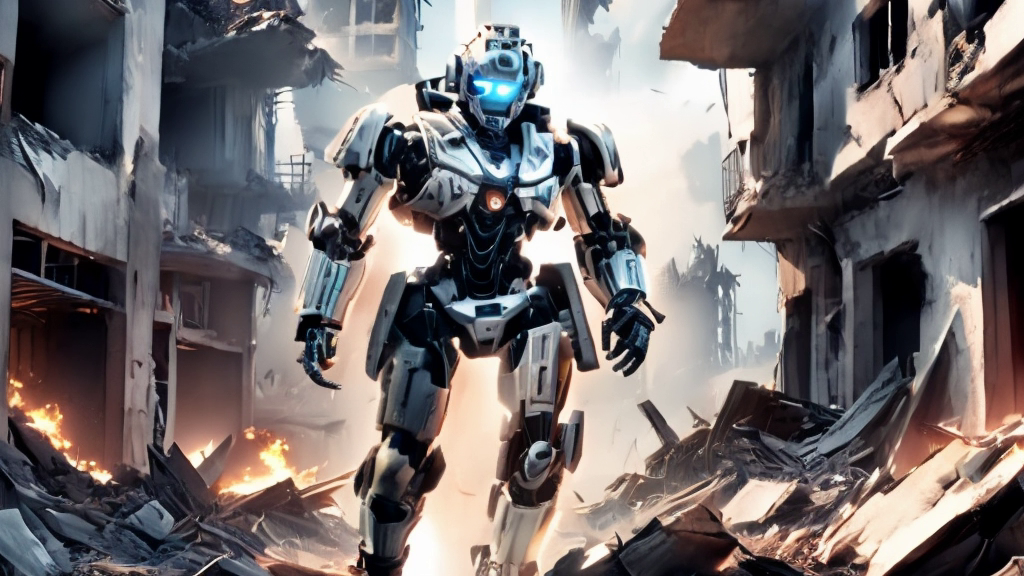}
            \put(15,1.5){\makebox[0pt]{\adjincludegraphics[width=0.065\linewidth,trim={{.41\width} {.4\height} {.44\width} {.1\height}},clip, cfbox=purple 2pt 0cm]{misc/dslv_showcase/a_robot_is_walking_through_a_destroyed_c-r-dslv20-00.png}}}
        \end{overpic}
        & 
        \begin{overpic}[width=0.24\linewidth,height=0.13\linewidth]{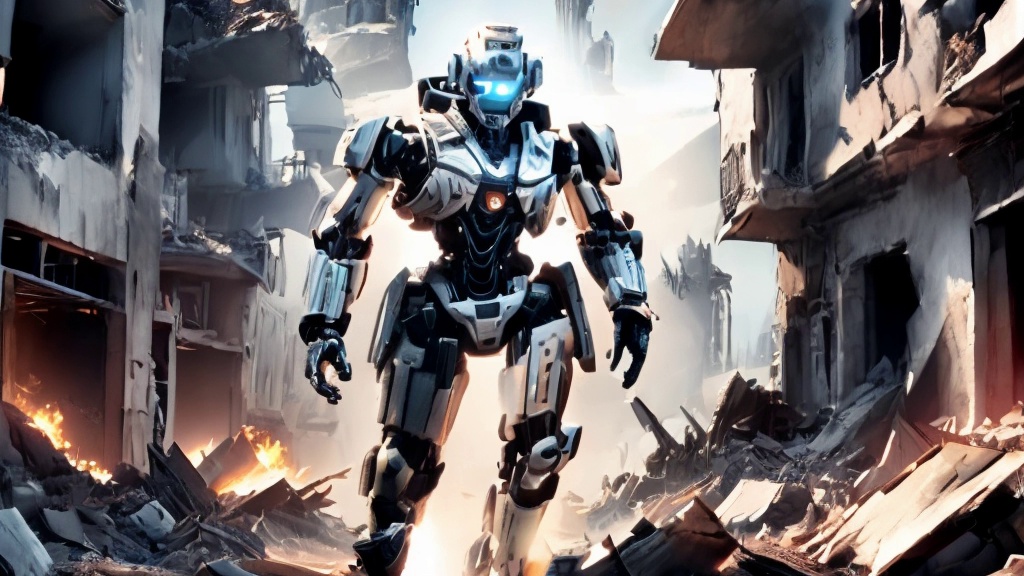}
            \put(15,1.5){\makebox[0pt]{\adjincludegraphics[width=0.065\linewidth,trim={{.41\width} {.4\height} {.44\width} {.1\height}},clip, cfbox=purple 2pt 0cm]{misc/dslv_showcase/a_robot_is_walking_through_a_destroyed_c-r-dslv10-00.jpg}}}
        \end{overpic}
        \\
        \\
        \includegraphics[width=0.24\linewidth,height=0.13\linewidth]{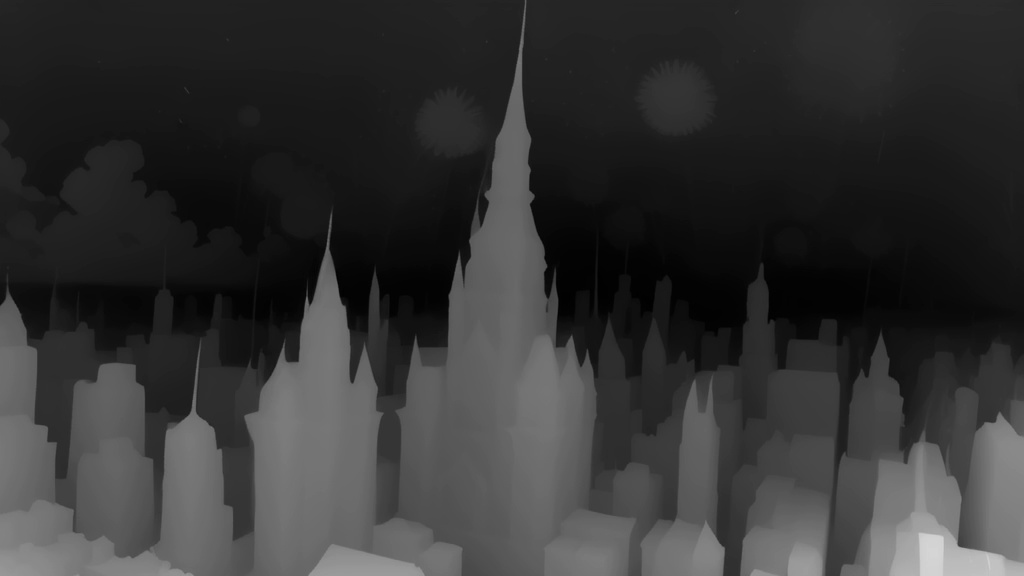}
        & 
        \includegraphics[width=0.24\linewidth,height=0.13\linewidth]{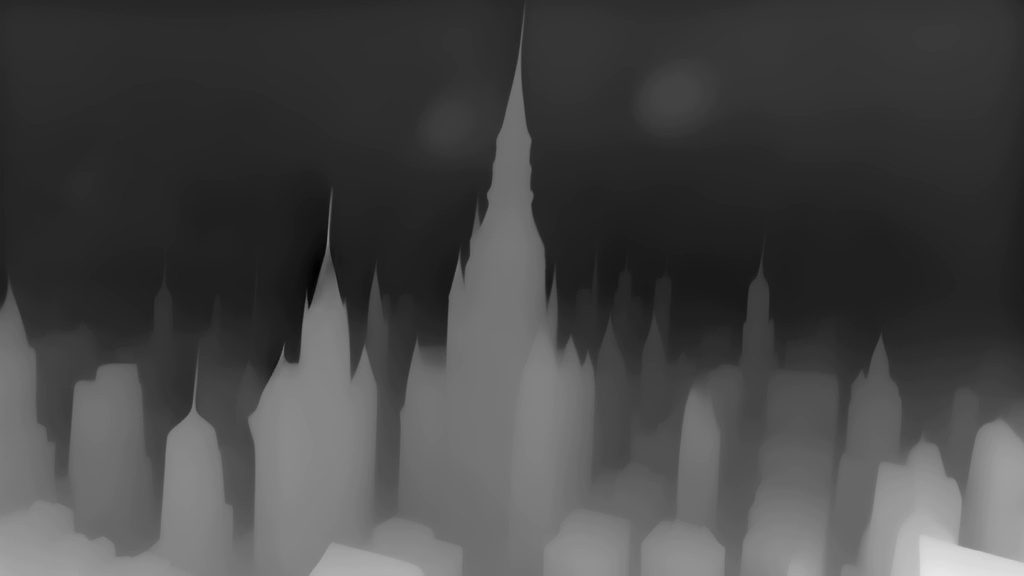} 
        & 
        \includegraphics[width=0.24\linewidth,height=0.13\linewidth]{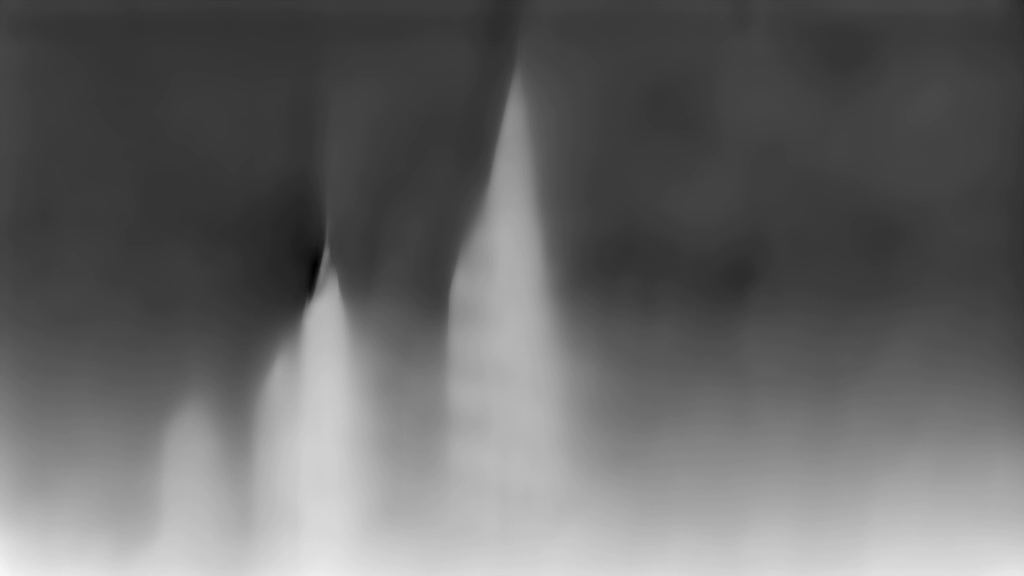} 
        & 
        \includegraphics[width=0.24\linewidth,height=0.13\linewidth]{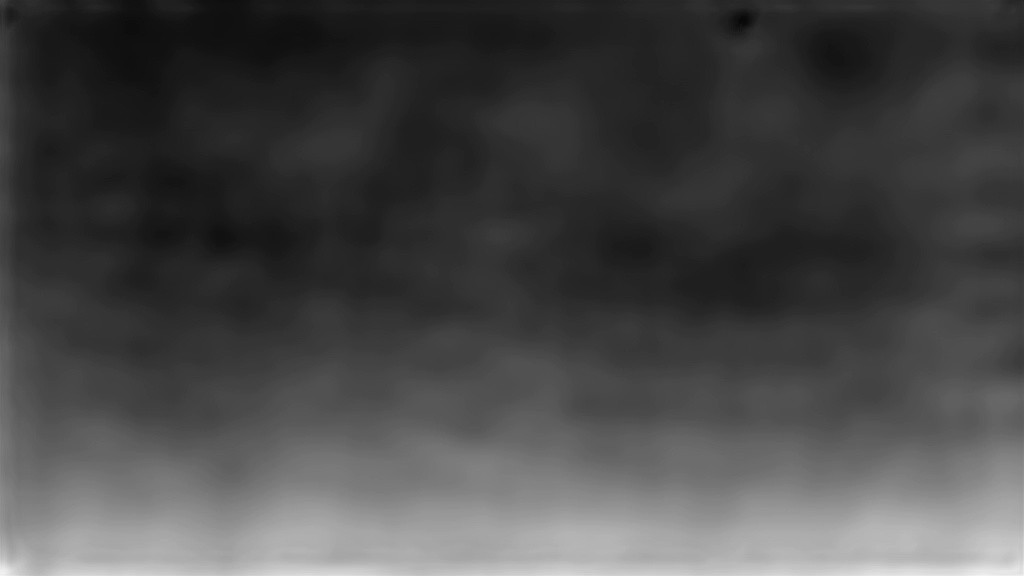}
        \\
        \begin{overpic}[width=0.24\linewidth,height=0.13\linewidth]{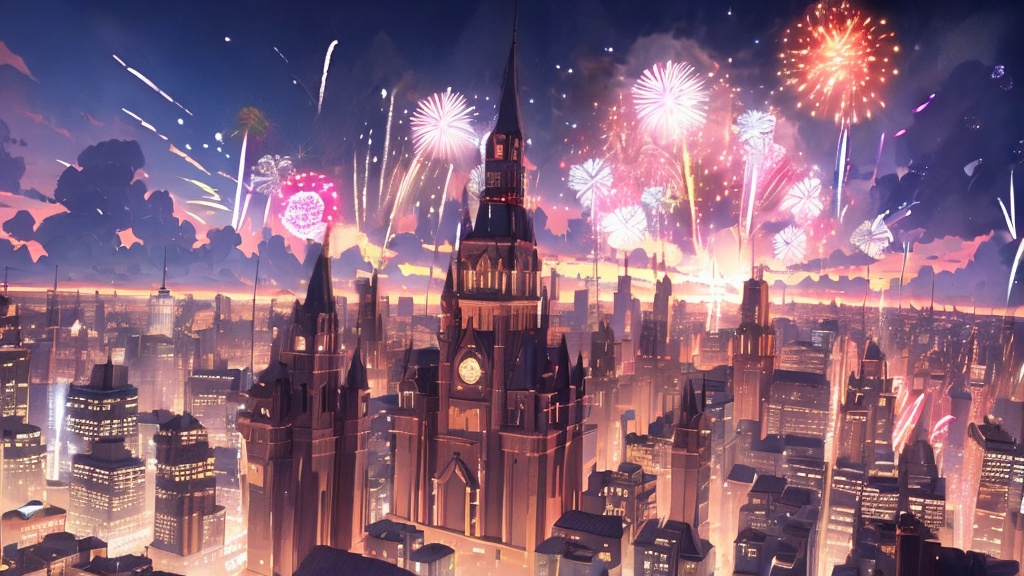}
            \put(18,1.5){\makebox[0pt]{\adjincludegraphics[width=0.075\linewidth,trim={{.21\width} {.3\height} {.64\width} {.3\height}},clip, cfbox=purple 2pt 0cm]{misc/dslv_showcase/fireworks_display-r-dslv40-00.jpg}}}
        \end{overpic}
        & 
        \begin{overpic}[width=0.24\linewidth,height=0.13\linewidth]{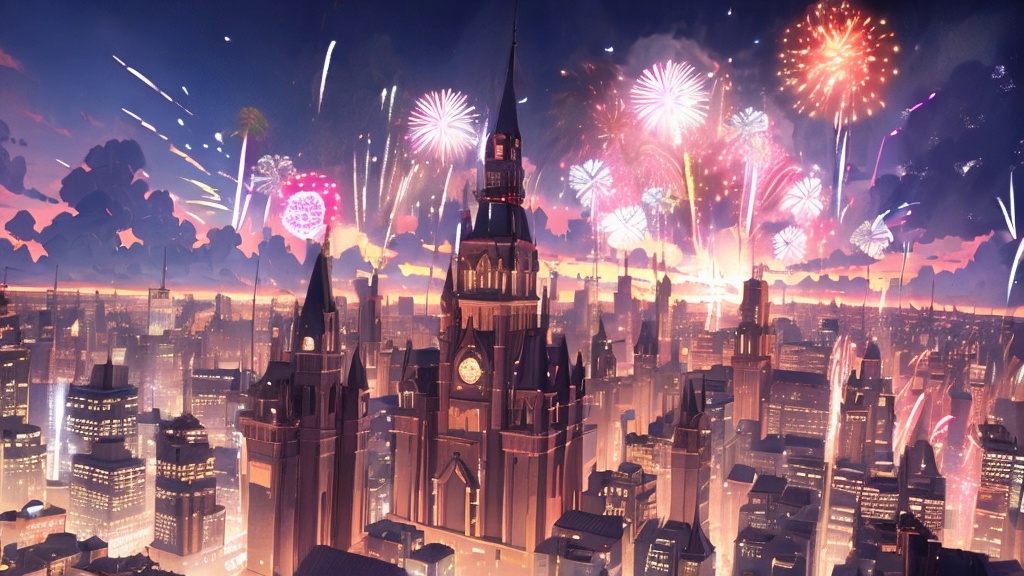}
            \put(18,1.5){\makebox[0pt]{\adjincludegraphics[width=0.075\linewidth,trim={{.21\width} {.3\height} {.64\width} {.3\height}},clip, cfbox=purple 2pt 0cm]{misc/dslv_showcase/fireworks_display-r-dslv30-00.jpg}}}
        \end{overpic}
        & 
        \begin{overpic}[width=0.24\linewidth,height=0.13\linewidth]{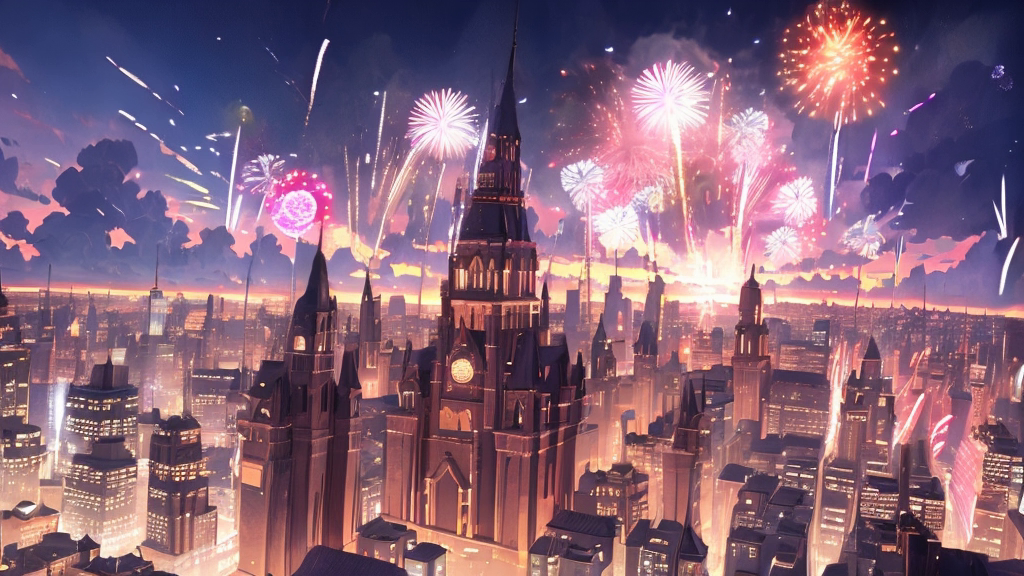}
            \put(18,1.5){\makebox[0pt]{\adjincludegraphics[width=0.075\linewidth,trim={{.21\width} {.3\height} {.64\width} {.3\height}},clip, cfbox=purple 2pt 0cm]{misc/dslv_showcase/fireworks_display-r-dslv20-00.png}}}
        \end{overpic}
        & 
        \begin{overpic}[width=0.24\linewidth,height=0.13\linewidth]{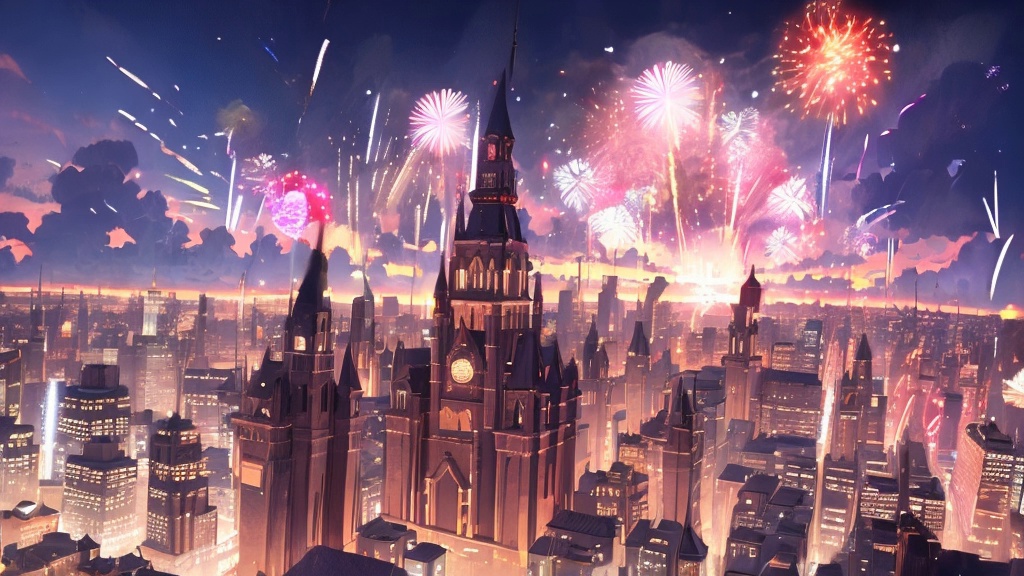}
            \put(18,1.5){\makebox[0pt]{\adjincludegraphics[width=0.075\linewidth,trim={{.21\width} {.3\height} {.64\width} {.3\height}},clip, cfbox=purple 2pt 0cm]{misc/dslv_showcase/fireworks_display-r-dslv10-00.jpg}}}
        \end{overpic}
        \\
    \end{tabular}
    \Description{Different dissolving levels. Top: a simpler case with fewer intricate structures. Bottom: a complex case with fine textures and detailed depth variations.}
    \caption{Different dissolving levels. Top: a simpler case with fewer intricate structures. Bottom: a complex case with fine textures and detailed depth variations.}
    \label{fig:dslv_res}
\end{figure*}

\subsection{Ablations}

\subsubsection{Dissolved Depth Maps}
\label{sec:dslv_depth_ablation}
As mentioned in our main text, we found that the optimal degree of dissolving strength depends on scene complexity. For simpler scenes with fewer intricate structures, a stronger dissolving effect (\textit{e.g.} $t=40$) produces better results, whilst a weaker dissolving effect (\textit{e.g.} $t=20$) is more effective for more complex scenes with fine textures and detailed depth variations. Here, we present some visual results to support this finding in~\Cref{fig:dslv_res}.

\paragraph{Downsampling vs. Depth Dissolving}
Apart from using dissolved depth maps, downsampling is another intuitive method to create depth maps at a lower frequency. As shown in~\Cref{tab:downsample_depth}, having downsampled depth maps may generally create better numerical results. However, the strengths of stereoscopic effects might be presented in a random manner for different videos.

\begin{table}[h]
    \footnotesize
    \centering
    \setlength{\tabcolsep}{2pt}
    \caption{Numerical results for different scales of the depth maps. We report MEt3R scores in this table.}
    \label{tab:downsample_depth}
    \begin{tabular}{r|ccc|ccccc}
        \toprule
        & D. Pro &  D. Anything &  V. D. Anything &  \multicolumn{5}{c}{D. Crafter}\\
        \cmidrule(lr){5-9}
        scale. &  &  &  & n/a & 10 & 20 & 30 & 40 \\
        \midrule
        $1$ & 6.78 & 6.71 & 6.79 & 6.70 & 7.34 & 6.90 & 6.49 & 6.72\\
        \midrule
        $1 / 2$ & 6.59 & 6.57 & 5.91 & 6.47 & 6.47 & 5.91 & 5.52 & 5.67 \\
        $1 / 4$ & 5.81 & 6.07 & 6.20 & 6.57 & 5.67 & 5.54 & 4.95 & 4.96\\
        \bottomrule
    \end{tabular}
\end{table}
\begin{table}[h]
\footnotesize
\centering
\setlength{\tabcolsep}{3pt}
\caption{Results of Gaussian and median blurs with 7$\times$7 and 11$\times$11 kernels.}
\label{tab:blur_dissolve}
\begin{tabular}{l|cc|cc|c}
\toprule
    & Gauss. 7$\times$7 & Gauss. 11$\times$11
    & Med. 7$\times$7 & Med. 11$\times$11
    & Dissolved Depth\\
\midrule
MEt3R $\downarrow$ & 5.44 & 5.31 & 5.31 & 5.32 & \textbf{4.95} \\
\bottomrule
\end{tabular}
\end{table}

\paragraph{Blurring vs. Depth Dissolving}
Similar to the downsampling approach above, the blurring methods are data-agnostic, whereas depth dissolving is semantic-aware, preserving critical scene semantics. 
Notably, blurred depth maps produced by simple heuristic filters follow the same principle. While less effective than our approach, they still yield measurable improvements. Together, these observations demonstrate that dissolved depth maps provide a principled means of imposing coarse geometric structure without the drawbacks of high-frequency depth priors.

\begin{minipage}{\linewidth}
\begin{minipage}[t]{0.46\linewidth}
\centering
\setlength{\tabcolsep}{3pt}
\setlength{\abovecaptionskip}{1.5em}
\setlength{\belowcaptionskip}{1em}
\captionsetup{hypcap=false}
\captionof{table}{Ablation on restart parameters $K$ (window) and $L$ (rounds).}
\captionsetup{hypcap=true} 
\label{tab:restart_ablation}
\small
\setlength{\tabcolsep}{2pt}
\begin{tabular}{cccccc}
\hline
$K$ & $L$ & Total Steps & MEt3R $\downarrow$ \\
\hline
6  & 5 & 30  & 0.0525 \\
6  & 7 & 42  & \textbf{0.0513} \\
6  & 9 & 54  & 0.0546 \\
11 & 5 & 55  & 0.0555 \\
11 & 7 & 77  & 0.0545 \\
11 & 9 & 99  & 0.0580 \\
21 & 5 & 105 & 0.0609 \\
21 & 7 & 147 & 0.0607 \\
21 & 9 & 189 & 0.0620 \\
\hline
\end{tabular}
\end{minipage}
\hspace{0.1em}
\begin{minipage}[t]{0.42\linewidth}
\centering
\setlength{\tabcolsep}{2pt}
\setlength{\abovecaptionskip}{1.5em}
\setlength{\belowcaptionskip}{1em}
\captionsetup{hypcap=false}
\captionof{table}{Ablation on iterative refinement rounds $N$ using $K=6, L=7$.}
\captionsetup{hypcap=true} 
\label{tab:refinement_ablation}
\small
\setlength{\tabcolsep}{2pt}
\begin{tabular}{cccc}
\hline
$K \times L$ & $N$ & No. Steps & MEt3R $\downarrow$ \\
\hline
6$\times$7 & 0 & 42  & 0.0513 \\
6$\times$7 & 2 & 56  & 0.0524 \\
6$\times$7 & 4 & 70  & \textbf{0.0495} \\
6$\times$7 & 6 & 84  & 0.0523 \\
6$\times$7 & 8 & 98  & 0.0535 \\
\hline
\end{tabular}
\end{minipage}
\end{minipage}

\subsubsection{Noisy Restart and Iterative Refinement.}
Noisy restart selectively injects controlled noise into disoccluded regions during early diffusion steps, shaping global stereo disparity and structural coherence. Iterative refinement performs targeted re-denoising ($L=1$1) at specific steps without noise reintroduction, harmonizing filled regions with warped latents.
\begin{table}[t]
    \small
    \centering
    \caption{Performances obtained with different baseline distances.}
    \label{tab:baseline_sensitivity}
    \begin{tabular}{c|cccc}
        \toprule
         Baseline & $4\%$ &  $8\%$ &  $12\%$ &  $16\%$ \\
        \midrule
        CLIP-F $\uparrow$ & 98.24 & 97.24 & 97.02 & 96.78 \\
        MEt3R $\downarrow$ & 4.85 & 5.77 & 6.71 & 7.58 \\
        \bottomrule
    \end{tabular}
\end{table}%
\begin{figure}[t]
     \centering
     \begin{subfigure}[b]{0.4\linewidth}
         \centering
         \includegraphics[width=\textwidth]{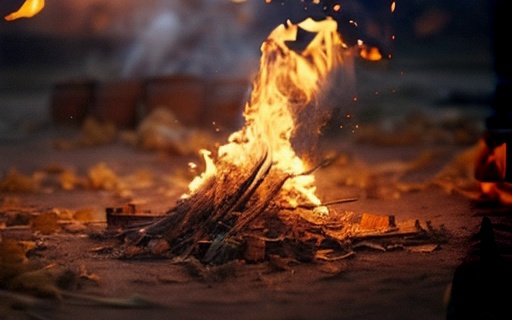}
     \end{subfigure}
     \begin{subfigure}[b]{0.4\linewidth}
         \centering
         \includegraphics[width=\textwidth]{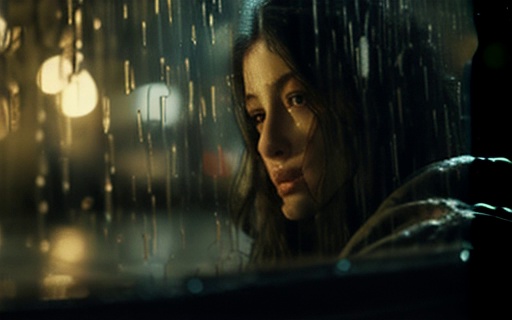}
     \end{subfigure}

     \begin{subfigure}[b]{0.4\linewidth}
         \centering
         \includegraphics[width=\textwidth]{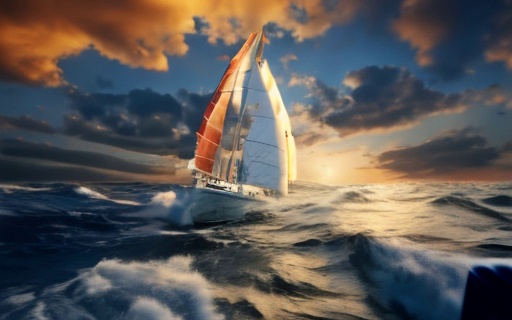}
         \caption{Good Results (MEt3R $<6$)}
     \end{subfigure}
     \begin{subfigure}[b]{0.4\linewidth}
         \centering
         \includegraphics[width=\textwidth,]{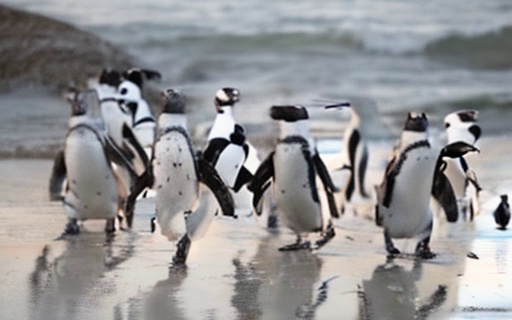}
         \caption{Bad Results (MEt3R $>6$)}
     \end{subfigure}
    \\
    \Description{We visualize the reconstruction results under $16\%$ baseline. The left column presents results that can be handled well, and the right column presents results that exhibit strong distortions.}
    \caption{We visualize the reconstruction results under $16\%$ baseline. The left column presents results that can be handled well, and the right column presents results that exhibit strong distortions.}
    \label{fig:baseline_16}
\end{figure}
\begin{figure*}
     \centering
     \rotatebox{90}{~~~~~~~~\footnotesize \makecell{No Dissolve}}
     \begin{subfigure}[b]{0.24\textwidth}
         \centering
        \includegraphics[width=\textwidth,height=0.5625\textwidth]{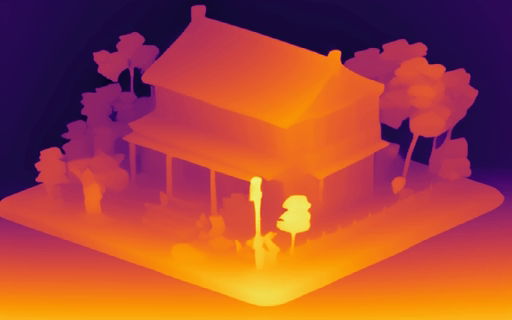}
     \end{subfigure}%
     \begin{subfigure}[b]{0.24\textwidth}
         \centering
        \includegraphics[width=\textwidth,height=0.5625\textwidth]{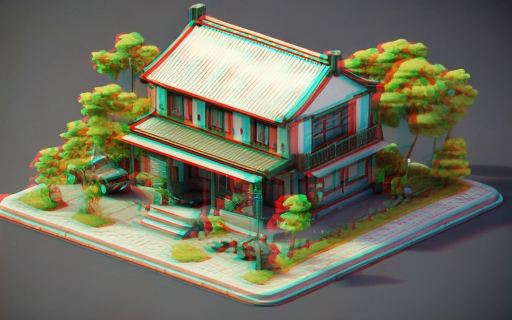}
     \end{subfigure}%
     \begin{subfigure}[b]{0.24\textwidth}
         \centering
        \includegraphics[width=\textwidth,height=0.5625\textwidth]{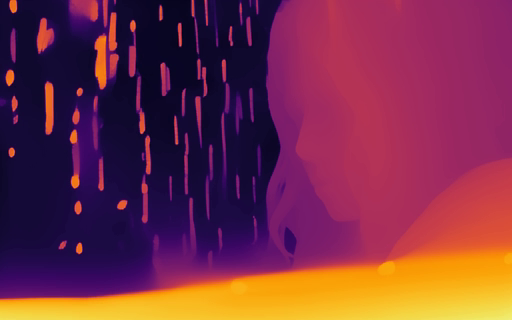}
     \end{subfigure}%
     \begin{subfigure}[b]{0.24\textwidth}
         \centering
        \includegraphics[width=\textwidth,height=0.5625\textwidth]{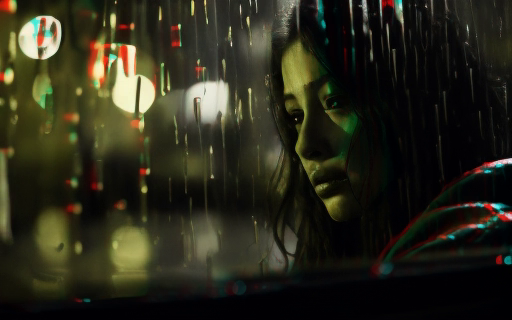}
     \end{subfigure}
     \\
     \rotatebox{90}{~~~~~~~~\footnotesize \makecell{$t=5$}}
     \begin{subfigure}[b]{0.24\textwidth}
         \centering
        \includegraphics[width=\textwidth,height=0.5625\textwidth]{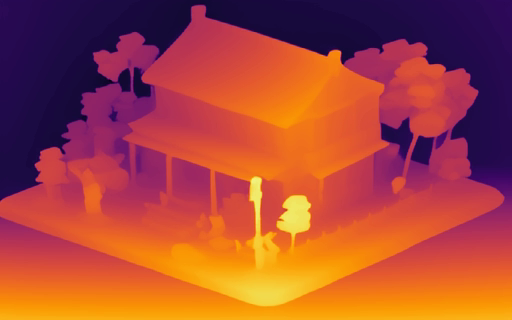}
     \end{subfigure}%
     \begin{subfigure}[b]{0.24\textwidth}
         \centering
        \includegraphics[width=\textwidth,height=0.5625\textwidth]{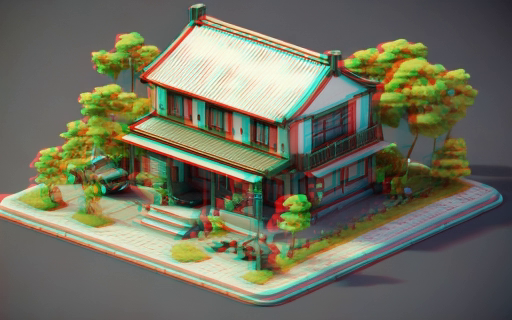}
     \end{subfigure}%
     \begin{subfigure}[b]{0.24\textwidth}
         \centering
        \includegraphics[width=\textwidth,height=0.5625\textwidth]{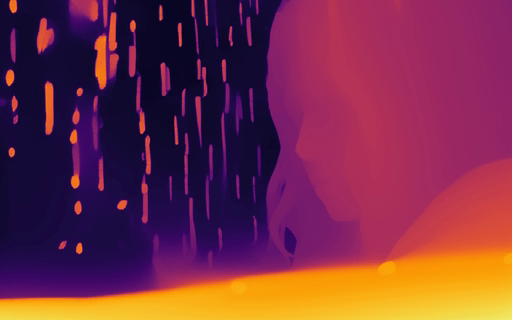}
     \end{subfigure}%
     \begin{subfigure}[b]{0.24\textwidth}
         \centering
        \includegraphics[width=\textwidth,height=0.5625\textwidth]{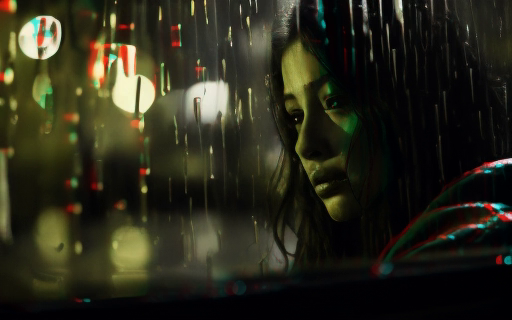}
     \end{subfigure}
     \\
     \rotatebox{90}{~~~~~~~~~~~\footnotesize \makecell{$t=15$}}
     \begin{subfigure}[b]{0.24\textwidth}
         \centering
        \includegraphics[width=\textwidth,height=0.5625\textwidth]{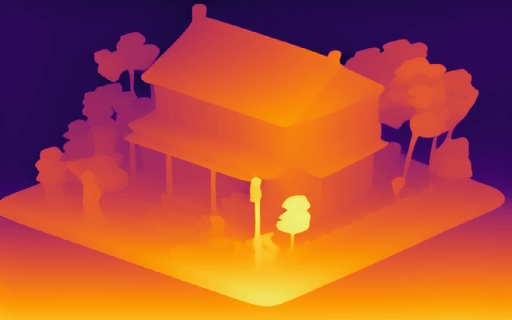}
     \end{subfigure}%
     \begin{subfigure}[b]{0.24\textwidth}
         \centering
        \includegraphics[width=\textwidth,height=0.5625\textwidth]{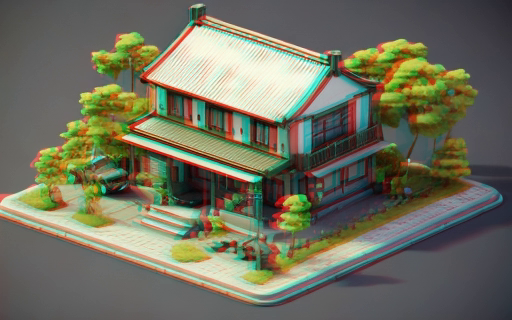}
     \end{subfigure}%
     \begin{subfigure}[b]{0.24\textwidth}
         \centering
        \includegraphics[width=\textwidth,height=0.5625\textwidth]{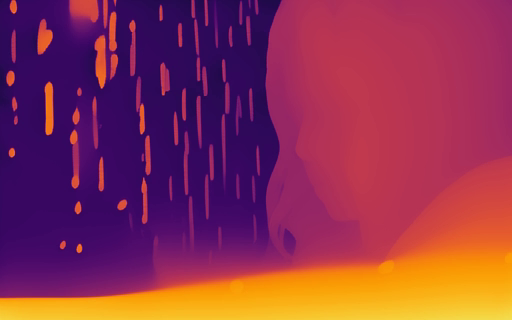}
     \end{subfigure}%
     \begin{subfigure}[b]{0.24\textwidth}
         \centering
        \includegraphics[width=\textwidth,height=0.5625\textwidth]{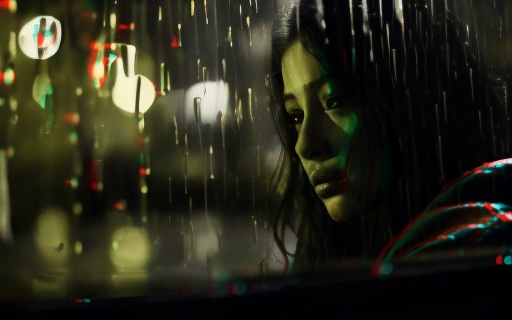}
     \end{subfigure}
     \\
     \rotatebox{90}{~~~~~~\footnotesize \makecell{$t=25$}}
     \begin{subfigure}[b]{0.24\textwidth}
         \centering
        \includegraphics[width=\textwidth,height=0.5625\textwidth]{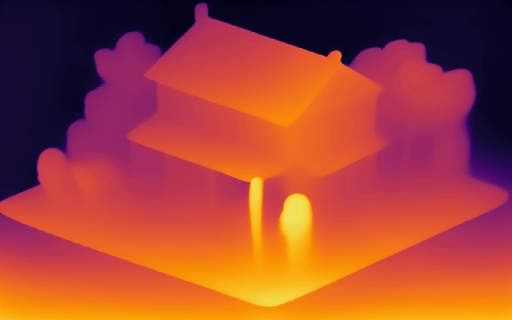}
     \end{subfigure}%
     \begin{subfigure}[b]{0.24\textwidth}
         \centering
        \includegraphics[width=\textwidth,height=0.5625\textwidth]{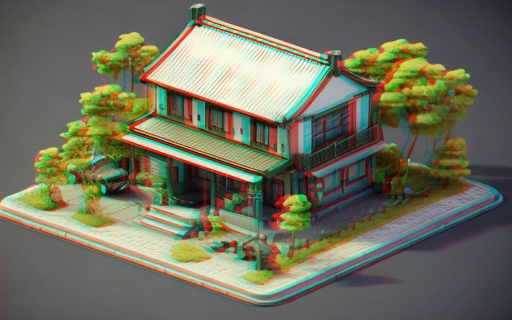}
     \end{subfigure}%
     \begin{subfigure}[b]{0.24\textwidth}
         \centering
        \includegraphics[width=\textwidth,height=0.5625\textwidth]{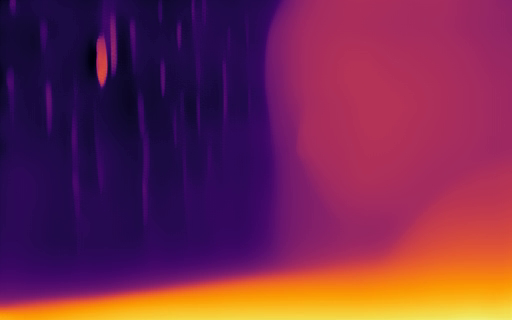}
     \end{subfigure}%
     \begin{subfigure}[b]{0.24\textwidth}
         \centering
        \includegraphics[width=\textwidth,height=0.5625\textwidth]{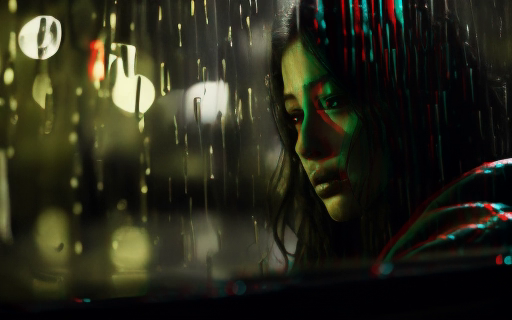}
     \end{subfigure}
     \\
     \rotatebox{90}{~~~~~~~~~~~~\footnotesize \makecell{$t=30$}}
     \begin{subfigure}[b]{0.24\textwidth}
         \centering
        \includegraphics[width=\textwidth,height=0.5625\textwidth]{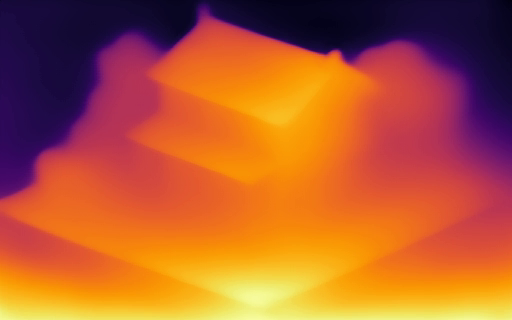}
     \end{subfigure}%
     \begin{subfigure}[b]{0.24\textwidth}
         \centering
        \includegraphics[width=\textwidth,height=0.5625\textwidth]{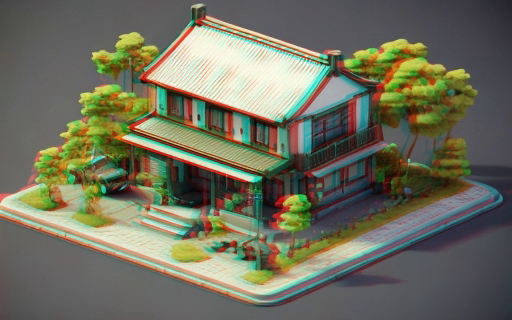}
     \end{subfigure}%
     \begin{subfigure}[b]{0.24\textwidth}
         \centering
        \includegraphics[width=\textwidth,height=0.5625\textwidth]{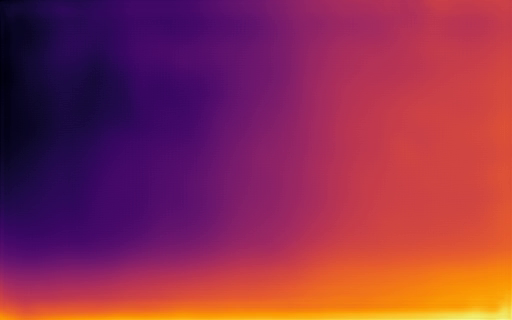}
     \end{subfigure}%
     \begin{subfigure}[b]{0.24\textwidth}
         \centering
        \includegraphics[width=\textwidth,height=0.5625\textwidth]{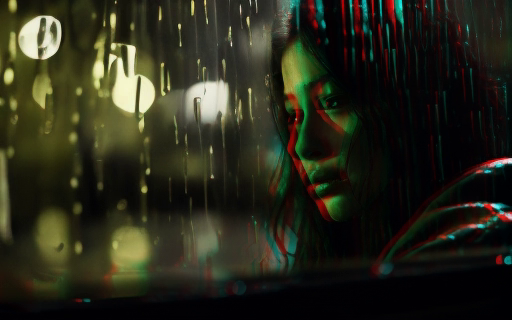}
     \end{subfigure}
     \\
     \rotatebox{90}{~~~~~~~~~~~~\footnotesize \makecell{$t=35$}}
     \begin{subfigure}[b]{0.24\textwidth}
         \centering
        \includegraphics[width=\textwidth,height=0.5625\textwidth]{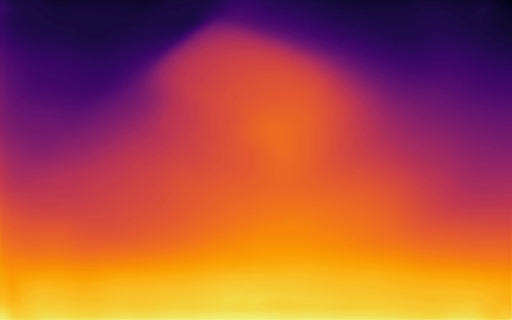}
     \end{subfigure}%
     \begin{subfigure}[b]{0.24\textwidth}
         \centering
        \includegraphics[width=\textwidth,height=0.5625\textwidth]{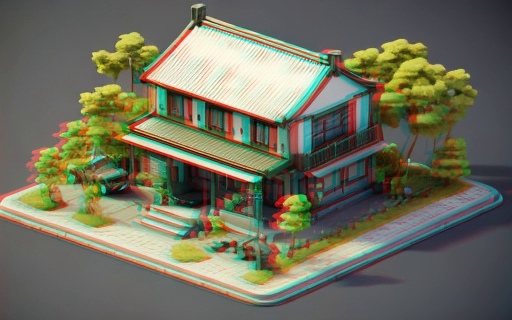}
     \end{subfigure}%
     \begin{subfigure}[b]{0.24\textwidth}
         \centering
        \includegraphics[width=\textwidth,height=0.5625\textwidth]{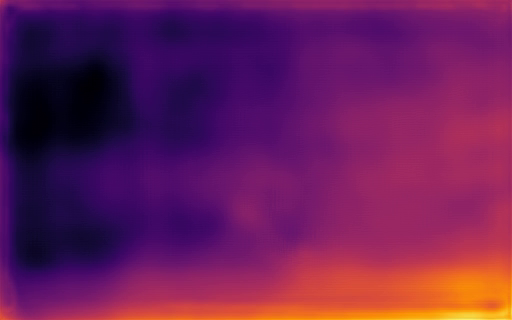}
     \end{subfigure}%
     \begin{subfigure}[b]{0.24\textwidth}
         \centering
        \includegraphics[width=\textwidth,height=0.5625\textwidth]{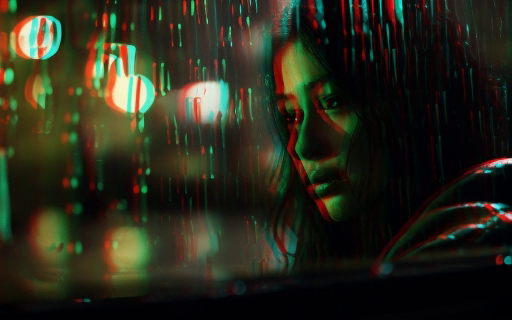}
     \end{subfigure}
     \\
     \rotatebox{90}{~~~~~~~~~\footnotesize \makecell{$t=45$}}
     \begin{subfigure}[b]{0.24\textwidth}
         \centering
        \includegraphics[width=\textwidth,height=0.5625\textwidth]{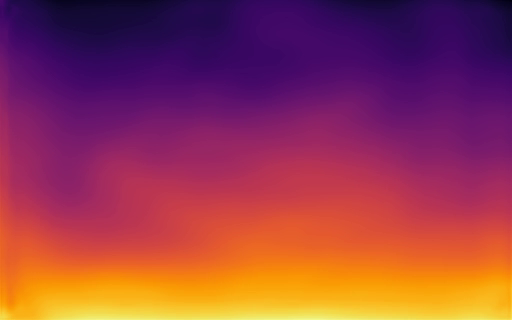}
     \end{subfigure}%
     \begin{subfigure}[b]{0.24\textwidth}
         \centering
        \includegraphics[width=\textwidth,height=0.5625\textwidth]{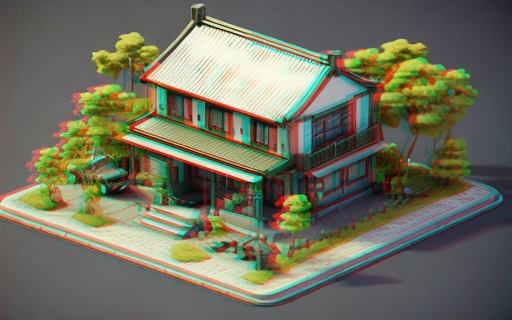}
     \end{subfigure}%
     \begin{subfigure}[b]{0.24\textwidth}
         \centering
        \includegraphics[width=\textwidth,height=0.5625\textwidth]{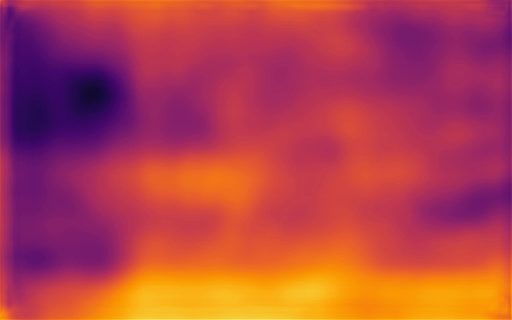}
     \end{subfigure}%
     \begin{subfigure}[b]{0.24\textwidth}
         \centering
        \includegraphics[width=\textwidth,height=0.5625\textwidth]{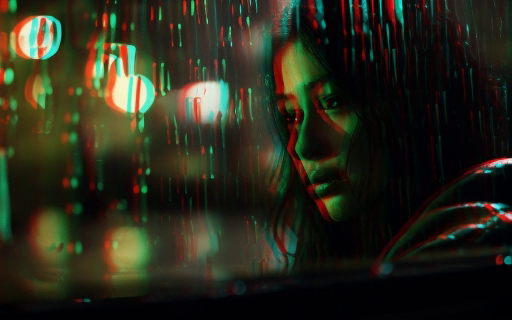}
     \end{subfigure}
    \Description{Visualization of how dissolved depth maps would affect the stereo effects. From top to bottom, we present stronger dissolving levels and their corresponding anaglyph visualization. In general, lower levels of dissolving result in smoother depth map edges without significantly altering structure. In contrast, higher dissolving levels may lead to over-smoothing, which compresses depth variations and makes object distances appear closer to one another, resulting in a stronger stereo effect. Additional examples are in~\cref{fig:dslv_res} in the supplementary material.}
    \caption{Visualization of how dissolved depth maps would affect the stereo effects. From top to bottom, we present stronger dissolving levels and their corresponding anaglyph visualization. In general, lower levels of dissolving result in smoother depth map edges without significantly altering structure. In contrast, higher dissolving levels may lead to over-smoothing, which compresses depth variations and makes object distances appear closer to one another, resulting in a stronger stereo effect. Additional examples are in~\cref{fig:dslv_res} in the supplementary material.}
    \label{fig:dslv_figure}
\end{figure*}

By varying the restart window $K$ and the number of denoising rounds $L$, we found that increasing $K$ in later sampling steps degrades performance, as shown in~\Cref{tab:restart_ablation}.
In~\Cref{tab:refinement_ablation}, by using the optimal settings ($K=6, L=7$), we vary the number of refinement rounds $N$.
This experiment shows that moderate refinement ($N=4$) achieves the best performance.

The reported ``total steps" in both tables correspond to the cumulative number of diffusion steps, which can be seen as a proxy for the runtime. Our method runs in total for 150 diffusion steps for stereo video generations, corresponding to threefold total computation compared to a common monocular video generator (\textit{e.g.} 50 steps).

\begin{figure*}
    \centering
     \setlength{\tabcolsep}{0pt}
     \renewcommand{\arraystretch}{0}
    \begin{tabular}{cc@{\hspace{2pt}}c@{\hspace{2pt}}c@{\hspace{2pt}}c}
        Left View & Right View & Left View & Right View & \multirow{1}{*}{\rotatebox{-90}{$t=0$}} \\
        \toprule
        \begin{overpic}[width=0.2\linewidth,height=0.117\linewidth]{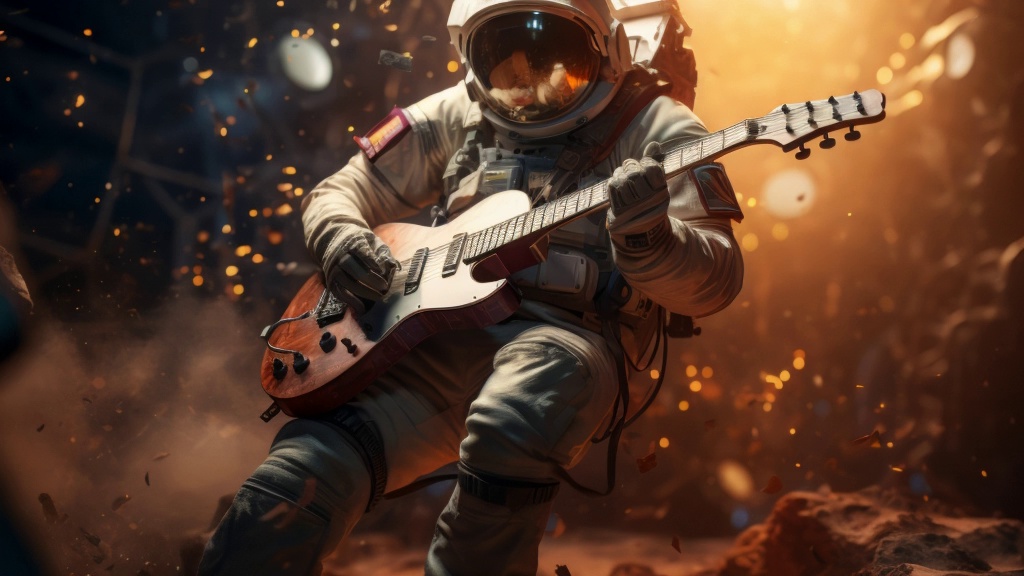}
            \put(15,1.5){\makebox[0pt]{\adjincludegraphics[width=0.05\linewidth,trim={{.29\width} {.42\height} {.58\width} {.35\height}},clip, cfbox=cyan 2pt 0cm]{misc/failed/frame_000__a_l_guitar.jpg}}}
            \put(84,30){\makebox[0pt]{\adjincludegraphics[width=0.05\linewidth,trim={{.57\width} {.6\height} {.33\width} {.22\height}},clip, cfbox=cyan 2pt 0cm]{misc/failed/frame_000__a_l_guitar.jpg}}}
        \end{overpic} & 
         \begin{overpic}[width=0.2\linewidth,height=0.117\linewidth]{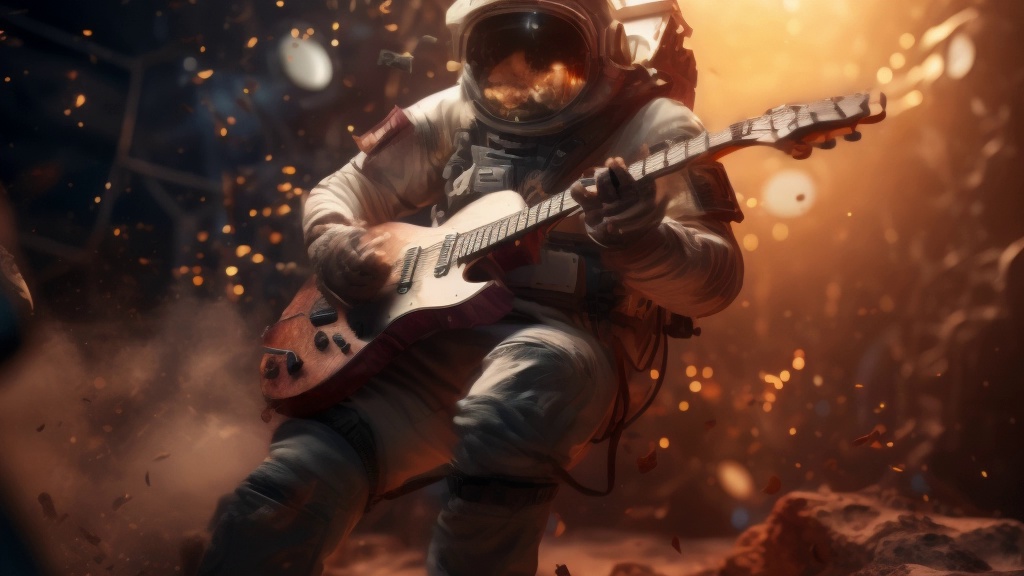}
            \put(15,1.5){\makebox[0pt]{\adjincludegraphics[width=0.05\linewidth,trim={{.27\width} {.42\height} {.6\width} {.35\height}},clip, cfbox=cyan 2pt 0cm]{misc/failed/frame_000__a_r_guitar.jpg}}}
            \put(84,30){\makebox[0pt]{\adjincludegraphics[width=0.05\linewidth,trim={{.55\width} {.6\height} {.35\width} {.22\height}},clip, cfbox=cyan 2pt 0cm]{misc/failed/frame_000__a_r_guitar.jpg}}}
        \end{overpic} &
        \begin{overpic}[width=0.2\linewidth,height=0.117\linewidth]{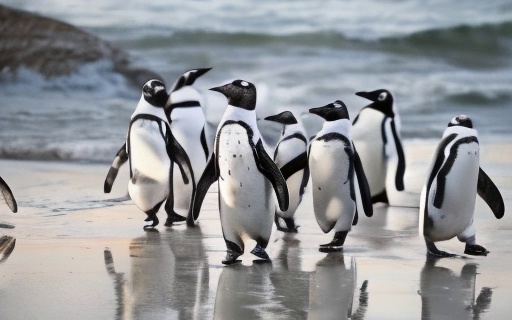}
            \put(84,35){\makebox[0pt]{\adjincludegraphics[width=0.05\linewidth,trim={{.4\width} {.62\height} {.47\width} {.2\height}},clip, cfbox=red 2pt 0cm]{misc/failed/frame_000__a_l_penguin.jpg}}}
        \end{overpic} & 
         \begin{overpic}[width=0.2\linewidth,height=0.117\linewidth]{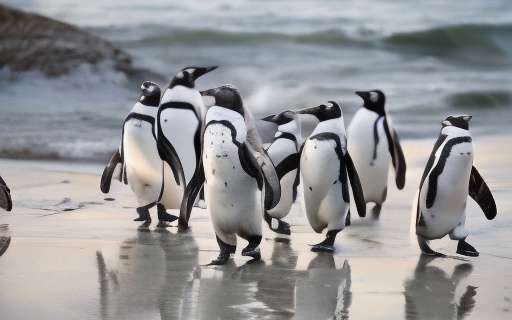}
            \put(84,35){\makebox[0pt]{\adjincludegraphics[width=0.05\linewidth,trim={{.38\width} {.62\height} {.49\width} {.2\height}},clip, cfbox=red 2pt 0cm]{misc/failed/frame_000__a_r_penguin.jpg}}}
        \end{overpic} & \multirow{1}{*}{\rotatebox{-90}{$t=7$}} \\
        \begin{overpic}[width=0.2\linewidth,height=0.117\linewidth]{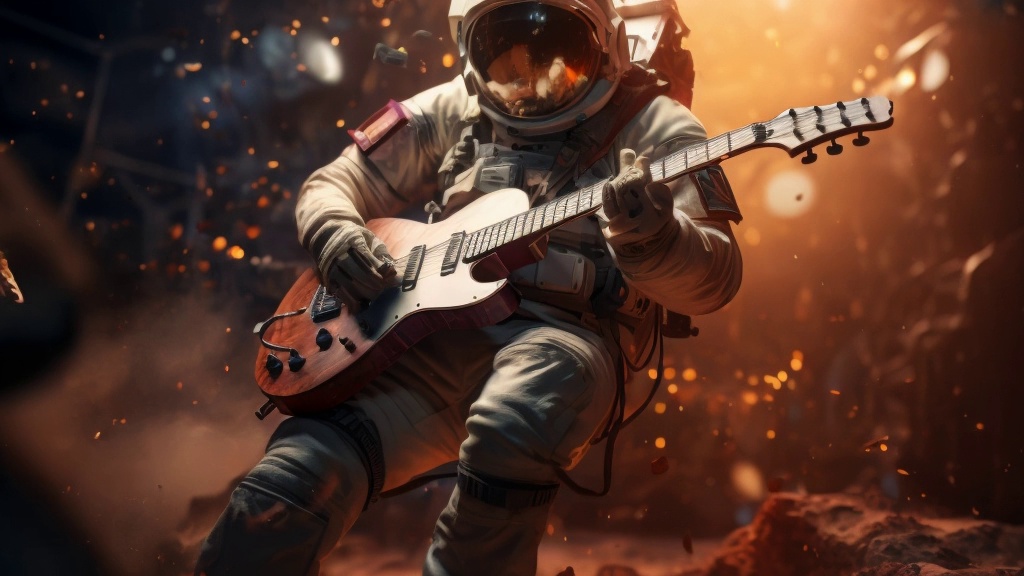}
            \put(15,1.5){\makebox[0pt]{\adjincludegraphics[width=0.05\linewidth,trim={{.29\width} {.42\height} {.58\width} {.35\height}},clip, cfbox=cyan 2pt 0cm]{misc/failed/frame_007__a_l_guitar.jpg}}}
            \put(84,30){\makebox[0pt]{\adjincludegraphics[width=0.05\linewidth,trim={{.57\width} {.6\height} {.33\width} {.22\height}},clip, cfbox=cyan 2pt 0cm]{misc/failed/frame_007__a_l_guitar.jpg}}}
        \end{overpic} & 
         \begin{overpic}[width=0.2\linewidth,height=0.117\linewidth]{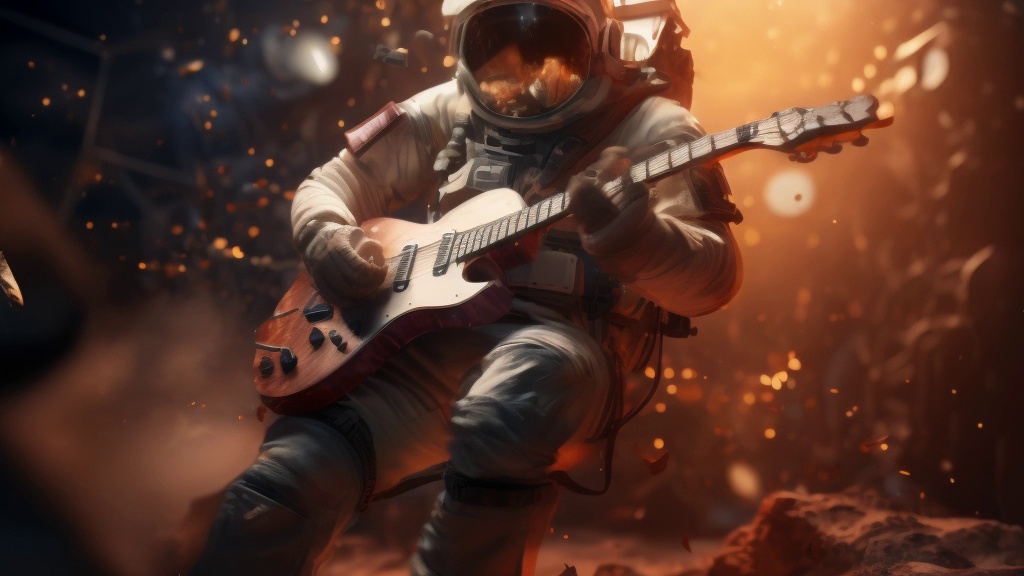}
            \put(15,1.5){\makebox[0pt]{\adjincludegraphics[width=0.05\linewidth,trim={{.27\width} {.42\height} {.6\width} {.35\height}},clip, cfbox=cyan 2pt 0cm]{misc/failed/frame_007__a_r_guitar.jpg}}}
            \put(84,30){\makebox[0pt]{\adjincludegraphics[width=0.05\linewidth,trim={{.55\width} {.6\height} {.35\width} {.22\height}},clip, cfbox=cyan 2pt 0cm]{misc/failed/frame_007__a_r_guitar.jpg}}}
        \end{overpic} &
        \begin{overpic}[width=0.2\linewidth,height=0.117\linewidth]{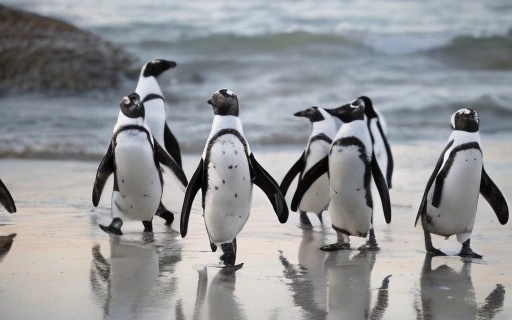}
            \put(84,35){\makebox[0pt]{\adjincludegraphics[width=0.05\linewidth,trim={{.38\width} {.62\height} {.49\width} {.2\height}},clip, cfbox=red 2pt 0cm]{misc/failed/frame_007__a_l_penguin.jpg}}}
        \end{overpic} & 
         \begin{overpic}[width=0.2\linewidth,height=0.117\linewidth]{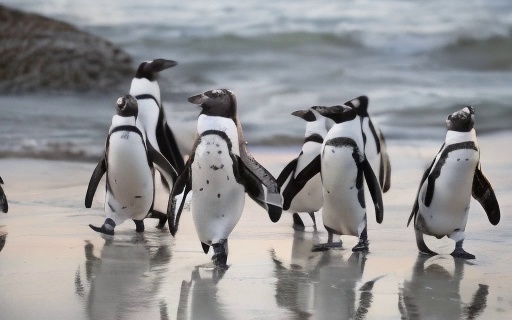}
            \put(84,35){\makebox[0pt]{\adjincludegraphics[width=0.05\linewidth,trim={{.36\width} {.62\height} {.51\width} {.2\height}},clip, cfbox=red 2pt 0cm]{misc/failed/frame_007__a_r_penguin.jpg}}}
        \end{overpic} & \multirow{1}{*}{\rotatebox{-90}{$t=14$}} \\
        \begin{overpic}[width=0.2\linewidth,height=0.117\linewidth]{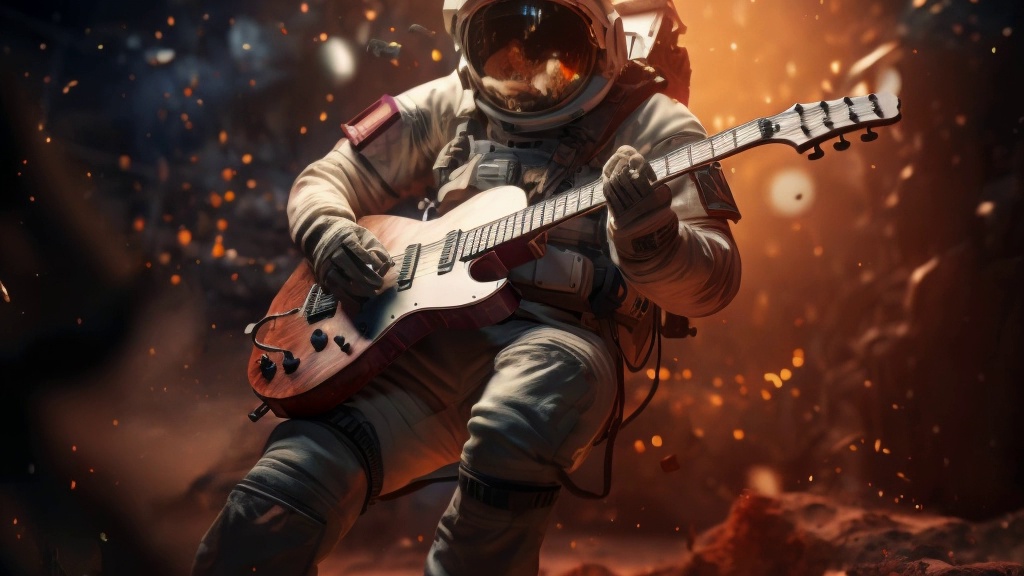}
            \put(15,1.5){\makebox[0pt]{\adjincludegraphics[width=0.05\linewidth,trim={{.29\width} {.42\height} {.58\width} {.35\height}},clip, cfbox=cyan 2pt 0cm]{misc/failed/frame_014__a_l_guitar.jpg}}}
            \put(84,30){\makebox[0pt]{\adjincludegraphics[width=0.05\linewidth,trim={{.57\width} {.6\height} {.33\width} {.22\height}},clip, cfbox=cyan 2pt 0cm]{misc/failed/frame_014__a_l_guitar.jpg}}}
        \end{overpic} & 
         \begin{overpic}[width=0.2\linewidth,height=0.117\linewidth]{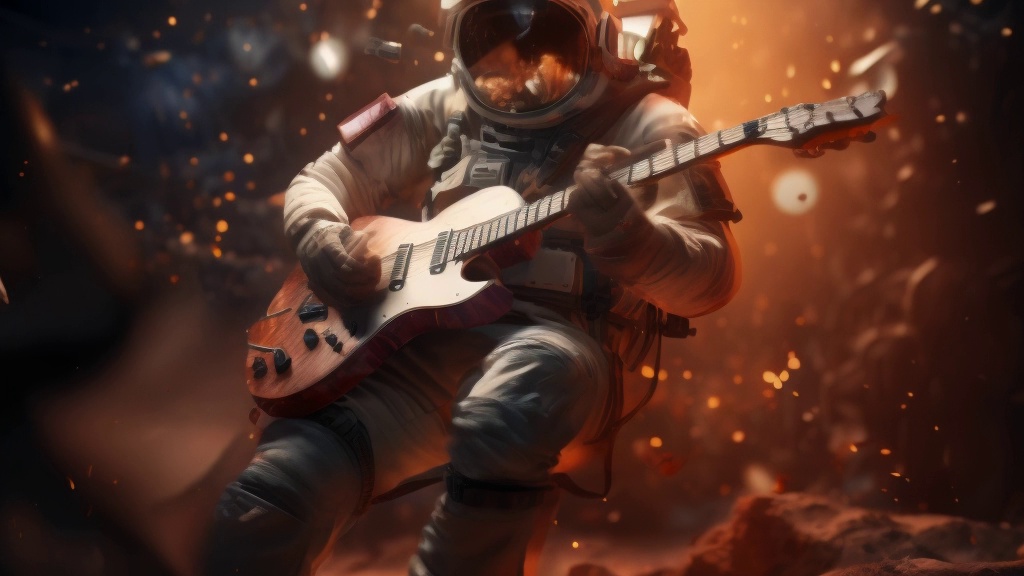}
            \put(15,1.5){\makebox[0pt]{\adjincludegraphics[width=0.05\linewidth,trim={{.27\width} {.42\height} {.6\width} {.35\height}},clip, cfbox=cyan 2pt 0cm]{misc/failed/frame_014__a_r_guitar.jpg}}}
            \put(84,30){\makebox[0pt]{\adjincludegraphics[width=0.05\linewidth,trim={{.55\width} {.6\height} {.35\width} {.22\height}},clip, cfbox=cyan 2pt 0cm]{misc/failed/frame_014__a_r_guitar.jpg}}}
        \end{overpic} &
        \begin{overpic}[width=0.2\linewidth,height=0.117\linewidth]{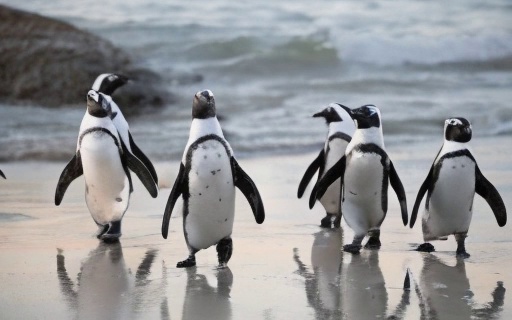}
            \put(84,35){\makebox[0pt]{\adjincludegraphics[width=0.05\linewidth,trim={{.36\width} {.62\height} {.51\width} {.2\height}},clip, cfbox=red 2pt 0cm]{misc/failed/frame_014__a_l_penguin.jpg}}}
        \end{overpic} & 
         \begin{overpic}[width=0.2\linewidth,height=0.117\linewidth]{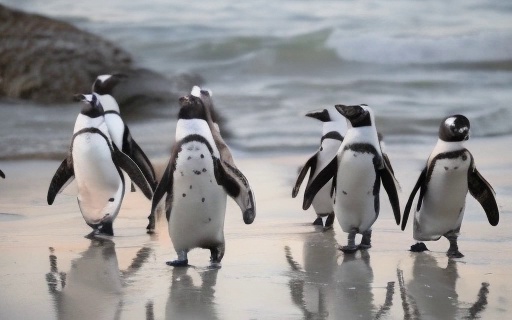}
            \put(84,35){\makebox[0pt]{\adjincludegraphics[width=0.05\linewidth,trim={{.34\width} {.62\height} {.53\width} {.2\height}},clip, cfbox=red 2pt 0cm]{misc/failed/frame_014__a_r_penguin.jpg}}}
        \end{overpic}
    \end{tabular}
    \Description{Demonstration of the failure cases. Our method can fail in the strong motion areas, such as the rapidly moving hand gesture and head pose.}
    \caption{Demonstration of the failure cases. Our method can fail in the strong motion areas, such as the rapidly moving hand gesture and head pose.}
    \label{fig:failed}
\end{figure*}

\subsubsection{Sensitivity to Baseline Distances}

We evaluate the sensitivity of our method to different baseline distances between the left and right views, expressed relative to the image width. We observe that smaller baselines ($<10\%$) led to a generally stable spatial and temporal consistency around occlusion boundaries or inpainting regions, while larger baselines ($>10\%$) may result in a greater change of distortion, especially on complex scenes. We present image samples with a stronger baseline value in~\Cref{fig:baseline_16}. In addition, statistical results are provided in~\Cref{tab:baseline_sensitivity}. The results indicate that with an increased baseline, both the semantic and multi-view consistency will be slightly dropped.

\begin{table}[b]
    \small
    \centering
    \setlength{\tabcolsep}{2pt}
    \caption{Performances with different depth models. Without the depth dissolving technique, all depth estimation models achieve similar results.}
    \label{tab:depth_abls}
    \begin{tabular}{c|cccc|c}
        \toprule
        & D. Pro &  D. Anything &  V. D. Anything &  \makecell{D. Crafter\\w/o dsl.} & \makecell{D. Crafter \\ w/ dsl.}\\
        \midrule
        MEt3R $\downarrow$ & 6.78 & 6.71 & 6.79 & 6.70 & \textbf{4.95}\\
        \bottomrule
    \end{tabular}
\end{table}

\subsection{Discussion}
\label{sec:evaluation}

\subsubsection{Why Depthcrafter?}
As the essential ingredient for creating stereoscopic effects, depth information significantly affects the quality of the generated right views.
Our method is compatible with any depth estimation method.
Besides DepthCrafter, we evaluated our method using several state-of-the-art depth models, including Depth Pro~\cite{bochkovskii2024depth}, Depth Anything~\cite{yang2024depth}, and Video Depth Anything~\cite{chen2025video}, to assess the impact of depth estimation accuracy on stereo generation.
For image-based depth models (\textit{i.e.}, DepthPro, DepthAnything), we applied a disparity propagation algorithm (see supplementary) for obtaining temporally consistent video depth.
For \textit{DepthCrafter}, depth maps can be further processed using our \emph{depth dissolving} technique, implemented as a reverse diffusion process in the depth latent space following the pretrained DepthCrafter architecture.
As shown in~\Cref{tab:depth_abls}, without the depth dissolving technique, similar performances can be observed for different depth estimation models. However, depth dissolving gives a clear advantage to Depthcrafter ( $26\%$ improvement on MEt3R from $6.70$ to $4.95$).
Visualizations of varying dissolving levels are provided in~\Cref{fig:dslv_figure}.\\

\subsubsection{Noise-Injection For Latent Refinement.}

Geometric operations such as warping are necessary for stereo generation, yet they inherently push latents away from the model prior. Without proper hole filling, the warped latents introduce distributional shifts that degrade generation quality. We argue that the missing regions are not merely spatial holes, but require stereo-consistent temporal and semantic completion. This is why inpainting-based approaches, including ProPainter (RGB-inpainting) and SVG (latent-inpainting), fail even when the missing regions are small, despite appearing superficially capable. Thus, our method achieves better results by filling holes using the same model prior, preserving distributional consistency throughout the generation process.

\paragraph{Noisy Restart}
Our noise injection strategies are conceptually related to the noise re-injection mechanism introduced in Time Reversal Fusion (TRF)~\cite{feng2024explorative}. TRF adopts a global noise perturbation strategy, whereas we use a stereo-aware, region-selective noise injection.
Our experiments indicate that stronger noise injection during noisy restart consistently produces a clearer and more stable stereo effect, as shown in~\Cref{fig:noisy_restart}. Yet both approaches reach a similar conclusion, that small perturbations have minimal effect at early denoising stages.
This behavior is well grounded in diffusion dynamics~\cite{ho2020denoising}, where the variance term $\beta_t$ shrinks toward later timesteps, reverse-process updates become too small to correct earlier structural choices. 
As a result, we apply stronger, stereo-targeted noisy restart in the early timesteps, ensuring that the model converges toward a stable and geometry-consistent solution.

\paragraph{Iterative Refinement}
As observed in Repaint~\cite{lugmayr2022repaint}, image inpainting models require sufficient diffusion steps to fully harmonize conditional inputs before the noise variance $\beta_t$ becomes too small in later timesteps. Otherwise, the model loses the capacity to correct structural inconsistencies. We observe an analogous phenomenon in our video generation method.
Warping depth maps at earlier diffusion stages only is better than warping depth maps at all diffusion stages. 
It is intuitive that a warping operation alters the distribution of RGB video latents to a distribution that is slightly out of the domain for the diffusion model.
Therefore, there is a trade-off between warping high-frequency details at later diffusion stages and moving latents outside the expected distribution.
We adopt early-stopping for depth warping. As mentioned in~\Cref{sec:exp_setup}, we disable the depth-based warping in the final 15 steps.

\subsubsection{Limitations.}

Our method exhibits strong robustness for videos at 256 and 512 resolutions, requiring minimal hyperparameter tuning. However, it may fail with high-resolution videos involving small objects under strong motion. In such cases, rapid fluctuations in the generated depth maps can cause instability in the latent space due to excessive and complex warping. To address this, the incorporated dissolved depth maps can help smooth artifacts and enhance overall performance. Nonetheless, this strategy demands careful tuning of dissolving levels (e.g., a stronger setting of 40 is recommended) and does not fully eliminate the issue. Residual inconsistencies in depth and semantic accuracy may persist, particularly in cases of extreme motion or occlusion, potentially leading to incorrect interpretations of depth and semantic information, as shown in~\Cref{fig:failed}. 

\begin{center}
\begin{tabular}{c}
     ~ \\
\end{tabular}

\begin{minipage}{0.7\linewidth}
    \centering
    \begin{overpic}[width=.8\linewidth]{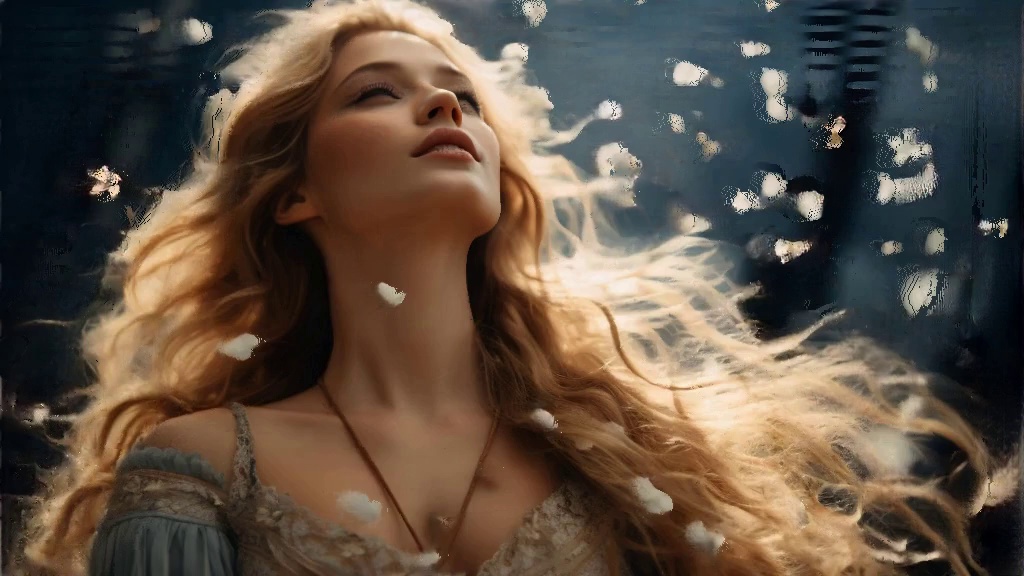}
        \put(100,18.5){\makebox[0pt]{\adjincludegraphics[height=0.35\linewidth,trim={{.71\width} {.65\height} {.10\width} {.05\height}},clip, cfbox=purple 2pt 0cm]{misc/main_figure/propainter/a_beautiful_woman_with_long_hair_and_a_d_masked_000.jpg}}}
        \put(-5,8.5){\makebox[0pt]{\adjincludegraphics[height=0.40\linewidth,trim={{.06\width} {.6\height} {.73\width} {.0\height}},clip, cfbox=purple 2pt 0cm]{misc/main_figure/propainter/a_beautiful_woman_with_long_hair_and_a_d_masked_000.jpg}}}
    \end{overpic}
    \setlength{\abovecaptionskip}{1.5em}
    \setlength{\belowcaptionskip}{1em}
    \Description{Artifacts generated by ProPainter.}
    \captionsetup{hypcap=false}
    \captionof{figure}{Artifacts generated by ProPainter.}
    \captionsetup{hypcap=true} 
    \label{fig:propainter_failure}
\end{minipage}
\end{center}

The observation from our user study and benchmark results is consistent with~\cite{tamir2024makes}, which reveals that preferences in VR often differ from those observed on traditional screens. For example, though \textit{StereoCrafter} achieves better epistolary consistency, \textit{TrajectoryCrafter} delivers better stereo effects.
In addition, \textit{ProPainter} generates a great amount of artifacts on the occluded regions as shown in~\Cref{{fig:propainter_failure}}, which we anticipate would perform poorly in user studies. However, when viewed in stereo (e.g., with both eyes), many viewers did not notice the strong frame quality degradation of the right view.
This discrepancy underscores the importance of developing evaluation metrics specifically tailored to immersive VR experiences, ensuring they accurately reflect perceptual feedback rather than relying on screen-based metrics.

\section{Conclusion}

This work introduced \textit{DissolveStereo}, a novel zero-shot stereo video generation approach. \textit{DissolveStereo} incorporates a \textit{noisy restart} strategy for stereo-aware latent initialization and an \textit{iterative refinement} process to enhance latent consistency. More importantly, we proposed \textit{dissolved depth maps} that retain only low-frequency structural depth, reducing high-frequency noise and improving the coherence and stability of the stereoscopic effects.
Our comprehensive evaluations, including statistical analysis and user studies, demonstrate the effectiveness of our method in generating high-quality stereo videos with enhanced depth consistency and temporal smoothness.
While noisy restart and iterative refinement are tailored to diffusion-style denoising, we believe dissolved depth maps are more general and can be extended beyond diffusion models to a broader class of approaches.
Future research will focus on developing a method for determining the optimal dissolving level and exploring the incorporation of user guidance for personalized control over the generated stereo videos.


\bibliographystyle{ACM-Reference-Format}
\bibliography{software}

\clearpage
\setcounter{page}{1}

\begin{figure*}
     \centering
     \setlength{\tabcolsep}{0pt}
     \renewcommand{\arraystretch}{0}
     \adjustboxset{
         width=16pc,
          margin=1pt, valign=t  
          }
     \begin{tabular}{c @{\hspace{2pt}}|@{\hspace{2pt}}ccc@{\hspace{2pt}}|@{\hspace{2pt}}c}
         Starting Frame {\color{gray} \footnotesize (Input)} & \multicolumn{3}{c|@{\hspace{2pt}}}{Generated Interpolated Frames {\color{gray} \footnotesize (Output)}} & Ending Frame {\color{gray} \footnotesize (Input)} \\
         \toprule
         \multirow[t]{2}{*}{
            \makecell[t]{
             \includegraphics[width=.19\textwidth]{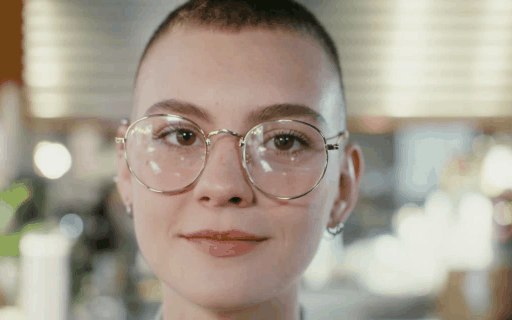} \\
             \textit{\color{gray} \footnotesize Prompt: a smiling girl \hfill}
             }
         }
         &
         \includegraphics[width=.19\textwidth]{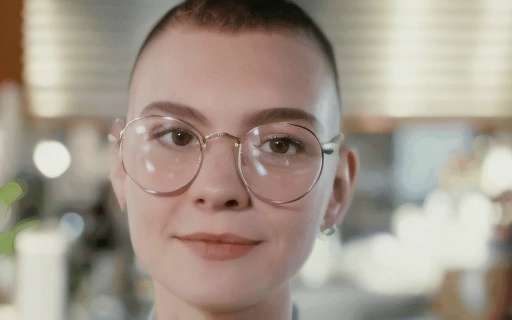} &
         \includegraphics[width=.19\textwidth]{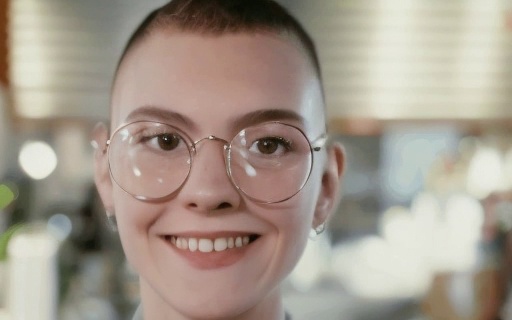} &
         \includegraphics[width=.19\textwidth]{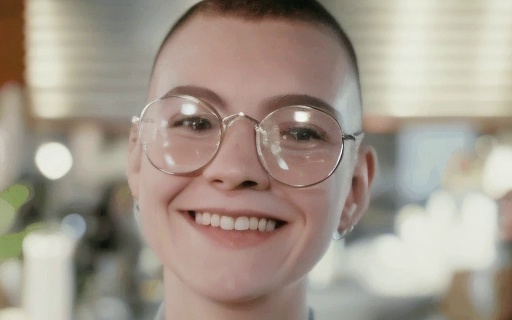} &
         \includegraphics[width=.19\textwidth]{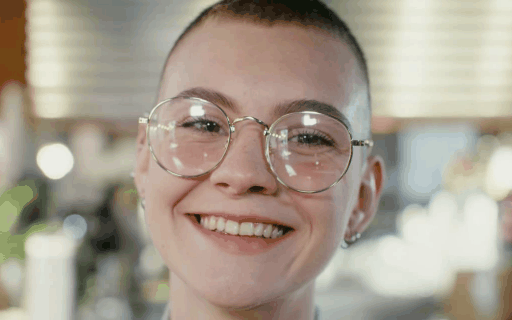} \\
          &
         \includegraphics[width=.19\textwidth]{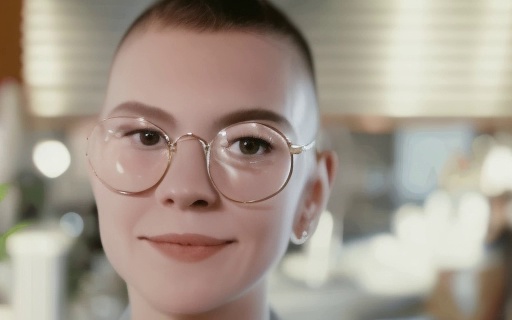} &
         \includegraphics[width=.19\textwidth]{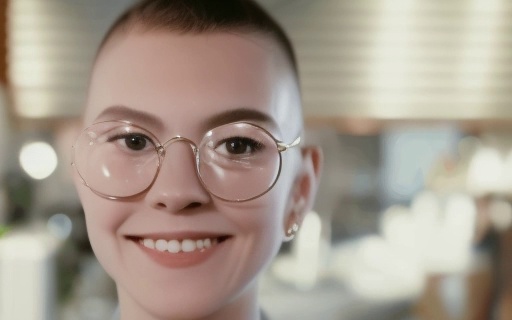} &
         \includegraphics[width=.19\textwidth]{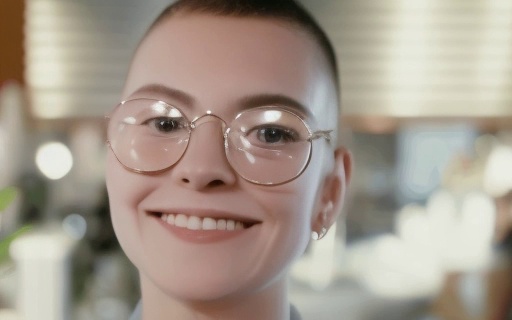} \\
          \\
        \midrule
         \multirow[t]{2}{*}{
            \makecell[t]{
             \includegraphics[width=.19\textwidth]{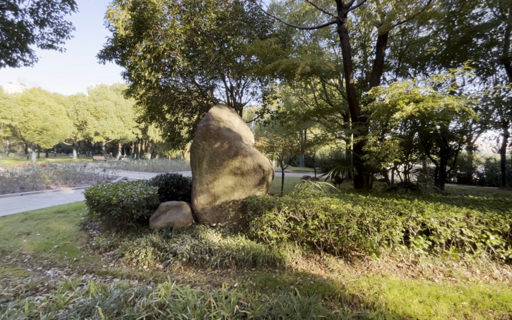} \\
             \textit{\color{gray} \footnotesize Prompt: rotating view \hfill}
             }
         }
         
         &
         \includegraphics[width=.19\textwidth]{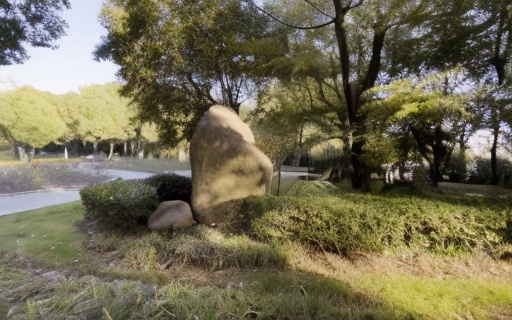} &
         \includegraphics[width=.19\textwidth]{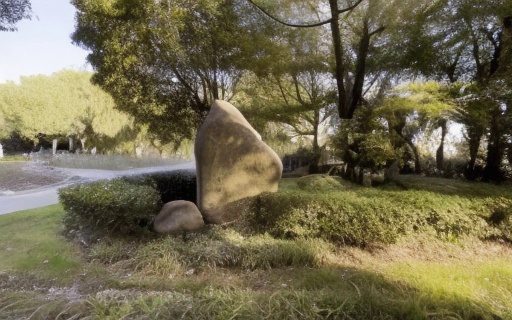} &
         \includegraphics[width=.19\textwidth]{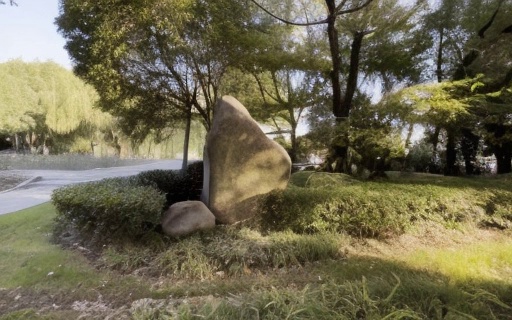} &
         \includegraphics[width=.19\textwidth]{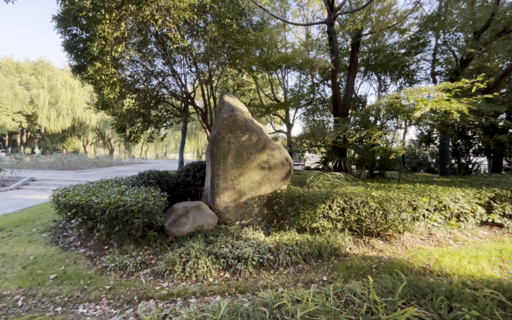} \\
         &
         \includegraphics[width=.19\textwidth]{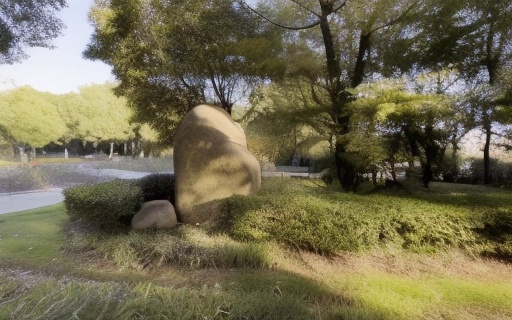} &
         \includegraphics[width=.19\textwidth]{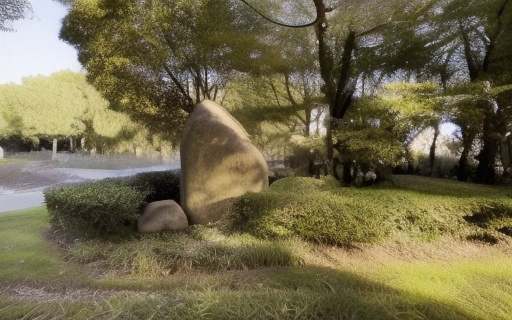} &
         \includegraphics[width=.19\textwidth]{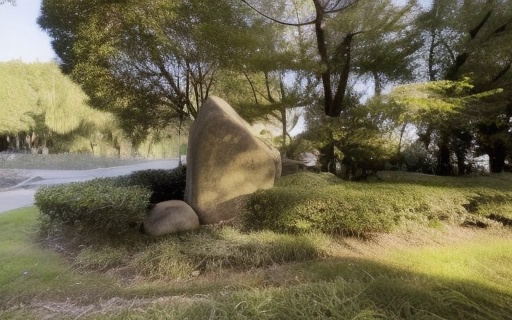} \\
          & 
          \\
          \midrule
         $t=0$ & $t=1$ & $t=8$ & $t=14$ & $t=15$ \\
     \end{tabular}
    \caption{Demonstration of generating stereo videos with interpolated inputs. We input the starting and ending frames as our input and generate stereo content with our method. For each example, top row is the left view and bottom row is the right view.}
    \label{fig:interp}
\end{figure*}

\section{An Efficient Warping Algorithm}
We present a vectorization technique for efficiently implementing pixel-wise image transformations that involve pixel displacement maps. Traditional implementations of such transformations rely on nested loops to process each pixel individually, which can be computationally intensive and inefficient, especially for high-resolution images or large batches.
Our method addresses these inefficiencies by leveraging advanced tensor operations in PyTorch to vectorize the computations while ensuring the correctness of the output, particularly in handling overlapping pixel assignments and maintaining the original assignment order.

\noindent\textbf{Problem Formulation.}
Let $I \in \mathbb{R}^{B \times C \times H \times W}$ be a batch of input images, where $B$ is the batch size, $C$ is the number of channels, and $H$ and $W$ are the height and width of the images, respectively. Let $D \in \mathbb{R}^{B \times 1 \times H \times W}$ be the corresponding batch of depth maps or optical flow fields. The goal is to create a derived image $I'$ by displacing pixels from the input image $I$ based on the values in $D$, scaled by a factor $s$ and an exponent $\alpha$. This transformation can be formalized as:
\begin{equation}
I'_{b, c, h, w'} = I_{b, c, h, w}, \quad \text{where} \quad w' = w + \text{offset}_{b, h, w},
\end{equation}
\begin{equation}
\text{offset}_{b, h, w} = \left( D_{b, 0, h, w}^\alpha \times s \right).
\end{equation}
The challenge arises when multiple source pixels $(b, c, h, w)$ map to the same destination pixel $(b, c, h, w')$, leading to overlapping assignments. It is crucial to ensure that the final value of $I'$ at any pixel $(b, c, h, w')$ corresponds to the last assignment in the original processing order, consistent with the behavior of the nested loops.

\noindent\textbf{Baseline Implementation.}
The traditional implementation uses three nested loops over the batch, height, and width dimensions:

{\small
\begin{verbatim}
for batch in range(B):
    for row in range(H):
        for col in range(W):
            col_d = col + D[
                batch, 0, row, col] ** alpha * s
            if 0 <= col_d < W:
                I_prime[batch, :, row, col_d] =
                    I[batch, :, row, col]
\end{verbatim}
}

This approach ensures correct handling of overlapping assignments by processing pixels in a specific order (e.g., from right to left when $s \geq 0$) and overwriting previous values as needed. However, it is computationally inefficient due to the explicit loops.

\noindent\textbf{Proposed Vectorization Technique.}
Our method eliminates the explicit loops by vectorizing operations across the batch, height, and width dimensions using PyTorch tensors. The key challenge is to replicate the assignment order of the nested loops to correctly handle overlapping assignments. First, we create tensors representing the batch, row, and column indices for all pixels. For positive movement, such as $s \geq 0$, we process columns from right to left ($W - 1$ to $0$), while we process columns from left to right ($0$ to $W - 1$) for $s < 0$. For each column index $col$, we then compute the offset $D$ for each pixel and calculate the destination column indices $col_d=col + D$. Next, a Boolean mask is applied to filter out indices where $col_d$ is outside the image boundaries. Note that the order for all valid indices is reversed to prioritize the last assignments in the original loop order. This conditional reversal ensures that, when duplicates are removed, the last assignment in the original loop order is preserved, correctly handling overlapping assignments. Next, we compute unique destination indices and keep the first occurrence, which corresponds to the last assignment due to reversal. We can then simply apply the computed index map to assign all pixels in bulk.

\begin{table}[h]
    \centering
    \small
    \caption{Performance evaluation of our vectorized method for displacement handling. Our implementation speeds up the operation by 936 and 2022 times for the image resolution of 128 and 512, respectively.}
    \label{tab:exp_vec}
    \begin{tabular}{c|cc}
        \toprule
         Resolution & $128\times 128$ & $512\times 512$ \\
         \midrule
        Baseline (ms) & 1401.18 & 22651.53 \\
        Ours \hfill (ms) & {\color{magenta} ~~~~~~1.47} & {\color{magenta} ~~~~~~11.78} \\
        \bottomrule
    \end{tabular}
\end{table}

\noindent\textbf{Experimental Validation.}
We validate our method by comparing the outputs and execution times of the original and vectorized implementations, as shown in \Cref{tab:exp_vec}. The values are computed with $20$ executions for each run. The results show that our implementation offers an effective solution for efficiently implementing pixel-wise image transformations that involve displacement based on depth maps or optical flow fields.

\section{Additional Experiments}

\subsection{Multi-Resolution Support}
\begin{figure}[h]
    \centering
     \setlength{\tabcolsep}{0pt}
     \renewcommand{\arraystretch}{0.5}
    \begin{tabular}{cccc}
         Input & $t=0$ & $t=7$ & $t=15$  \\
         \includegraphics[width=0.24\linewidth,height=0.24\linewidth]{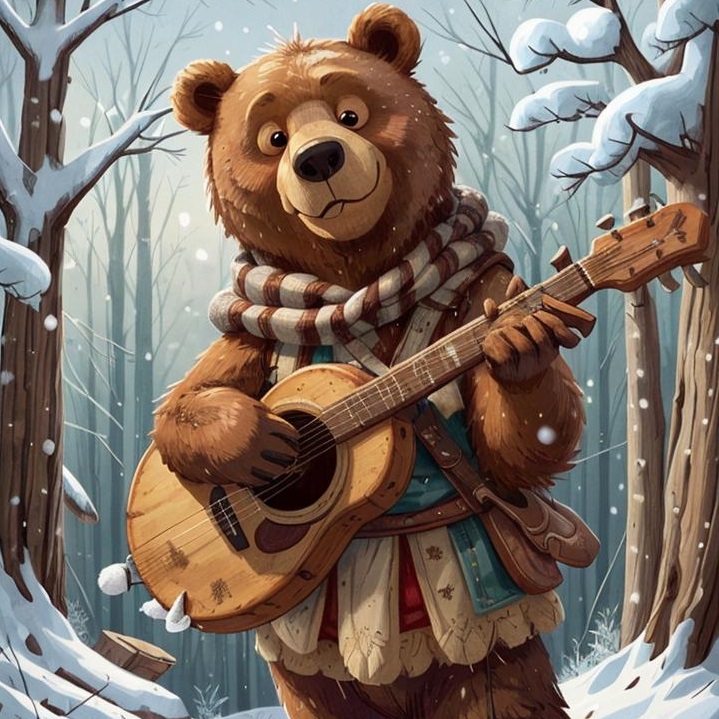}~ &
         \includegraphics[width=0.24\linewidth,height=0.24\linewidth]{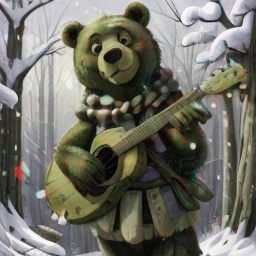} &
         \includegraphics[width=0.24\linewidth,height=0.24\linewidth]{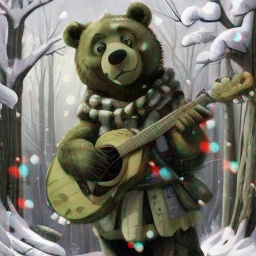} &
         \includegraphics[width=0.24\linewidth,height=0.24\linewidth]{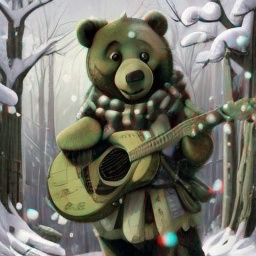} 
         \\
         \multicolumn{4}{l}{\color{gray}\scriptsize ~~Prompt: bear playing guitar happily, snowing}
         \\
         \includegraphics[width=0.24\linewidth,height=0.24\linewidth]{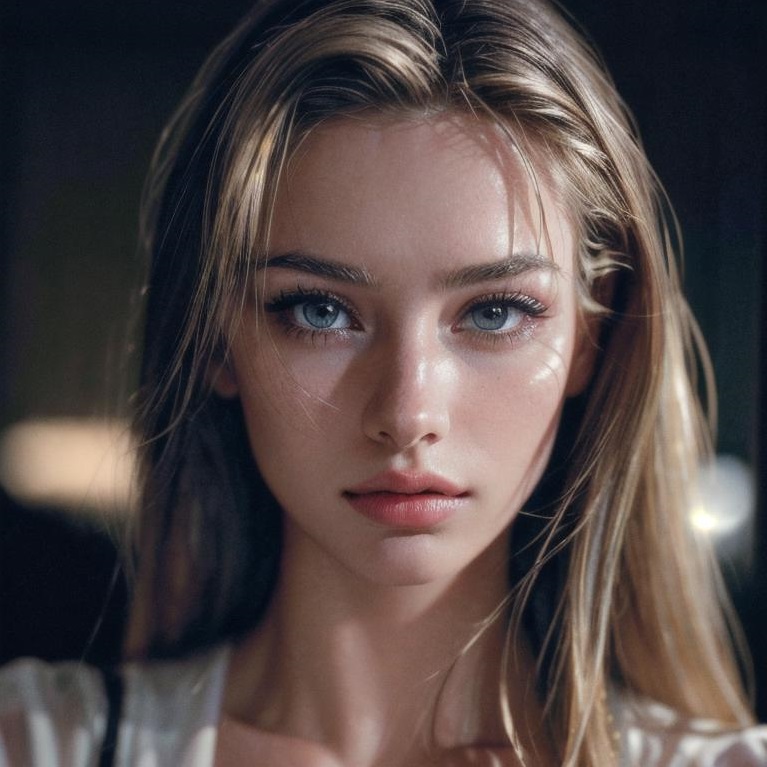}~ &
         \includegraphics[width=0.24\linewidth,height=0.24\linewidth]{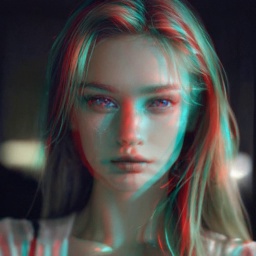} &
         \includegraphics[width=0.24\linewidth,height=0.24\linewidth]{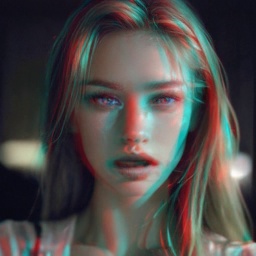} &
         \includegraphics[width=0.24\linewidth,height=0.24\linewidth]{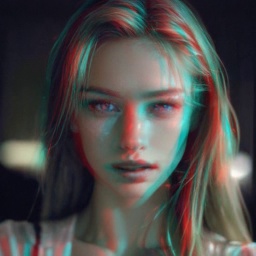} 
         \\
         \multicolumn{4}{l}{\color{gray}\scriptsize ~~Prompt: girl talking and blinking}
    \end{tabular}
    \caption{Demonstration of $256\times 256$ examples. Shown in anaglyph for visualizing the stereoscopic effect.}
    \label{fig:256}
\end{figure}

Our method supports various resolutions, such as $576\times 1024$, $320\times 512$, and $256\times 256$.
The images shown in our main text and supplementary materials are $576\times 1024$ and $320\times 512$ samples. In addition, we show $256\times 256$ samples in~\Cref{fig:256} for demonstration purposes.
By default, we recommend using dissolving strength of $t=45$ for $576\times 1024$ videos with $34$ as the last warping step, $t=40$ for $320\times 512$ videos with $44$ as the last warping step. For $256\times 256$ resolutions, we found the dissolving is no longer required to obtain good results.

\begin{figure}
    \centering
     \setlength{\tabcolsep}{0pt}
     \renewcommand{\arraystretch}{0}
    \begin{tabular}{ccc@{\hspace{2pt}}c}
        Input Image & Left View & Right View & \multirow{1}{*}{\rotatebox{-90}{$t=0$}} \\
        \toprule
         \multirow[t]{2}{*}{
            \makecell[t]{\includegraphics[width=0.27\linewidth,height=0.16\linewidth]{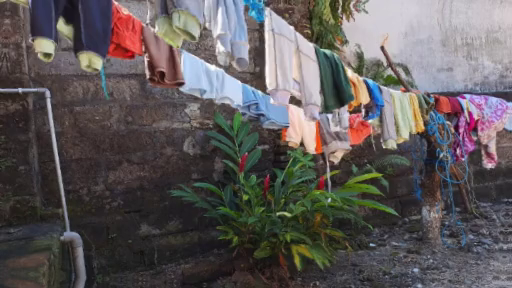} \\
             \textit{\color{gray} \scriptsize Prompt: clothes swaying} \\  \textit{\color{gray} \scriptsize in the wind} \hfill
             }
         } &
         \includegraphics[width=0.27\linewidth,height=0.16\linewidth]{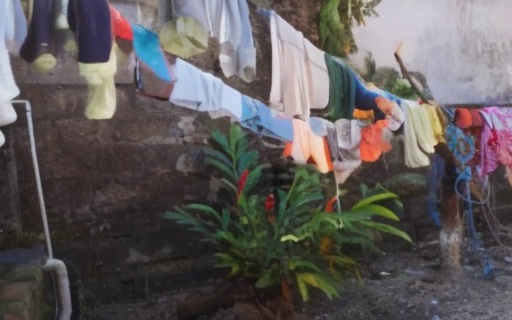} & 
         \includegraphics[width=0.27\linewidth,height=0.16\linewidth]{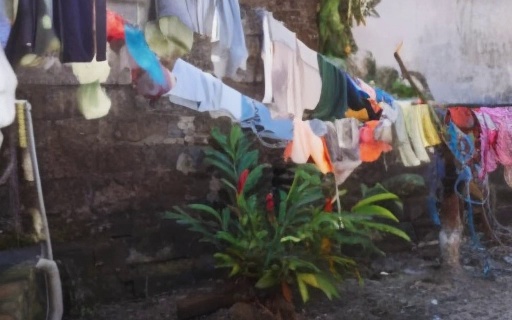} & \multirow{1}{*}{\rotatebox{-90}{$t=7$}} \\
         & 
         \includegraphics[width=0.27\linewidth,height=0.16\linewidth]{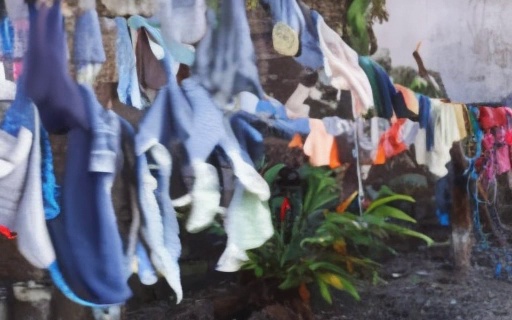} & 
         \includegraphics[width=0.27\linewidth,height=0.16\linewidth]{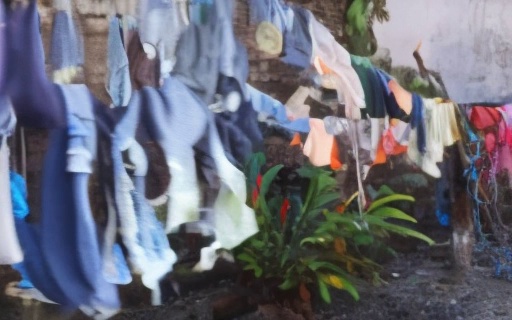} & \multirow{1}{*}{\rotatebox{-90}{$t=14$}} \\
         &
         \includegraphics[width=0.27\linewidth,height=0.16\linewidth]{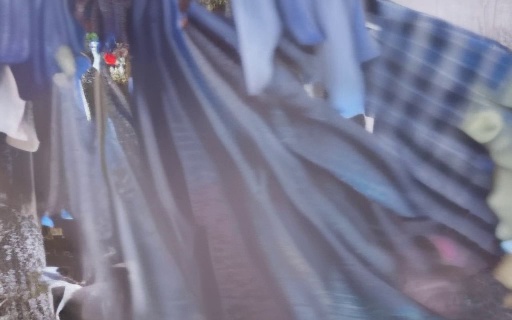} & 
         \includegraphics[width=0.27\linewidth,height=0.16\linewidth]{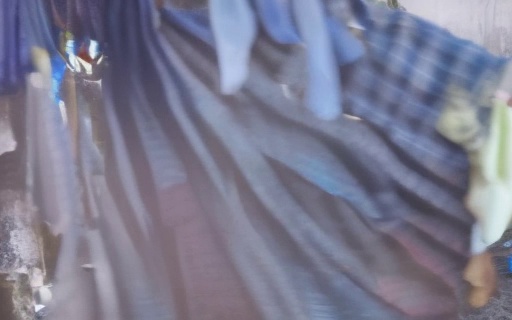} & \multirow{1}{*}{\rotatebox{-90}{$t=0$}} \\
         
         \multirow[t]{2}{*}{
            \makecell[t]{\includegraphics[width=0.27\linewidth,height=0.16\linewidth]{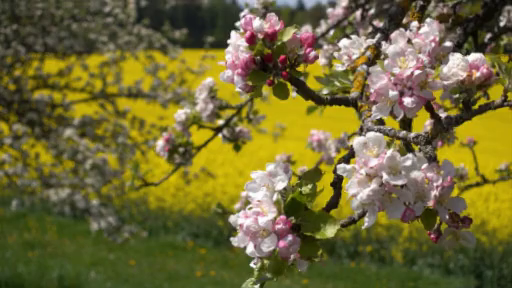} \\
             \textit{\color{gray} \scriptsize Prompt: flowers swaying} \\  \textit{\color{gray} \scriptsize in the wind} \hfill
             }
         } &
         \includegraphics[width=0.27\linewidth,height=0.16\linewidth]{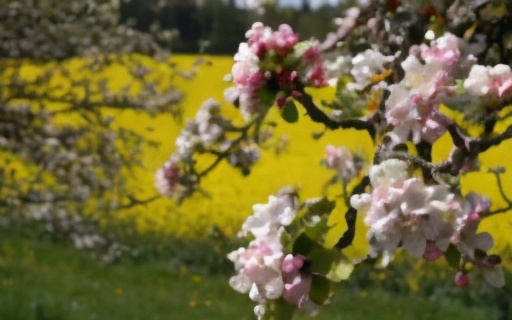} & 
         \includegraphics[width=0.27\linewidth,height=0.16\linewidth]{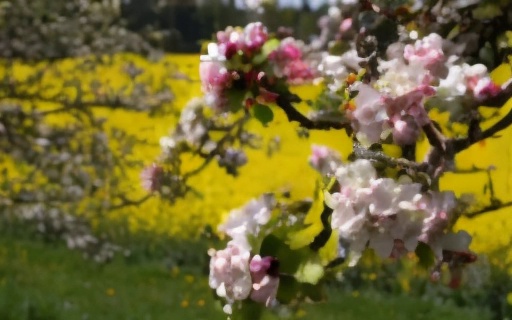} & \multirow{1}{*}{\rotatebox{-90}{$t=7$}} \\
         & 
         \includegraphics[width=0.27\linewidth,height=0.16\linewidth]{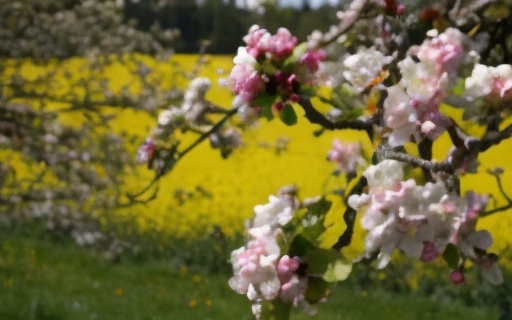} & 
         \includegraphics[width=0.27\linewidth,height=0.16\linewidth]{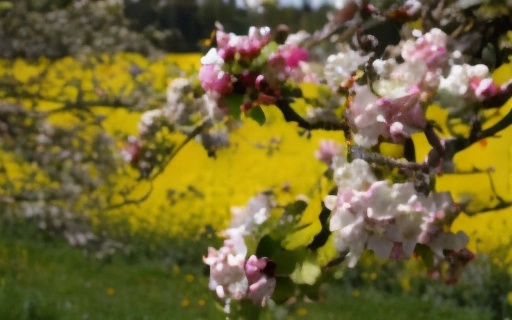} & \multirow{1}{*}{\rotatebox{-90}{$t=14$}} \\
         &
         \includegraphics[width=0.27\linewidth,height=0.16\linewidth]{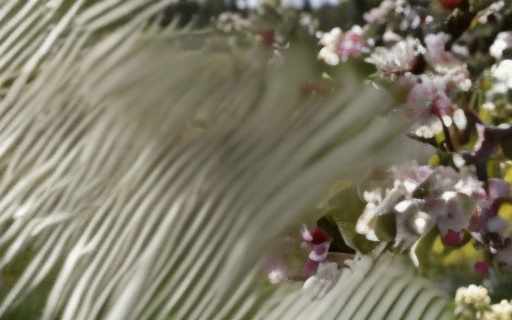} & 
         \includegraphics[width=0.27\linewidth,height=0.16\linewidth]{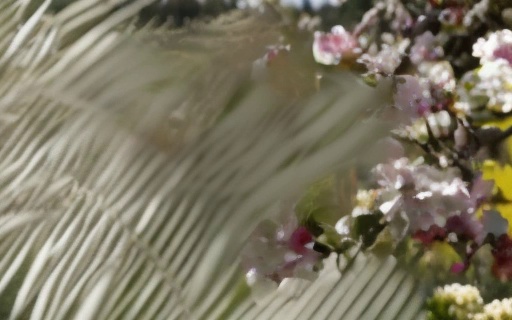} \\
    \end{tabular}
    \caption{Demonstration of generating looped stereo videos. Starting with a single reference image, we produce a stereo video clip that seamlessly loops, offering visually continuous playback.}
    \label{fig:looped}
\end{figure}

\begin{figure*}
    \centering
     \setlength{\tabcolsep}{0pt}
     \renewcommand{\arraystretch}{0}
    \begin{tabular}{cccc}
        Depth Pro & Depth Anything & DepthCrafter & Video Depth Anything \\
        \toprule
        \includegraphics[width=0.24\linewidth,height=0.13\linewidth]{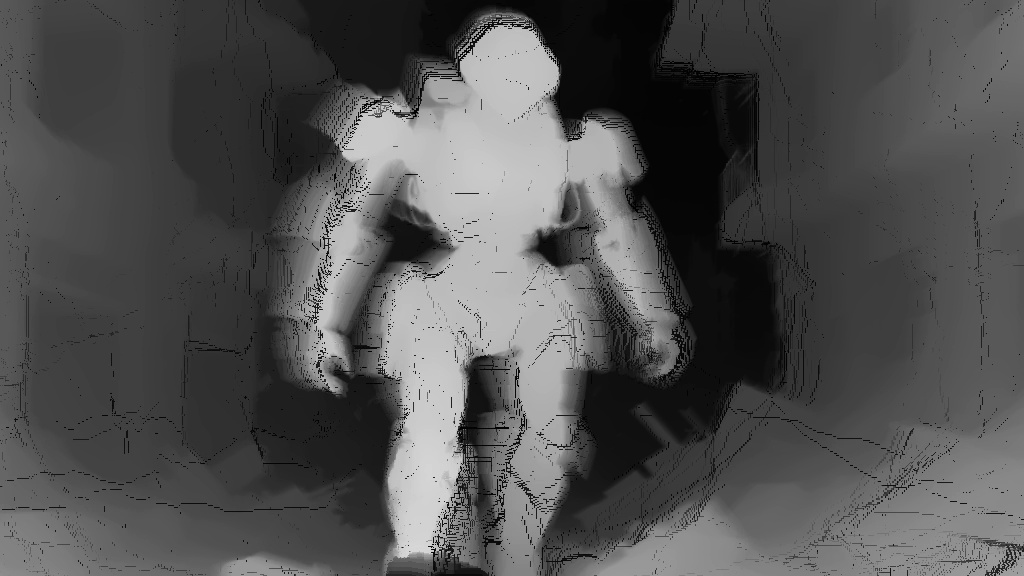}
        & 
        \includegraphics[width=0.24\linewidth,height=0.13\linewidth]{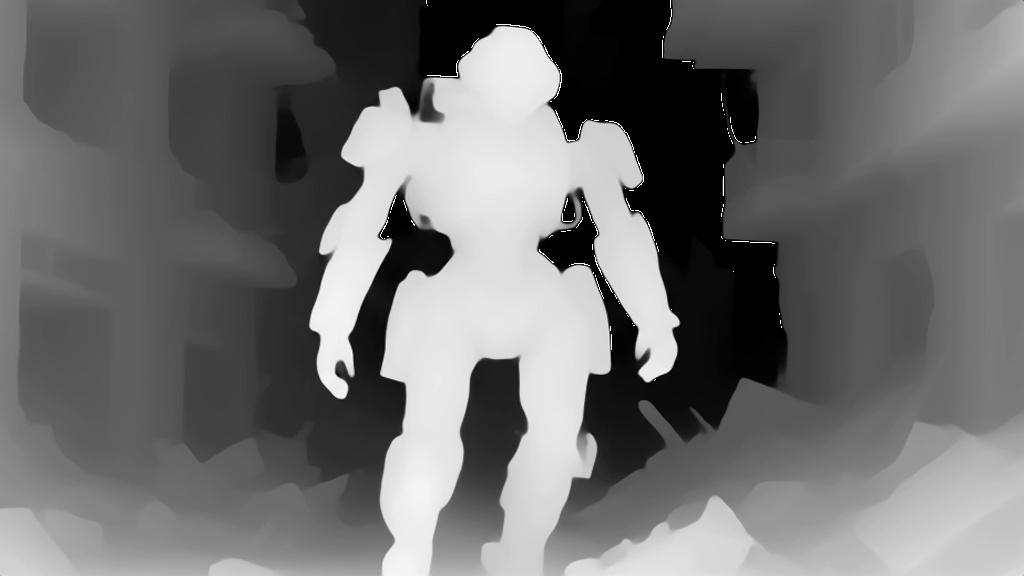} 
        & 
        \includegraphics[width=0.24\linewidth,height=0.13\linewidth]{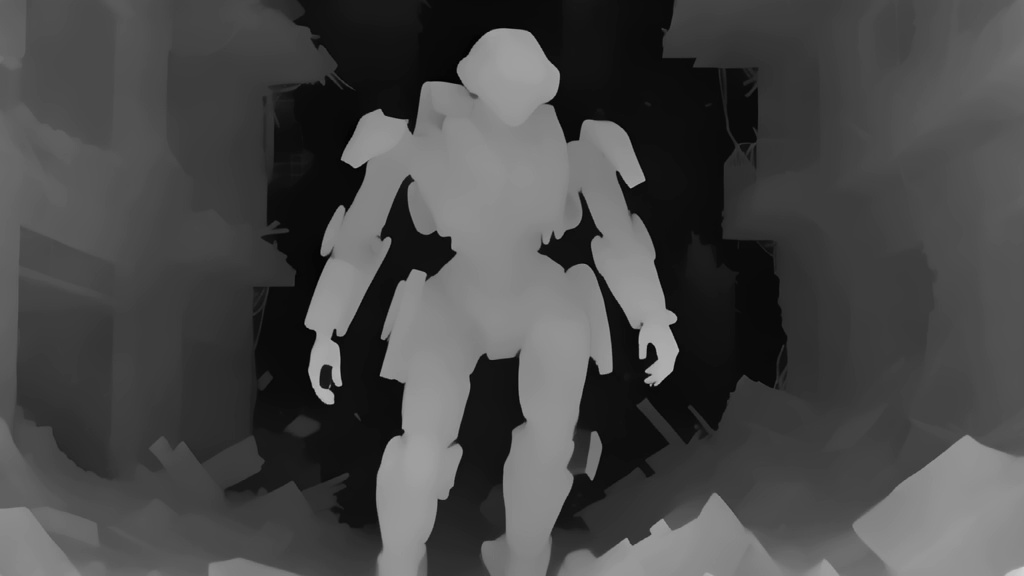} 
        & 
        \includegraphics[width=0.24\linewidth,height=0.13\linewidth]{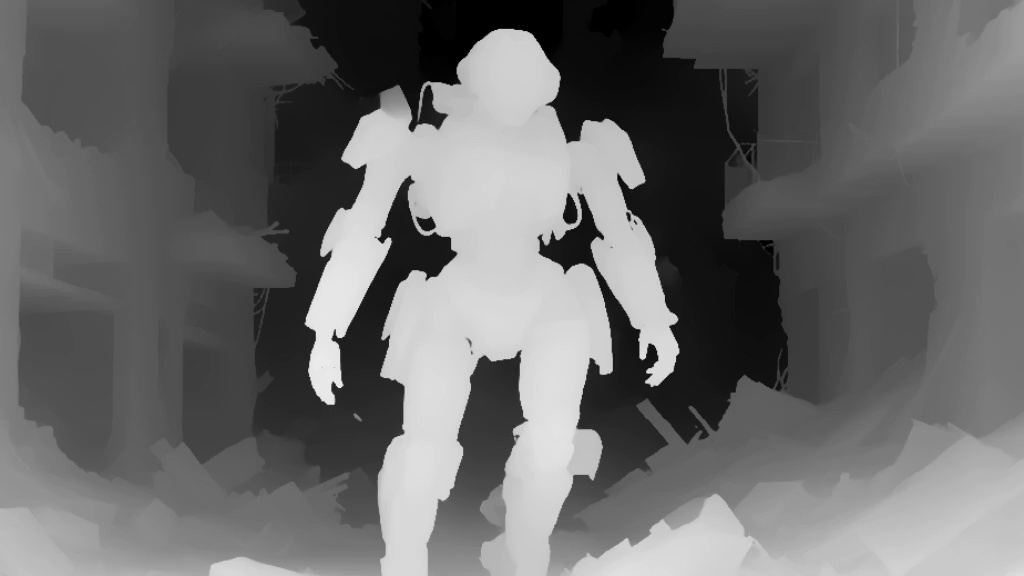} 
        \\
        \includegraphics[width=0.24\linewidth,height=0.13\linewidth]{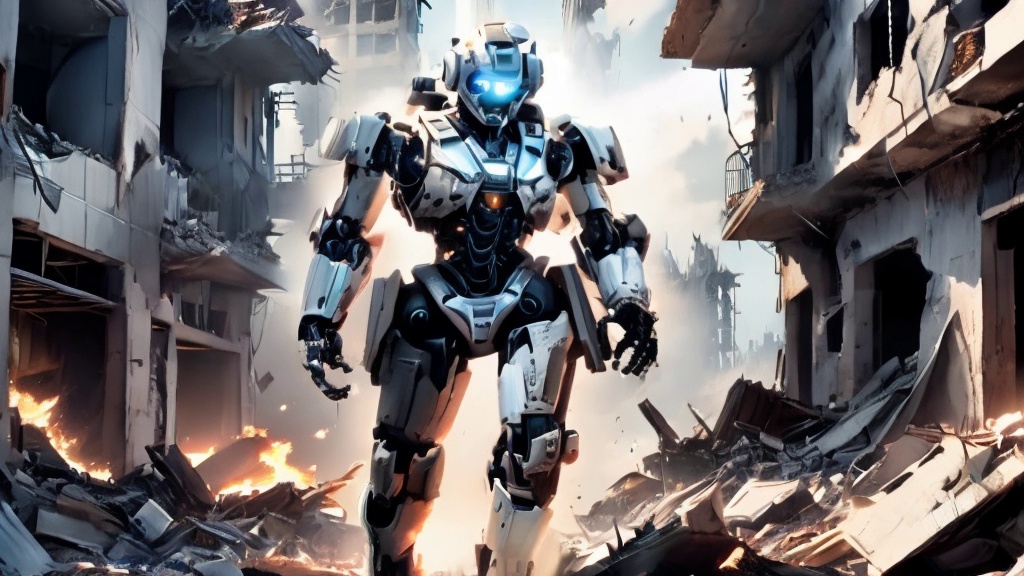}
        & 
        \includegraphics[width=0.24\linewidth,height=0.13\linewidth]{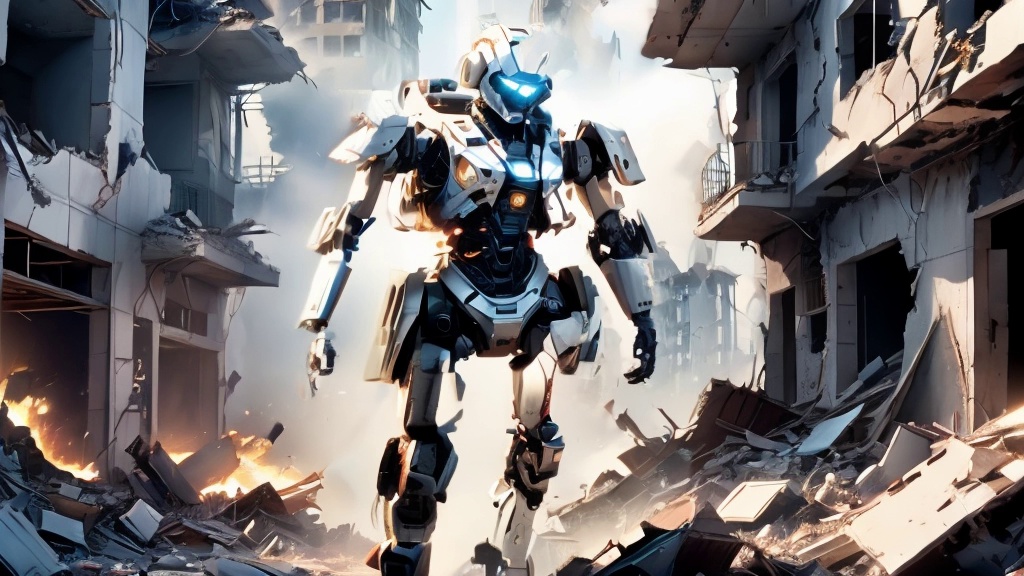} 
        & 
        \includegraphics[width=0.24\linewidth,height=0.13\linewidth]{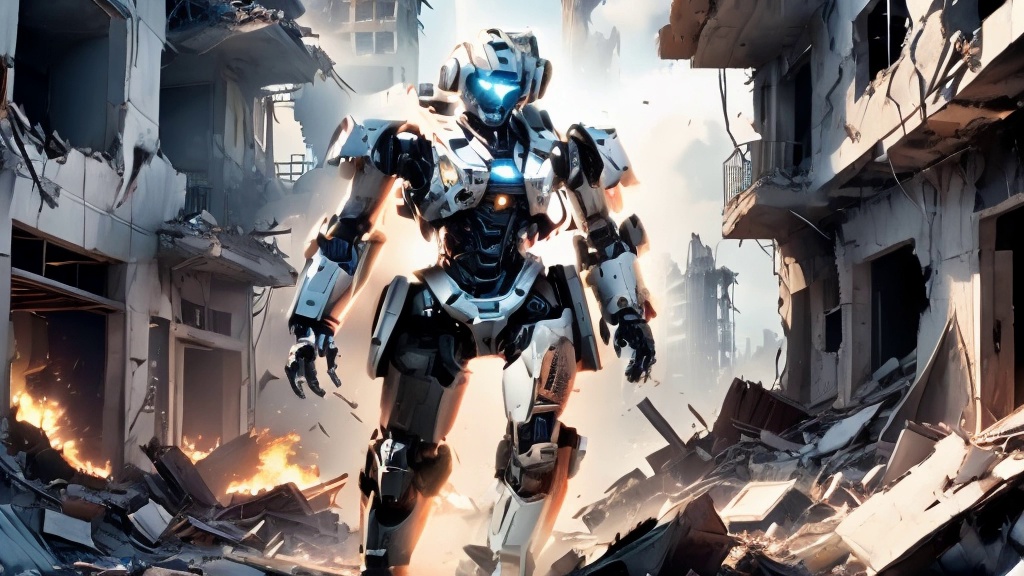} 
        & 
        \includegraphics[width=0.24\linewidth,height=0.13\linewidth]{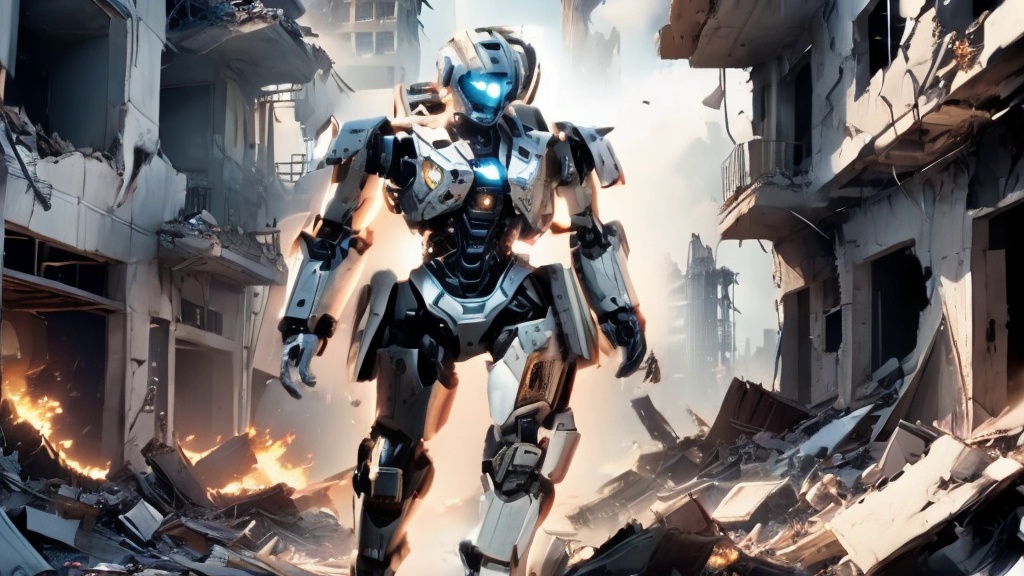}
        \\
        \includegraphics[width=0.24\linewidth,height=0.13\linewidth]{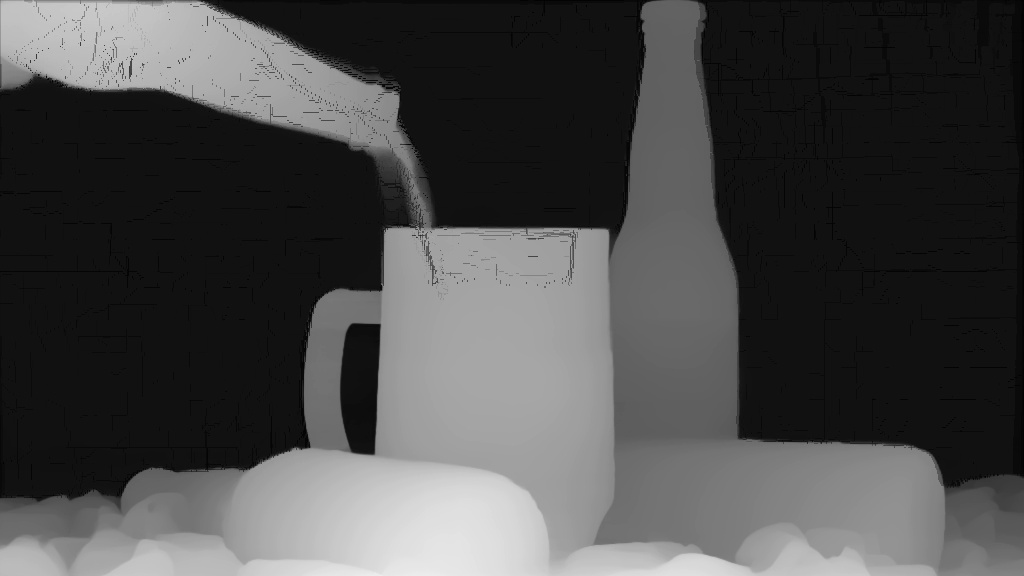}
        & 
        \includegraphics[width=0.24\linewidth,height=0.13\linewidth]{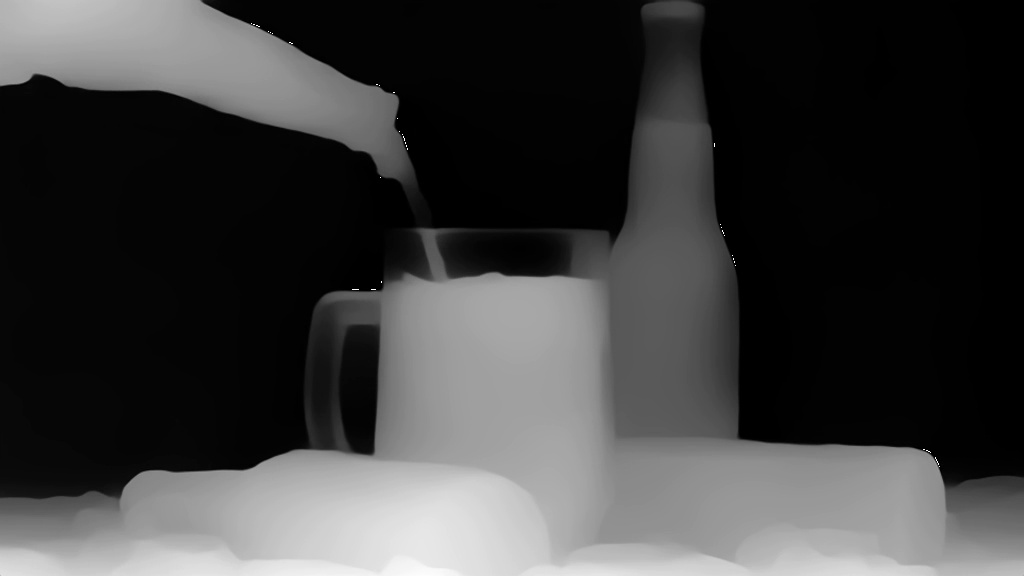} 
        & 
        \includegraphics[width=0.24\linewidth,height=0.13\linewidth]{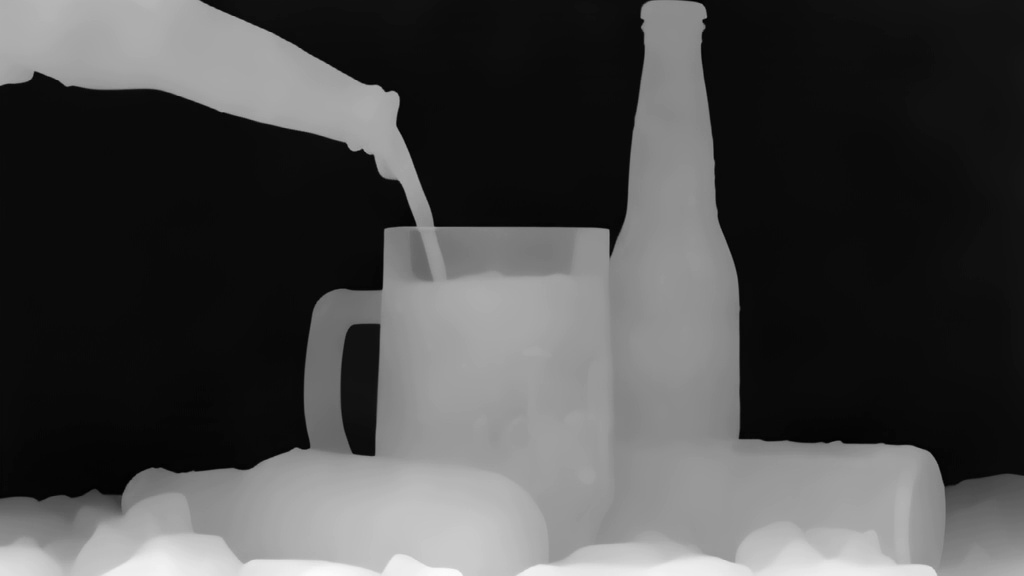} 
        & 
        \includegraphics[width=0.24\linewidth,height=0.13\linewidth]{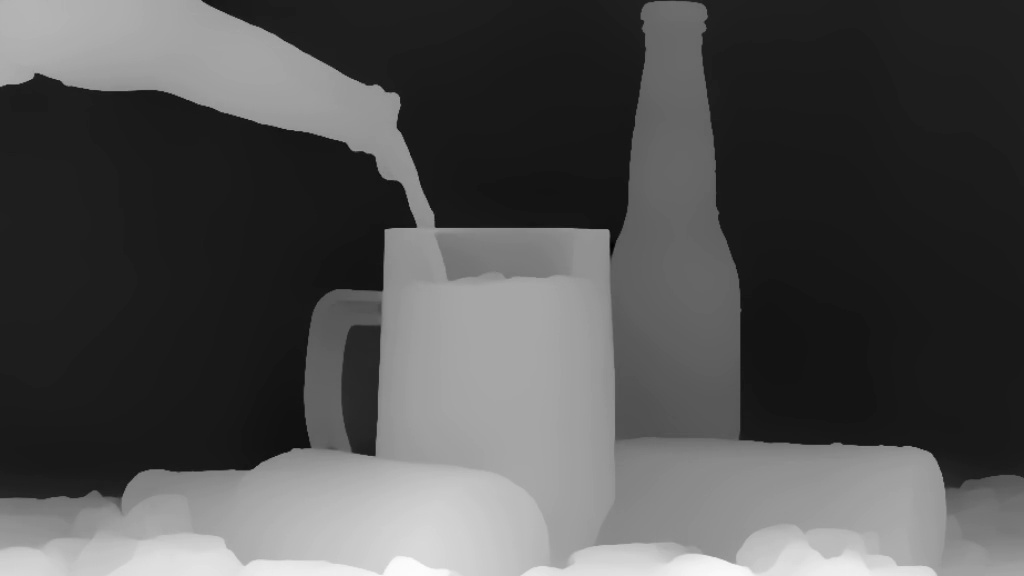} \\
        \includegraphics[width=0.24\linewidth,height=0.13\linewidth]{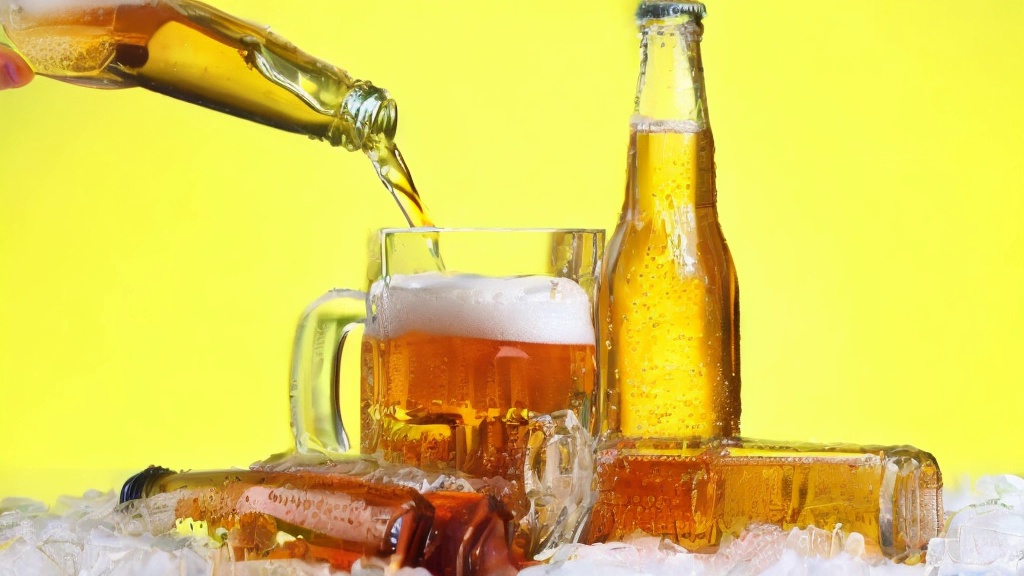}
        & 
        \includegraphics[width=0.24\linewidth,height=0.13\linewidth]{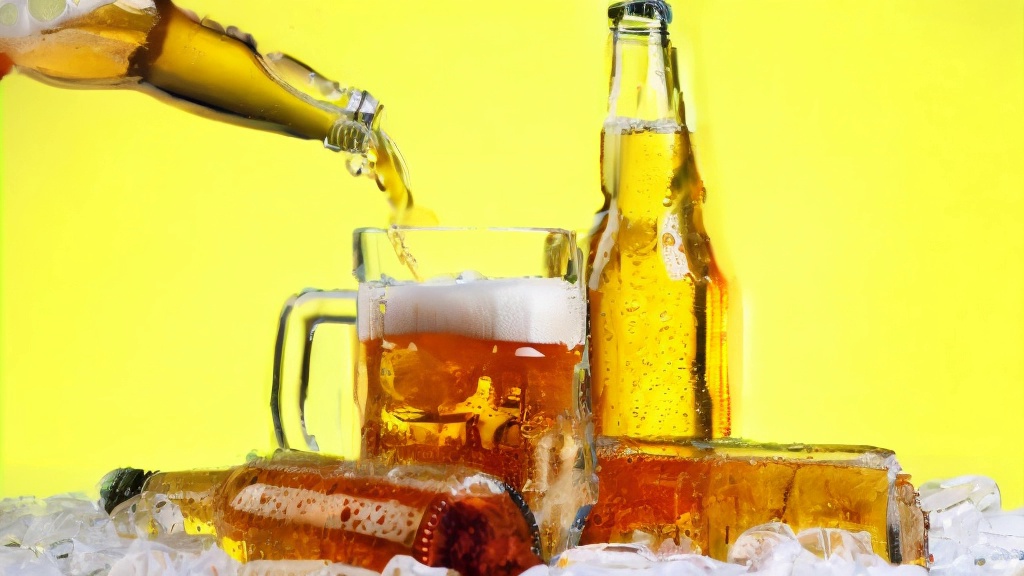} 
        & 
        \includegraphics[width=0.24\linewidth,height=0.13\linewidth]{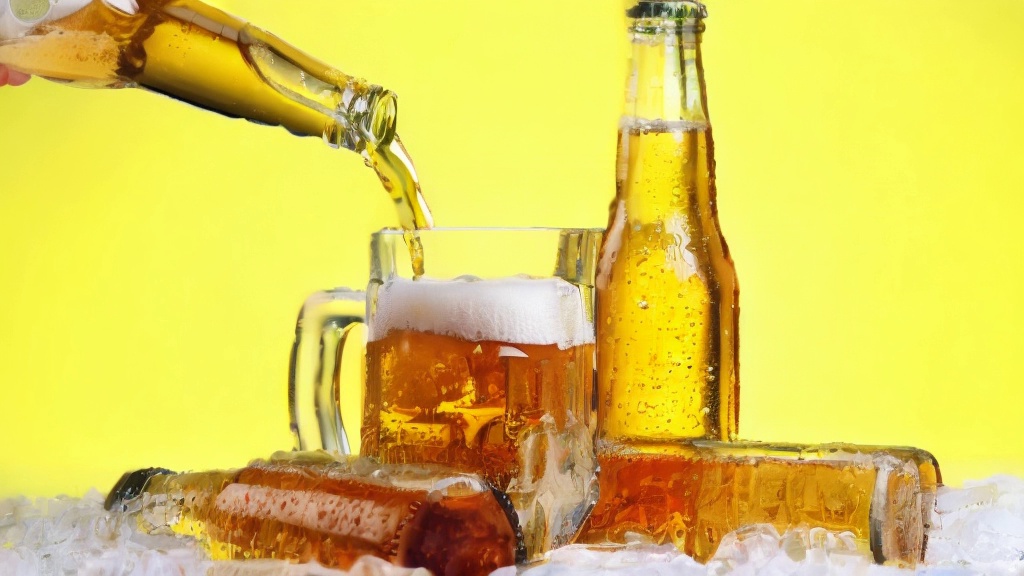} 
        & 
        \includegraphics[width=0.24\linewidth,height=0.13\linewidth]{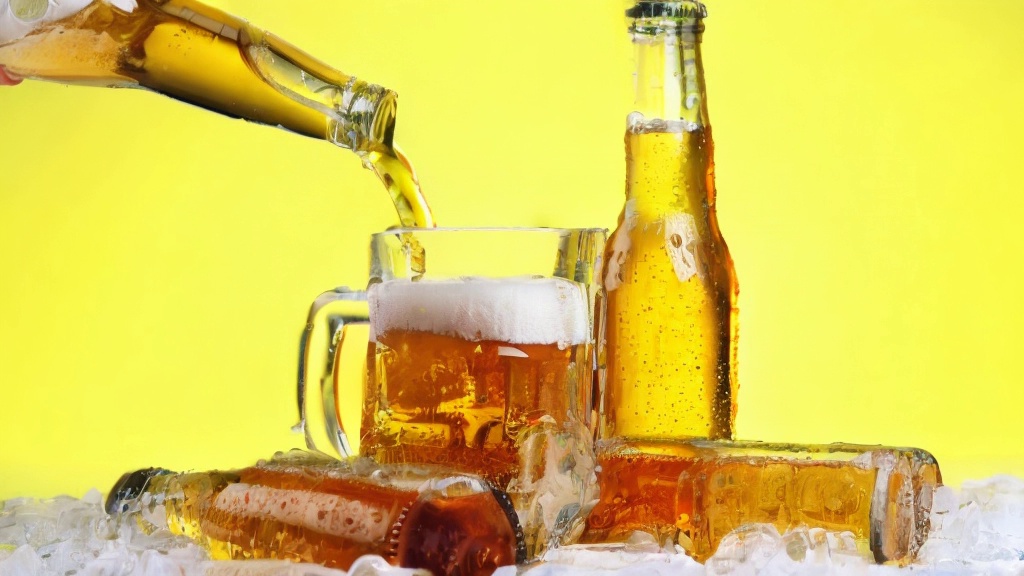}
        \\
    \end{tabular}
    \caption{A cross comparison of depth maps from the disparity propagated image-based models and the video depth models. The visual quality indicates that the DepthPro may obtain a better performance, whilst the dissolving technique is not applied.}
    \label{fig:disp_prop_showcase}
\end{figure*}

\begin{figure*}
    \centering
    \setlength{\tabcolsep}{0pt}
    \begin{tabular}{cccc}
         & $M=35$ &  $M=40$ &  $M=45$ \\
        \begin{overpic}[width=0.2\textwidth]{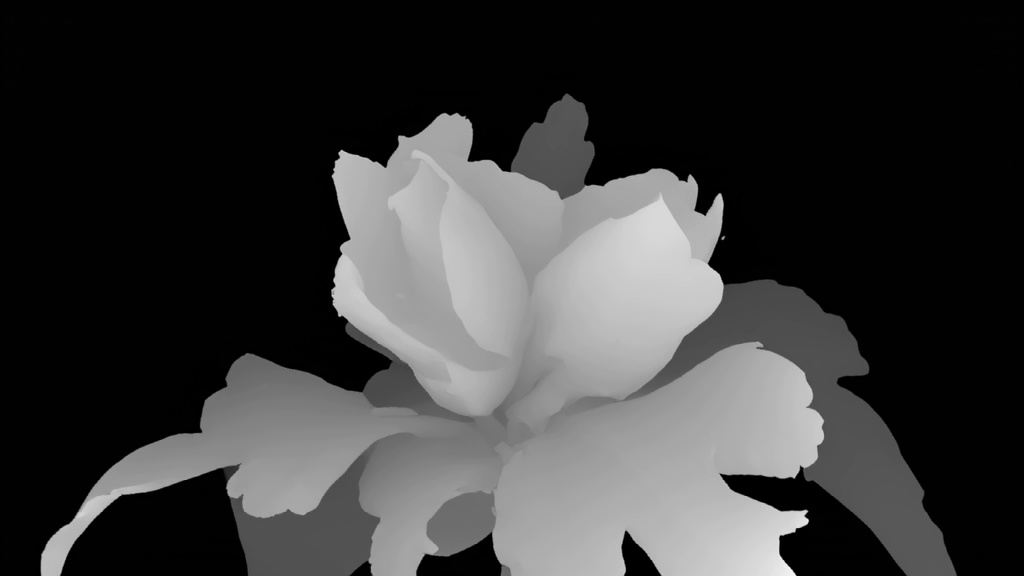}
            \put(25,60){\makebox[0pt]{\small DepthCrafter}}
        \end{overpic}
        &
        \begin{overpic}[width=.245\linewidth]{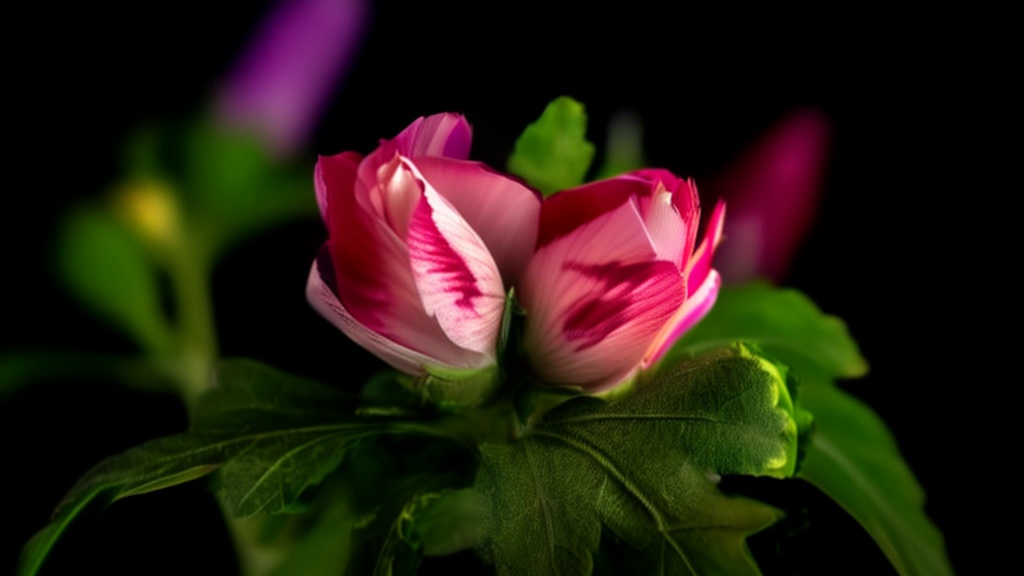}
            \put(13,1.5){\makebox[0pt]{\adjincludegraphics[width=0.055\linewidth,trim={{.28\width} {.3\height} {.6\width} {.2\height}},clip, cfbox=purple 2pt 0cm]{misc/bloom_depth_map/bloom35/frame_012__a.jpg}}}
        \end{overpic}
        & 
        \begin{overpic}[width=.245\linewidth]{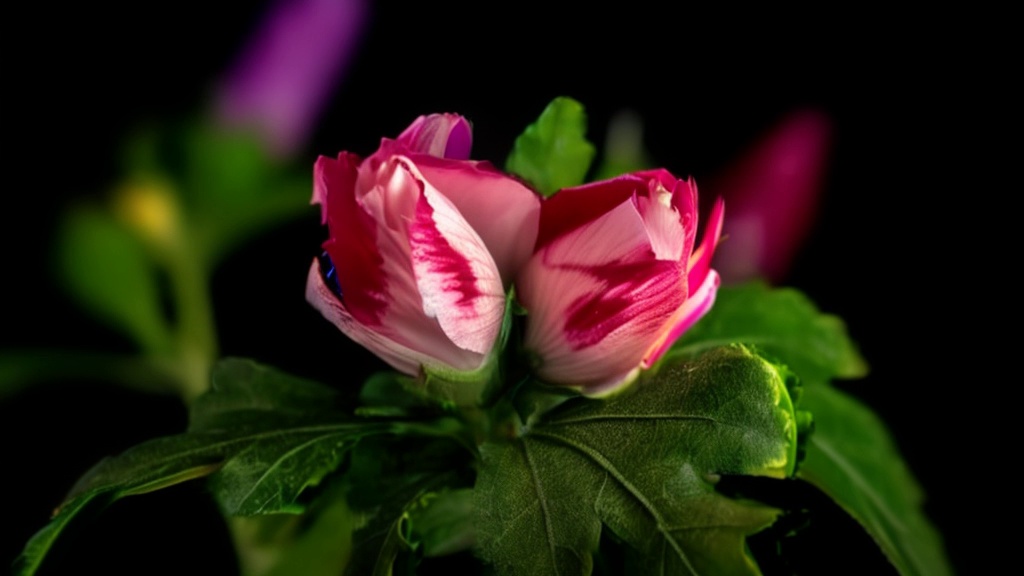}
            \put(13,1.5){\makebox[0pt]{\adjincludegraphics[width=0.055\linewidth,trim={{.28\width} {.3\height} {.6\width} {.2\height}},clip, cfbox=purple 2pt 0cm]{misc/bloom_depth_map/bloom40/frame_012__a.jpg}}}
        \end{overpic}
         & 
        \begin{overpic}[width=.245\linewidth]{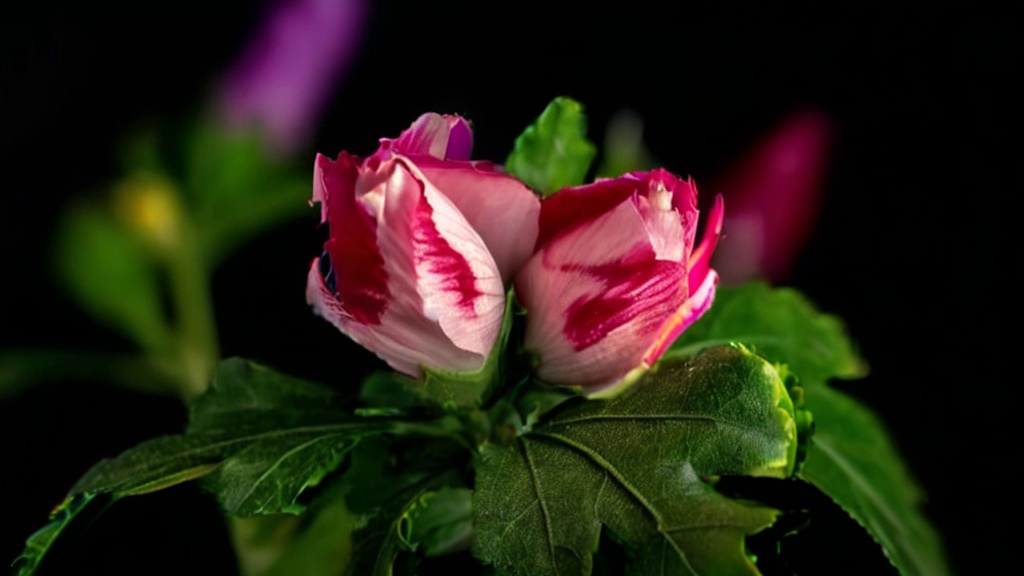}
            \put(13,1.5){\makebox[0pt]{\adjincludegraphics[width=0.055\linewidth,trim={{.28\width} {.3\height} {.6\width} {.2\height}},clip, cfbox=purple 2pt 0cm]{misc/bloom_depth_map/bloom45/frame_012__a.jpg}}}
        \end{overpic}
        \\
        \begin{overpic}[width=0.2\textwidth]{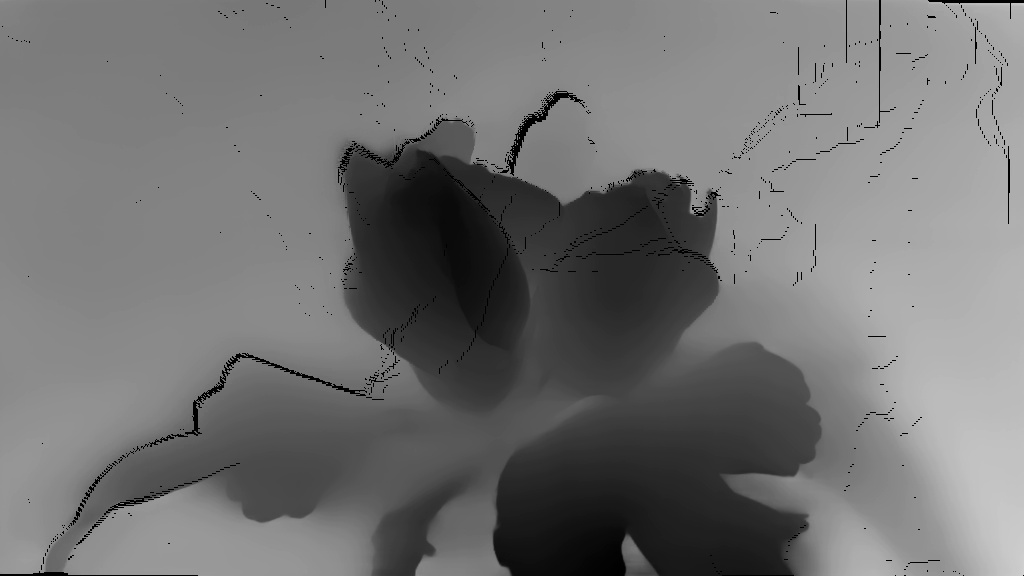}
            \put(40,60){\makebox[0pt]{\small Disparity Propagation}}
        \end{overpic}
        &
        \begin{overpic}[width=.245\linewidth]{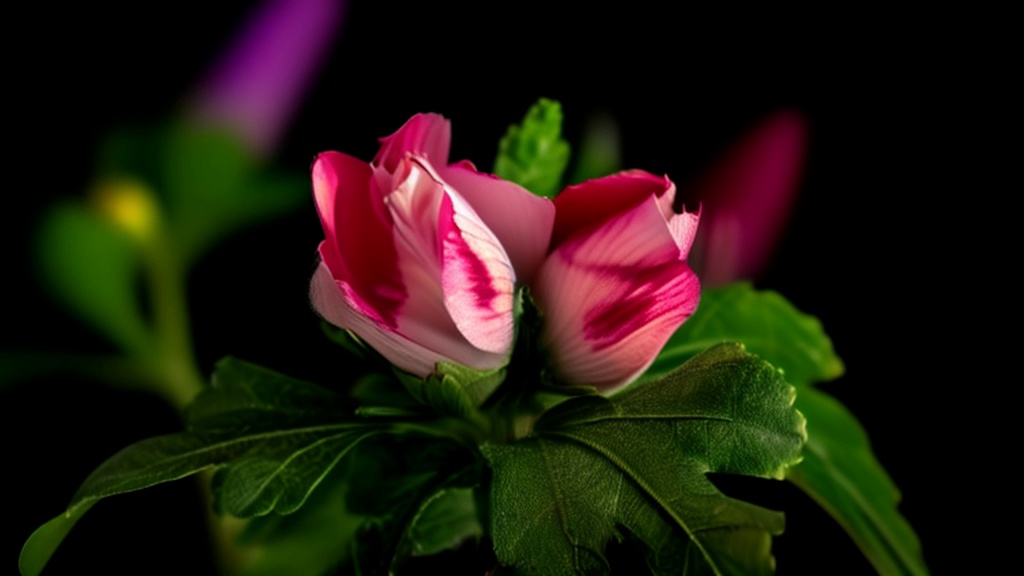}
            \put(13,1.5){\makebox[0pt]{\adjincludegraphics[width=0.055\linewidth,trim={{.28\width} {.3\height} {.6\width} {.2\height}},clip, cfbox=purple 2pt 0cm]{misc/bloom_depth_map/bloom35_prop/frame_012__a.jpg}}}
        \end{overpic}
         & 
        \begin{overpic}[width=.245\linewidth]{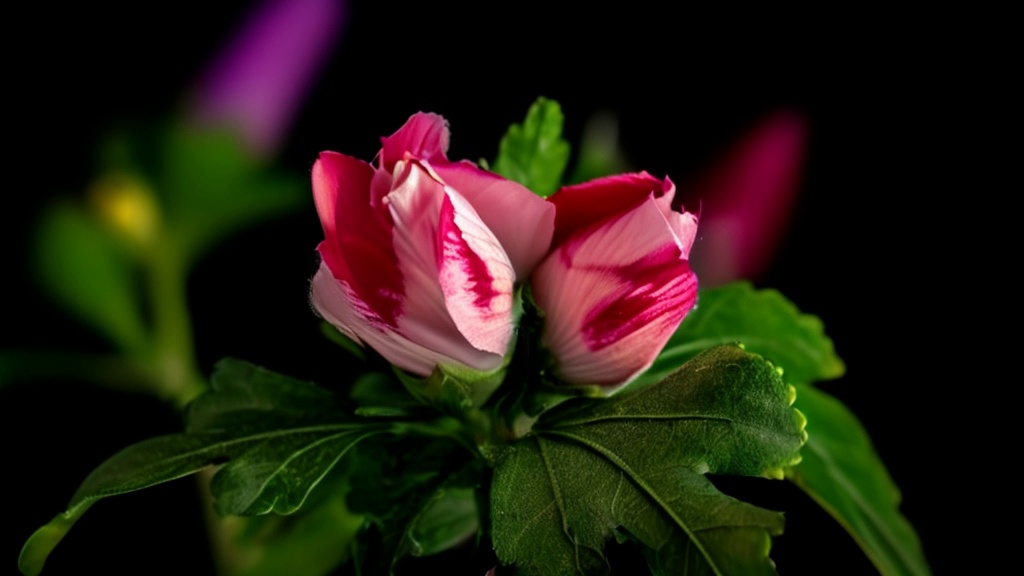}
            \put(13,1.5){\makebox[0pt]{\adjincludegraphics[width=0.055\linewidth,trim={{.28\width} {.3\height} {.6\width} {.2\height}},clip, cfbox=purple 2pt 0cm]{misc/bloom_depth_map/bloom40_prop/frame_012__a.jpg}}}
        \end{overpic}
         & 
        \begin{overpic}[width=.245\linewidth]{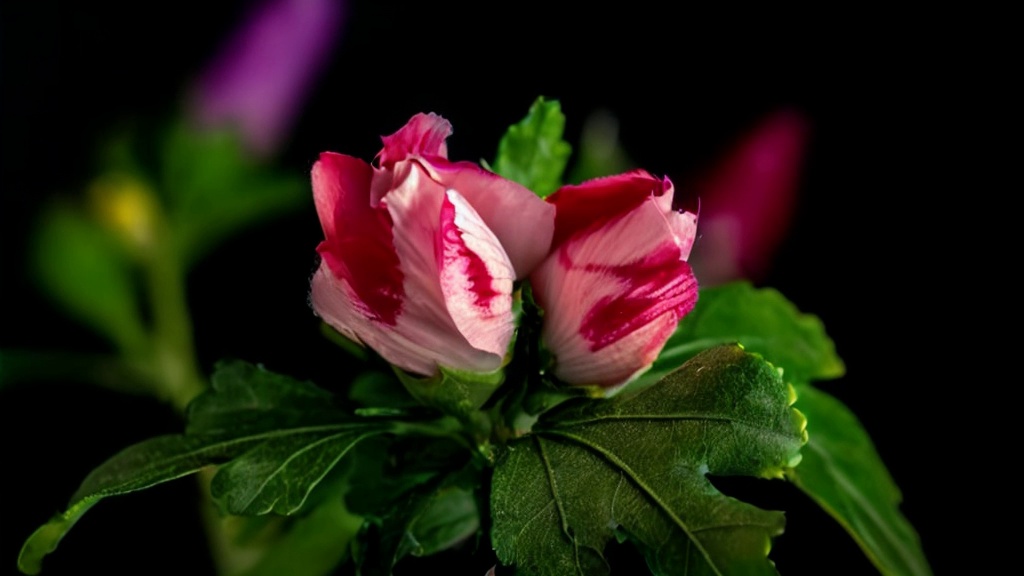}
            \put(13,1.5){\makebox[0pt]{\adjincludegraphics[width=0.055\linewidth,trim={{.28\width} {.3\height} {.6\width} {.2\height}},clip, cfbox=purple 2pt 0cm]{misc/bloom_depth_map/bloom45_prop/frame_012__a.jpg}}}
        \end{overpic}
        \\
    \end{tabular}
    \caption{Demonstration of the differences between using DepthCrafter and the Disparity Propagation method for obtaining disparity values. $M$ denotes the early-stop sampling step for the warping operation. Both depth images and generated samples are taken from the $12^{th}$ frame. We can see that DepthCrafter offers a better depth map, but it may create sharp textures for a larger $M$ while the disparity propagation method may mitigate such effects, though a less accurate depth map is provided.}
    \label{fig:prop_vs_crafter}
\end{figure*}

\subsection{Frame Interpolation And Looped Videos}

Our method can be adapted to different video generation methods. We demonstrate generating stereo videos by using 1) frame interpolation, which generates stereo videos with a starting frame and an ending frame, and 2) looped videos, which generate a stereo video that keeps repeating without breaking the continuity.
We present our generated videos in~\Cref{fig:interp,fig:looped}. This video generation process operates under a stricter condition, wherein only the required frames are fixed, while all other frames are initialized to zeros. Our approach follows a similar configuration as outlined in the main text. In addition, once the results are obtained, we remove the fixed frame indices because such hard constraints can disrupt the smoothness and natural continuity of the generated videos. From our experiments, we observe that those hard constraints eliminate the stereoscopic effects on those frames. This step ensures a more coherent and visually appealing video output while adhering to the imposed frame constraints during the generation process.

\subsection{Extend To Other Video Diffusion Models}

\begin{figure}[b]
    \centering
     \setlength{\tabcolsep}{0pt}
    \begin{tabular}{ccc}
         \includegraphics[width=0.33\linewidth]{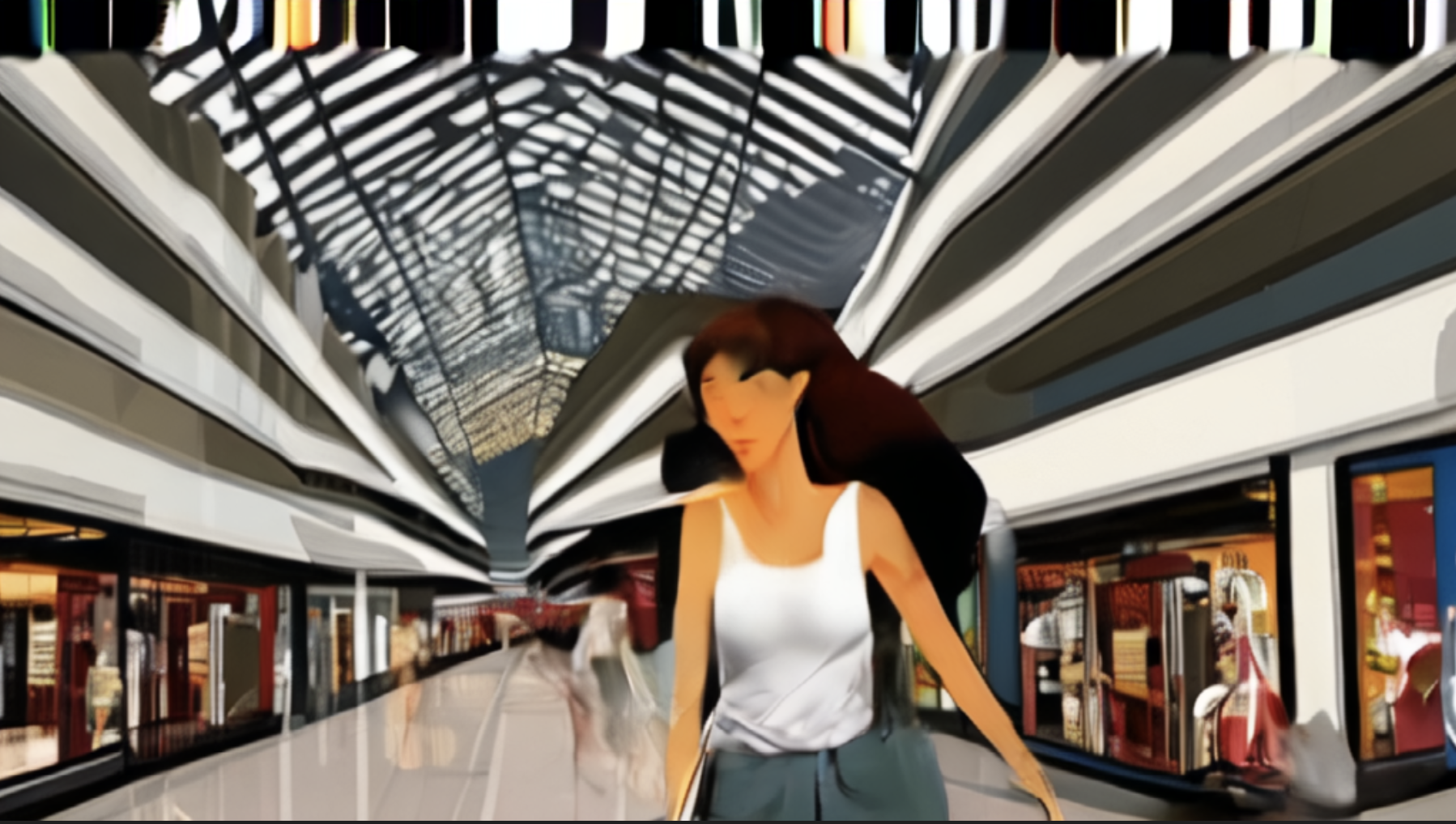}
         &
         \includegraphics[width=0.33\linewidth]{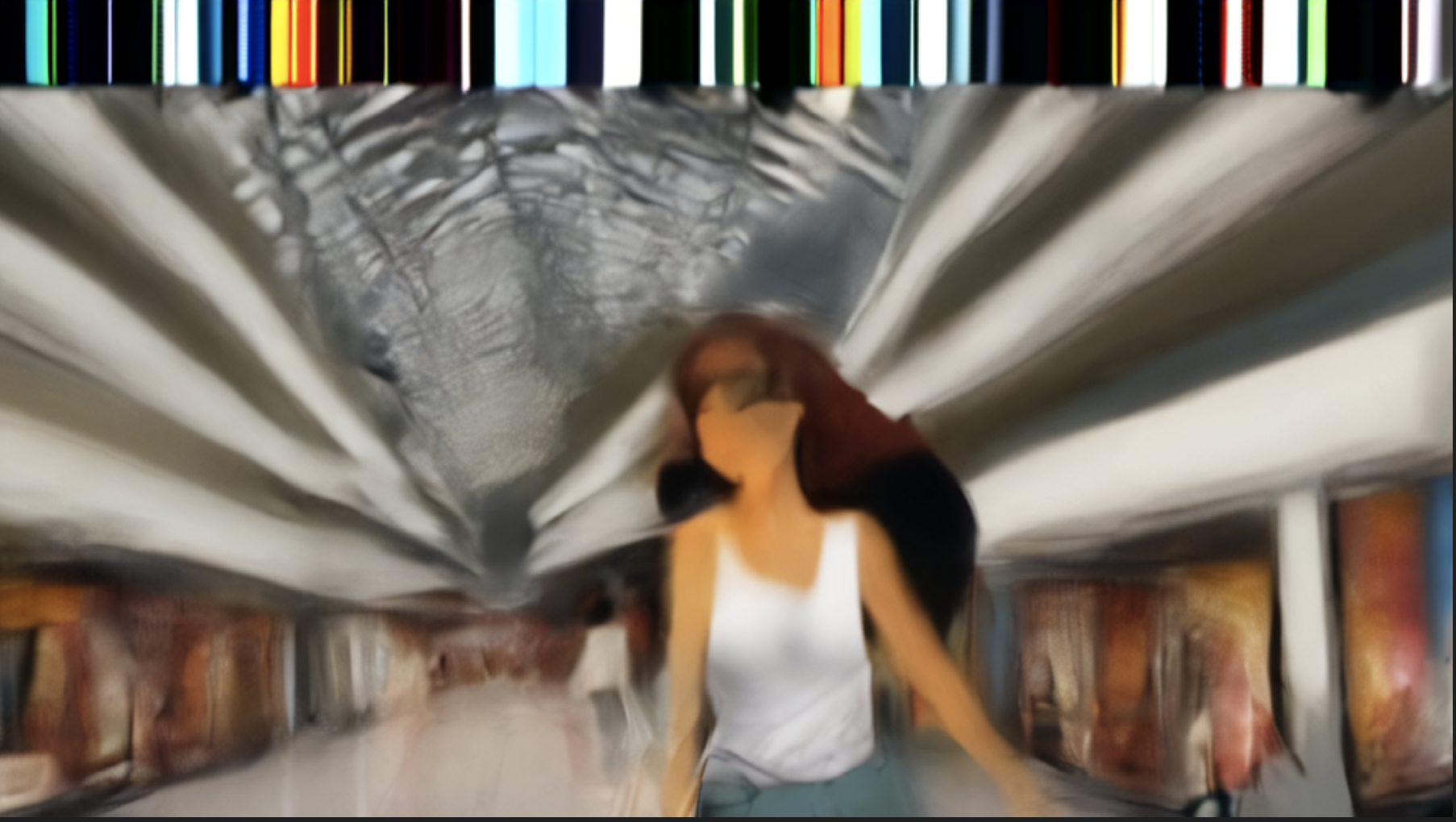}
         &
         \includegraphics[width=0.33\linewidth]{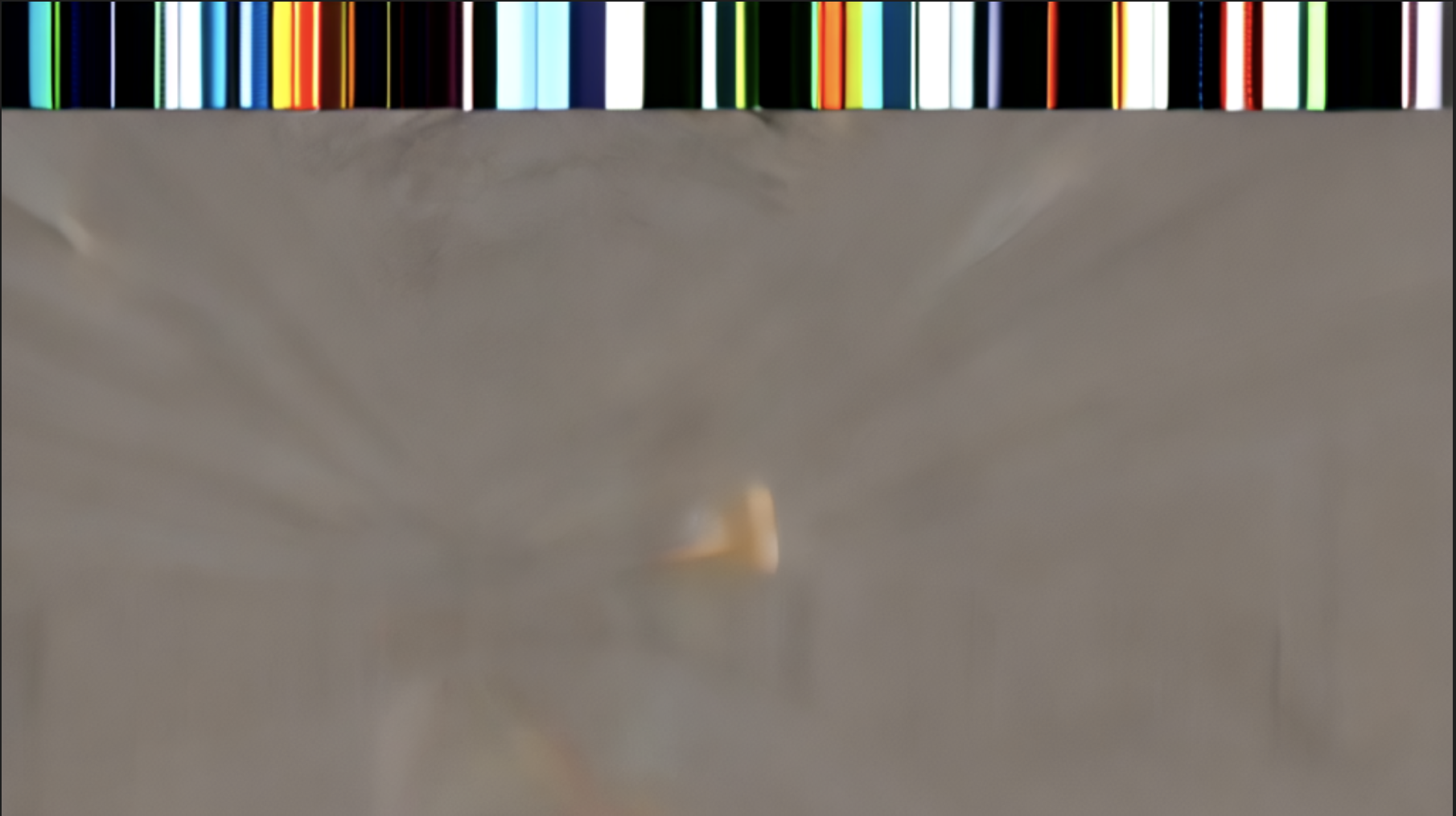}
         \\
         $np=3$ & $np=6$ & $np=8$ \\
    \end{tabular}
    \caption{Warp LVDM latent space with direct translation with border padding on $t=10$. $np$ denotes the number of rows of pixels that are shifted.}
    \label{fig:lvdm-crash}
\end{figure}

\paragraph{A Pitfall In Vanilla Latent Video Diffusion Models}
Our method can be extended to other diffusion-based models. However, whilst experimenting with the naive latent video diffusion models (LVDMs), we found that LVDMs can not handle repetitive patterns in the latent space properly.
This can happen when inpainting the warped latent features with the source view latent.
To further illustrate that, we experiment by applying an affine transformation (\textit{s.t} translation) to the latent space with border padding. We can observe that the generated images are degenerated, as shown in~\Cref{fig:lvdm-crash}.

\begin{figure}[b]
    \centering
    \begin{overpic}[width=.9\linewidth,trim={0 8em 0 0},clip]{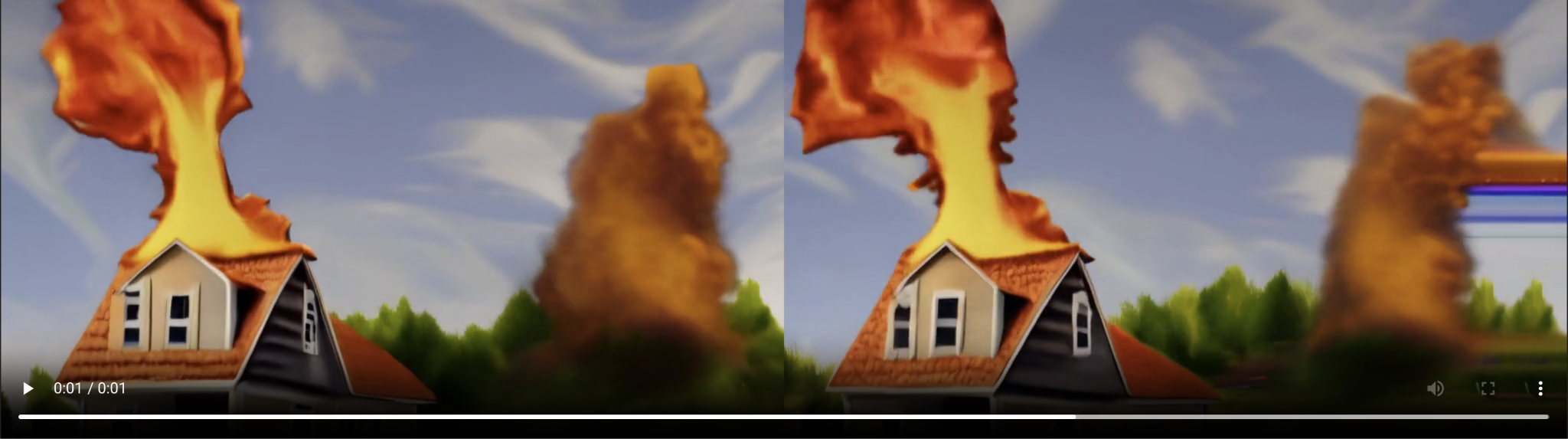}
    \put(92,1.5){\makebox[0pt]{\adjincludegraphics[height=0.235\linewidth,trim={{.8\width} 15em 0 0},clip, cfbox=red 2pt 0cm]{misc/LVDM/demo.png}}}
    \put(43,1.5){\makebox[0pt]{\adjincludegraphics[height=0.235\linewidth,trim={{.3\width} 15em {.5\width} 0},clip, cfbox=red 2pt 0cm]{misc/LVDM/demo.png}}}
    \end{overpic}
    \caption{Demo of repetitive latent warping for LVDMs. Left: filling with source view (w/o repetitive patterns). Right: filling with random noise (w/ repetitive patterns).}
    \label{fig:lvdm-demo}
\end{figure}

In our experiment, such an issue can be addressed by preventing the potential repetitive patterns, by inpainting with random Gaussian noises, or by our proposed noisy restart method (which uses both the source view latent and a small amount of random Gaussian noises). We show the improved results in~\Cref{fig:lvdm-demo}.

\paragraph{Enforcing Larger Motion with MOG-VFI}

To enforce stronger motion, we adapted those components to a more recent model, namely MoG-VFI, for video generation via frame interpolation. We first generated video clips using Google Veo3, then took the 1st and 24th frames as inputs to MoG-VFI to enforce a stronger motion. Even without careful hyperparameter tuning, the adapted model achieved a MEt3R score of 5.08, suggesting our method can transfer beyond the DynamiCrafter instantiation and support larger motion videos. The sample adaptation code is provided in our GitHub repository.

\section{Perfect Depth Maps Can Be Imperfect}

This section introduces a disparity propagation approach that can generate depth maps with consistent depth values, then we compare this solution against the video depth estimation approach.

\subsection{Disparity Propagation}
\label{sec:disp_prop}

With the aim of a more stable video with fewer artifacts, we first obtain a video depth map that allows smoother transitions and minimizes abrupt depth changes.
Therefore, we propose \textit{disparity propagation}, which propagates depth information from a single, initial depth frame across the entire video using frame-by-frame optical flow warping. This approach provides a smoother, more consistent disparity map over time, as it mitigates frame-specific artifacts by avoiding the per-frame recalculation of depth.

Given an initial depth map $D_t$ for the frame at time $t$, we estimate the depth map at the previous $t+1$ frame and the next $t-1$ frame by warping $D_t$ based on the forward and backward optical flow, $F_{t \to t+1}$ and $F_{t \to t-1}$.
We use a warping operation to apply the optical flow $F_{t \to t+1}$ to adjust $D_t$ based on the motion between the frames. The forward warping function can be expressed as:
\begin{equation}
     D_{t+1}(x, y) = D_t\big(x + F_{t \to t+1}^x(x, y), \; y + F_{t \to t+1}^y(x, y)\big).
\end{equation}
Note that the backward warping $D_t$ to $D_{t-1}$ can also be achieved with this operation since $F_{t+1 \to t} = - F_{t \to t+1}$.
We repeat the process for each precedent and subsequent frame to propagate the initial depth across the entire video sequence. This produces a sequence of disparity maps $S_D=\{D_0, D_1, \dots, D_T\}$ that are inherently smooth and consistent over time.
In our work, we compute the disparity propagated depth maps for each frame, obtaining $\{S_D^{0}, .., S_D^{T}\}$. For each frame at time $t$, we take an average of $\{S_D^{0}[t], .., S_D^{T}[t]\}$ as the final depth.

\subsection{Disparity Propagation vs. Video Depth Estimation}
Existing video depth estimation models often overemphasize subtle scene movements, such as the drift of particles or minor background elements.

The direct use of video depth estimation results may mistakenly warp the low-frequency latent component using the prediction of high-frequency depth maps.
This can lead the model to assign high importance to irrelevant objects or minor details, which misleads the generation model into emphasizing non-crucial elements.
Moreover, although these models generally produce consistent depth estimates across the scene, they occasionally introduce significant artifacts. 
For instance, objects may appear to shift abruptly from far to near within the frame, or objects ignored in previous frames may suddenly appear, leading to sudden jumps in perceived depth. Such inconsistencies can result in collapsed or distorted video generation.
As shown in~\Cref{fig:prop_vs_crafter}, the depth map generated from a video depth estimator may mess up the border values due to the potential inconsistency around the edges, which is especially crucial for warping since the values may be moved to the wrong location. 

In our current implementation, we still prefer a video depth estimator since this depth propagation approach may fail to generate a meaningful disparity map to handle the scenarios that come with strong motion. 
We demonstrate different disparity propagated depth and video depth estimation results in~\cref{fig:disp_prop_showcase}. However, supported by our aforementioned experiment, one potential solution might be employing a progressive depth map that can gradually increase the depth information for high-frequency objects. In the future, we believe a progressive depth map may further benefit the generation process.

\end{document}